\documentclass{article}

\PassOptionsToPackage{numbers,compress}{natbib}
\usepackage[preprint]{neurips_2026}

\usepackage[utf8]{inputenc}
\usepackage[T1]{fontenc}
\usepackage{hyperref}
\hypersetup{hypertexnames=false}
\newcommand{\mainref}[1]{\ref{#1}}
\usepackage{url}
\usepackage{booktabs}
\usepackage{amsfonts}
\usepackage{nicefrac}
\usepackage{microtype}
\usepackage{xcolor}
\usepackage{graphicx}
\usepackage{subcaption}
\usepackage{amsmath}
\usepackage{amsthm}
\usepackage{ragged2e}
\usepackage{algorithm}
\usepackage{algpseudocode}
\usepackage{tabularx}
\usepackage{longtable}
\usepackage{array}
\usepackage{makecell}
\usepackage{amssymb}
\usepackage{multirow}
\usepackage{enumitem}
\usepackage{threeparttable}
\usepackage{siunitx}
\usepackage{caption}
\usepackage{listings}
\usepackage{tikz}
\usetikzlibrary{arrows.meta,positioning}
\usepackage{tcolorbox}
\tcbuselibrary{skins, breakable}

\newcolumntype{Y}{>{\RaggedRight\arraybackslash}X}
\newcolumntype{N}{S[
  mode=text,
  detect-family=true,
  detect-weight=true,
  table-format=1.2,
  table-number-alignment=center
]}

\newtcolorbox{promptbox}[2][]{
    colback=white,
    colframe=gray!50,
    coltitle=black,
    fonttitle=\bfseries,
    title=#2,
    breakable,
    enhanced,
    sharp corners,
    boxrule=0.5pt,
    left=3mm, right=3mm, top=3mm, bottom=3mm,
    width=\linewidth,
    grow to right by=0pt,
    grow to left by=0pt,
    #1
}

\lstdefinelanguage{json}{
  basicstyle=\ttfamily\small,
  numbers=left,
  numberstyle=\tiny,
  numbersep=5pt,
  frame=single,
  breaklines=true,
  showstringspaces=false,
  tabsize=2,
  morestring=[b]",
  literate=
    *{0}{{{\color{orange!70!black}0}}}{1}
     {1}{{{\color{orange!70!black}1}}}{1}
     {2}{{{\color{orange!70!black}2}}}{1}
     {3}{{{\color{orange!70!black}3}}}{1}
     {4}{{{\color{orange!70!black}4}}}{1}
     {5}{{{\color{orange!70!black}5}}}{1}
     {6}{{{\color{orange!70!black}6}}}{1}
     {7}{{{\color{orange!70!black}7}}}{1}
     {8}{{{\color{orange!70!black}8}}}{1}
     {9}{{{\color{orange!70!black}9}}}{1}
     {:}{{{\color{black}:}}}{1}
     {,}{{{\color{black},}}}{1}
     {\{}{{{\color{black}\{}}}{1}
     {\}}{{{\color{black}\}}}}{1}
     {[}{{{\color{black}[}}}{1}
     {]}{{{\color{black}]}}}{1},
}

\title{ProCARE: Real-World Study Automation via Profile-Grounded Evidence Contracts}

\author{%
Zhaoyu Fan\textsuperscript{1,3}\thanks{Equal contribution.} \enspace
Bowen Han\textsuperscript{1,3}\footnotemark[1] \enspace
Jincheng Ou\textsuperscript{1} \enspace
Yanwei Ren\textsuperscript{1,2} \enspace
Zehua He\textsuperscript{1} \enspace
Kaiyu Huang\textsuperscript{1} \\
\bfseries Pengcheng He\textsuperscript{1} \enspace
Shanshan Ni\textsuperscript{1} \enspace
Chengchen Gong\textsuperscript{1} \enspace
Shuai Xue\textsuperscript{3} \enspace
Qingjiang Shi\textsuperscript{1,4}\thanks{Corresponding author. Email: \texttt{shiqj@tongji.edu.cn}.}\\
\textsuperscript{1}Tongji University\\
\textsuperscript{2}University of Electronic Science and Technology of China\\
\textsuperscript{3}Shanghai Hengfang Health Technology Co., Ltd.\\
\textsuperscript{4}Shenzhen Research Institute of Big Data
}

\begin{document}
\maketitle
\begingroup
\renewcommand{\thefootnote}{}
\footnotetext[0]{GitHub repository: \url{https://github.com/ZyFanTJ/ProCARE}.}
\endgroup
\begin{abstract}
Real-world studies (RWS) analyze routinely collected healthcare data, such as electronic health records and insurance claims, to estimate treatment effects, monitor safety, and identify risk factors outside randomized trials. Automating RWS with large language model agents is appealing but brittle: valid analyses depend on dataset-specific clinical semantics, including index dates, follow-up windows, patient/event-level units, and outcome boundaries that are often implicit in observational data. This leads to schema mismatches, temporal errors, inconsistent analytical granularity, and unsupported clinical claims. We present ProCARE, a profile-grounded framework that reformulates RWS automation as constrained compilation. ProCARE converts heterogeneous data profiles into Evidence Contracts specifying available variables, temporal attributes, distributions, and evidentiary limits, and compiles study objectives into RWS-IR, a strongly typed intermediate representation for variables, time relations, analysis units, and statistical tasks. Deterministic validation and localized repair then check and revise generated workflows across schema mapping, cohort construction, variable derivation, analysis, and interpretation. On 36 EHR-based benchmark tasks, ProCARE outperforms strong LLM-agent baselines in code executability, semantic consistency, and conclusion traceability, with automated violation diagnoses showing moderate-to-substantial agreement with blinded expert review.
\end{abstract}

\section{Introduction}

Real-world studies (RWS) derive clinical evidence from observational healthcare data such as electronic health records and insurance claims~\cite{sherman2016real}. Unlike generic table analysis, RWS workflows are constrained by relational structure, clinical variable semantics, temporal anchors, analysis grain, and evidence scope. Cohorts must be assembled across heterogeneous tables, variables must be assigned valid clinical roles, exposure and follow-up must be anchored consistently, and reported findings must remain within the support of the executed analysis. These requirements make RWS automation a structured scientific generation problem rather than a plain code-generation task.

LLM-based agents have been applied to data science and scientific workflow automation~\cite{zheng2023largelanguagemodelsscientific,jiang2025medagentbench,kim2025limitations}, but their standard interfaces do not preserve all constraints required by schema-rich clinical data. Textualizing multi-table datasets can obscure entity keys, visit structure, table linkage, missingness, and time-dependent variables~\cite{rajkomar2018scalable,hripcsak2015observational,bodensohn2025unveiling}. RWS requires domain-specific reasoning about endpoint definitions, censoring, confounding, proxy variables, and claim scope~\cite{singhal2023large,wang2023scientific}. We observe three recurring failure modes: \emph{query-to-specification failures}, where study concepts are bound to missing, incompatible, or ambiguous fields; \emph{specification-to-code failures}, where free-form code violates the declared join graph, analysis grain, temporal ordering, or artifact protocol; and \emph{artifact-to-claim failures}, where descriptive statistics or proxy artifacts are converted into unsupported clinical claims. These failures indicate missing contracts between the data, study specification, executable workflow, and final report—a structural bottleneck that bounds even strong baselines like Claude Code.

ProCARE addresses this gap by formulating RWS automation as \emph{contract-valid generation}. Its object is a heterogeneous data profile $\Phi$, which records schema structure, field statistics, clinical semantics, temporal roles, validity flags, and assumptions. Rather than serving as a context summary, $\Phi$ acts as a profile-grounded evidence contract. ProCARE compiles each user goal into a typed RWS intermediate representation $Z$, turns $Z$ into executable artifacts through a profile-constrained coding harness, validates execution with deterministic static and runtime gates that produce localized violation sets $\mathcal{U}_t$, and admits report claims only through evidence certificates $e_q$ that link claims to validated artifacts, executed code, profile entries, or recorded assumptions~\cite{schmidgall2025agent}. The output is an auditable artifact bundle containing the profile, RWS-IR, generated code, execution logs, result and figure manifests, validation evidence, assumptions, evidence certificates, and final report.

\textbf{Contributions.} (i) a heterogeneous RWS profile and typed RWS-IR that define a study contract over data structure, clinical semantics, temporal roles, analysis grain, and assumptions; (ii) evidence certificates that restrict factual report claims to validated artifacts, executed code, profile entries, IR fields, or assumptions; and (iii) an end-to-end RWS evaluation over 36 HCC tasks and 18 public-health transfer tasks, measuring executability, code quality, artifact completeness, and claim traceability.
\section{Related Work}

\textbf{LLM agents for data analysis.}
Recent LLM agents combine reasoning, tool use, code execution, and feedback for multi-step data-analysis and scientific workflows~\citep{yao2023react,shinn2023reflexion,qin2024toolllm,wu2023autogen}. Data-analysis systems and benchmarks such as InfiAgent-DABench, Data Interpreter, DSBench, and DataSciBench evaluate agents that translate natural-language goals into programs, tables, visualizations, or model outputs~\citep{hu2024infiagent,hong2025datainterpreter,jing2025dsbench,zhang2025datascibench}. Despite this progress, these systems still report failures in state tracking, dependency management, numerical consistency, and data-dependent repair. In RWS, such errors become methodological failures because executable code must preserve relational structure, analysis grain, temporal anchors, clinical roles, and claim scope. ProCARE addresses this setting with contract-valid generation over typed study objects, executable artifacts, and certified claims.

\textbf{Structured and clinical-data grounding.}
Another line of work improves LLM grounding over spreadsheets, tables, SQL databases, and structured clinical records. Table-centric methods compress, encode, or retrieve from large tabular structures~\citep{li2023sheetcopilot,dong2024spreadsheetllm,su2024tablegpt2,chen2024tablerag,yu2025tablerag}, while text-to-SQL, structured-EHR, and FHIR benchmarks evaluate reasoning over relational or resource-based clinical data~\citep{lee2024ehrsqlsharedtask,yang2026ehrstruct,lee2025fhiragentbench}. These methods improve structured-data access, but they mainly target retrieval, query answering, or task-level execution. RWS automation further requires variable bindings, joins, temporal anchors, artifacts, and claims to remain valid under one study-level contract. ProCARE enforces this contract by propagating a heterogeneous profile through goal-to-IR compilation, profile-constrained execution, and evidence-certified reporting.

\textbf{RWS evidence constraints.}
RWE generation is governed by methodological, reporting, and regulatory constraints beyond table access or code execution. STROBE and RECORD emphasize transparent reporting of study design, data sources, variables, bias, missingness, and provenance~\citep{vandenbroucke2007strobe,benchimol2015record}; target-trial emulation and TARGET guidance require explicit protocols, estimands, eligibility criteria, treatment strategies, follow-up, assumptions, and observational-data mappings~\citep{hernan2016targettrial,cashin2025target}. OMOP, FHIR, and FDA guidance further emphasize standardized representation and reliable provenance for RWD use~\citep{ohdsi2026omop,hl7fhir,fda2023rwdstandards}. Recent RWE automation benchmarks test whether LLM agents can conduct observational analyses or reproducible hypothesis testing from medical databases~\citep{li2026rwebench,gim2026latch}. ProCARE incorporates these constraints into generation by compiling typed RWS-IR, localizing execution failures as contract violations, and admitting factual claims only through evidence certificates.

\section{Method}
\label{sec:method}

\begin{figure*}[t]
    \centering
    \includegraphics[width=1\linewidth]{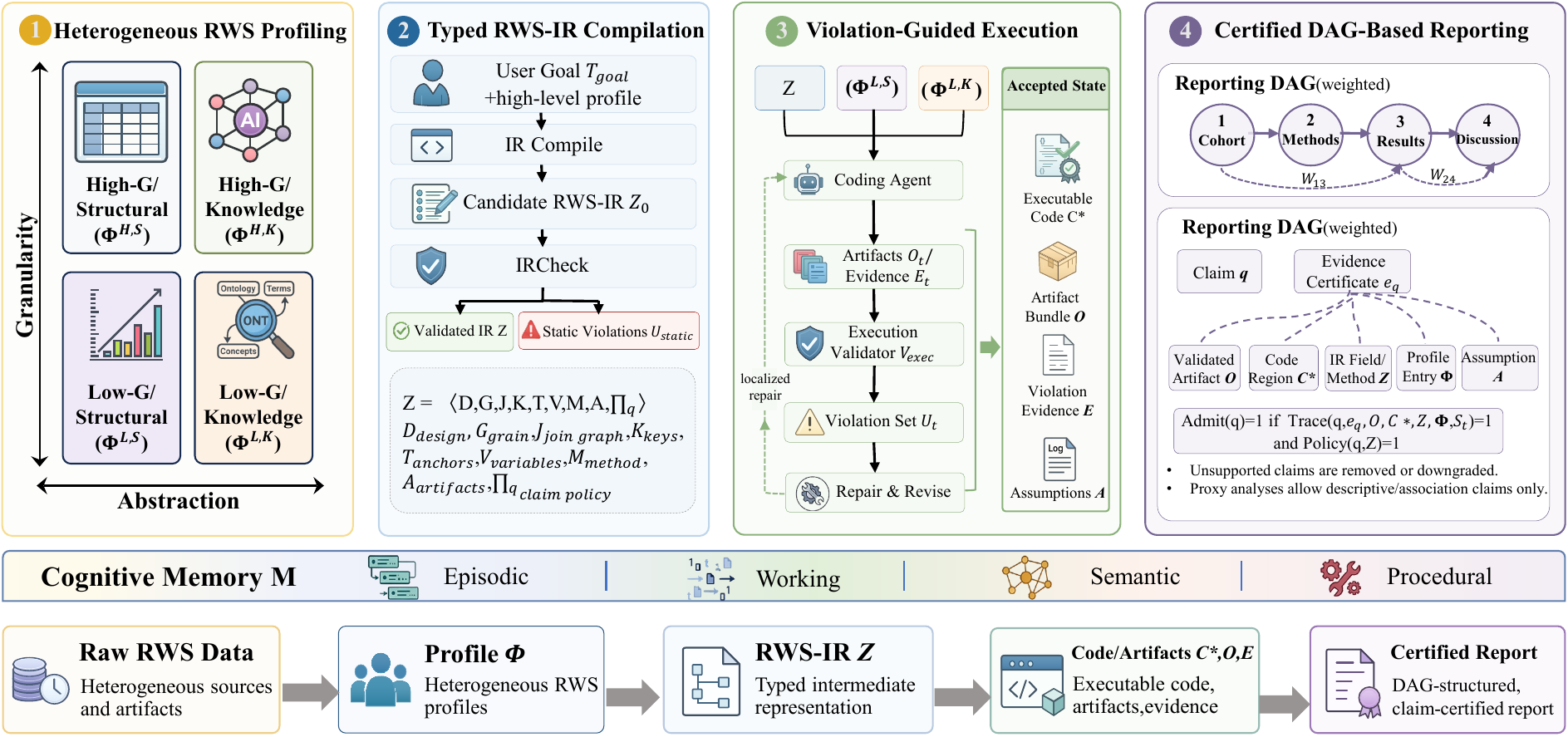}
    \caption{ProCARE architecture for contract-valid RWS automation. A heterogeneous RWS profile constrains RWS-IR compilation, violation-guided execution, and evidence-certified DAG reporting as a profile-grounded evidence contract.}
    \label{fig:framework}
\end{figure*}
ProCARE casts RWS automation as contract-valid generation over the three profile-constrained transitions in Figure~\ref{fig:framework}. Given a heterogeneous profile $\Phi$ and user goal $T_{\mathrm{goal}}$, it produces a typed study specification, validated code, result artifacts, a report, and admissible claims:
\begin{equation}
    \Phi, T_{\mathrm{goal}}
    \xrightarrow{\text{profile contract}}
    (Z, C^*, O, R, Q),
\end{equation}
where $Z$ is the RWS-IR, $C^*$ validated code, $O$ result and figure artifacts, $R$ the report, and $Q$ admitted claims. The contract requires every study choice, code operation, and claim to trace to profile entries, validated artifacts, or explicit assumptions.

ProCARE enforces this contract at three boundaries. Query-to-IR compilation admits only profile-supported study designs, variables, temporal anchors, and methods. IR-to-code execution accepts only runs that preserve the declared analysis grain, join graph, temporal ordering, and artifact requirements. Artifact-to-claim reporting admits only claims with evidence certificates linked to validated artifacts, executed code, IR fields, profile entries, or recorded assumptions.

The generation state is maintained as typed artifacts $\mathcal{S}_t=\{\Phi,Z_t,C_t,O_t,\mathcal{E}_t,\mathcal{U}_t,\mathcal{A}_t\}$. Here $\mathcal{E}_t$ stores validation evidence, $\mathcal{U}_t$ contract violations, and $\mathcal{A}_t$ assumptions and limitations.

\subsection{Heterogeneous Data Profiling}
\label{sec:profiling}
ProCARE first converts raw RWS inputs into a heterogeneous profile $\Phi=(\Phi^{\mathrm{H,S}}, \Phi^{\mathrm{H,K}}, \Phi^{\mathrm{L,S}}, \Phi^{\mathrm{L,K}})$, indexed by profile level (dataset/task vs.\ field/instance) and abstraction (structural vs.\ knowledge-driven). Table~\ref{tab:profile_matrix} summarizes the four blocks. Structural blocks specify joins, grain, temporal anchors, and artifact constraints; knowledge blocks record clinical roles, ontology mappings, proxy status, assumptions, and claim scope. Deterministic parsers extract structural profiles, while ontology checks over UMLS and SNOMED CT validate candidate clinical mappings for semantic type, units, ranges, and temporal-role compatibility. Failed or ambiguous mappings are marked as semantically unknown in $\Phi^{\mathrm{L,K}}$ with their failure causes.

Semantically unknown fields may support only structure-grounded operations: they cannot bind clinical roles, derive endpoints, or support claims beyond their validated level unless a proxy or explicit assumption is recorded. Thus, survival, safety, and utilization queries are compiled from validated entity levels, linkage paths, temporal anchors, clinical roles, and support levels. Unsupported choices are filtered, downgraded, or reported as limitations before IR compilation.

\begin{table}[t]
\centering
\scriptsize
\setlength{\tabcolsep}{4pt}
\renewcommand{\arraystretch}{0.95}

\caption{Heterogeneous data profile $\Phi$ organized by profile level and abstraction.}
\label{tab:profile_matrix}

\newcommand{\profilecell}[2]{%
\begin{tabular}[t]{@{}>{\raggedright\arraybackslash}p{2.7em}@{\hspace{0.3em}}%
>{\raggedright\arraybackslash}p{\dimexpr\linewidth-3.0em\relax}@{}}
#1 & #2
\end{tabular}%
}

\begin{tabularx}{\columnwidth}{@{}>{\bfseries}l X X@{}}
\toprule
& \textbf{High-level profile} \textit{(dataset/task)}
& \textbf{Low-level profile} \textit{(field/instance)} \\
\midrule

Structural &
\profilecell{$\Phi^{\mathrm{H,S}}$}
{Clinical form registry; subject/visit/table linkage; cohort-level missingness.}
&
\profilecell{$\Phi^{\mathrm{L,S}}$}
{Field type, unit, range, date fields, temporal anchors, and window-specific missingness.}
\\

Knowledge &
\profilecell{$\Phi^{\mathrm{H,K}}$}
{Cohort and index-date schema; exposure, outcome, censoring, confounder, and claim-scope roles.}
&
\profilecell{$\Phi^{\mathrm{L,K}}$}
{Ontology mapping; unit/value validation; temporal-role validation; field-level proxy and confounder roles.}
\\

\bottomrule
\end{tabularx}
\end{table}

\subsection{RWS-IR Compilation and Violation-Guided Execution}
\label{sec:planning-execution}

The IR and execution stages implement the first two contract transitions: query-to-IR and IR-to-code. The compiler turns the user goal into a typed RWS-IR, while the execution stage tests whether the generated analysis can be run without violating the profile, the IR, or the artifact policy.

\textbf{RWS-IR compilation.}
Given a research goal $T_{\mathrm{goal}}$, ProCARE first generates a candidate RWS-IR using the high-level profile:
\begin{equation}
    Z_0 =
    f_{\mathrm{IR}}
    \left(
        T_{\mathrm{goal}},
        \Phi^{\mathrm{H,S}} \oplus \Phi^{\mathrm{H,K}}
    \right),
\end{equation}
where $\oplus$ denotes structured fusion of profile blocks. The candidate IR is admitted only after static profile checking:
\begin{equation}
    (Z, \mathcal{U}_{\mathrm{static}})
    =
    \mathrm{IRCheck}_{\mathrm{static}}
    (Z_0, \Phi, \mathcal{S}_t).
\end{equation}

The accepted IR is $Z=\langle D,G,J,K,T,V,M,A,\Pi_q\rangle$, where $D$, $G$, $J$, $K$, $T$, $V$, $M$, $A$, and $\Pi_q$ denote study design, analysis grain, join graph, entity keys, temporal anchors, variable-role bindings, method specification, required artifacts, and claim policy, respectively.

Static checks enforce three conditions. First, all referenced tables, fields, and keys must appear in $\Phi^{\mathrm{L,S}}$. Second, clinical roles such as exposure, outcome, censoring, and confounder must be supported by $\Phi^{\mathrm{L,K}}$, marked as validated proxies, or explicitly recorded as assumptions. Third, the declared join graph, index date, follow-up window, and analysis grain must be compatible with the profiled schema. Only IRs with $\mathcal{U}_{\mathrm{static}}=\emptyset$ enter code generation. Otherwise, ProCARE repairs the IR, downgrades the unsupported request, or emits an infeasibility statement. Row-level temporal and grain checks are deferred to execution-time validation.

\noindent\textbf{Profile-constrained execution.}
Given an admitted IR, ProCARE generates executable analysis code inside a workspace containing $Z$, $\Phi^{\mathrm{L,S}}$, $\Phi^{\mathrm{L,K}}$, the claim policy $\Pi_q$, allowed tools, coding policies, artifact validators, and execution logs. A run is accepted only if it produces validated code $C^*$, complete output artifacts $O$, validation evidence $\mathcal{E}$, and disclosed assumptions or limitations $\mathcal{A}$.

Execution is controlled by deterministic gates. Schema gates require every table, field, join key, and output column to be traceable to $\Phi^{\mathrm{L,S}}$. Method gates require cohort filters, index dates, exposure definitions, endpoint or censoring rules, covariates, temporal windows, and statistical choices to match $Z$ and the validated semantic roles in $\Phi^{\mathrm{L,K}}$. Artifact gates require the selected entry point, such as \texttt{analysis.py} or \texttt{main.py}, to run successfully, avoid failure sentinels, and produce complete result and figure manifests.
After each execution attempt, validators return a violation set
$\mathcal{U}_t =
V_{\mathrm{exec}}(Z_t,C_t,O_t,\mathcal{E}_t,\Phi,\mathcal{S}_t)$.
Each violation records the failed contract clause, offending code or artifact, nearest valid profile entries, and permitted repair action. When $\mathcal{U}_t\neq\emptyset$, the next edit is conditioned on the localized failures,
$(Z_{t+1},C_{t+1})=
A_{\theta}(Z_t,C_t,\mathcal{U}_t,\mathcal{S}_t)$.
The violation set guides repair but does not imply global edit minimality.

If a requested endpoint, exposure, covariate, time window, or statistical analysis is unsupported by the profile, ProCARE terminates with a transparent failure trace or downgrades the task to a supported descriptive analysis with explicit limitations. Failed traces remain in $\mathcal{S}_t$ for inspection or subsequent runs. This makes profile constraints and methodological assumptions observable during code generation, rather than implicit in prompt context.

\subsection{Certified DAG-Based Reporting}
\label{sec:certified_reporting}

The reporting stage enforces the artifact-to-claim transition: only factual claims supported by validated evidence and allowed by the claim policy are admitted. For each claim $q$, the report generator emits an evidence certificate $e_q=\langle o_q,c_q,z_q,\phi_q,a_q\rangle$, binding the claim to a supporting artifact, executed code region, IR field or method, profile entry, and optional assumption or limitation. A claim is admitted only if both traceability and policy checks pass:
\begin{equation}
\mathrm{Admit}(q)=
\mathbb{I}\!\left[
\mathrm{Trace}(q,e_q,O,C^*,Z,\Phi,\mathcal{S}_t)=1
\land
\mathrm{Policy}(q,Z,\Pi_q)=1
\right].
\end{equation}

The trace condition grounds $q$ in validated artifacts, executed code, IR fields, profile entries, or disclosed assumptions; the policy condition checks whether its type and wording are allowed by $Z.\Pi_q$ at the available support level. Unsupported claims are removed or rewritten as limitations. Proxy-based or descriptive analyses may support descriptive or association claims, but not causal, safety-critical, or high-stakes claims without the required design assumptions and validations.

After claim filtering, a dependency DAG orders report sections: nodes are report components, and edges encode prerequisites such as cohort definition, variable construction, temporal anchors, sample support, and method-design consistency. The DAG controls section order, while evidence certificates control claim admissibility. The support score and traversal rule are provided in the appendix.

\section{Evaluation and Benchmark}
\label{sec:benchmark}

The benchmark evaluates complete RWS artifact bundles rather than isolated code outputs, covering study specification, code synthesis, visualization, and scientific reporting. Because ProCARE natively emits typed IRs, verifier traces, evidence certificates, and report-structure metadata, we distinguish artifact-level comparative metrics from contract-native diagnostics. Deterministic validators assess checkable constraints such as executability, schema use, temporal/grain consistency, artifact completeness, and claim support. Rubric-based judges are used only for residual semantic ambiguity, presentation quality, or stylistic criteria.

\noindent\textbf{Task construction.}
The primary benchmark contains 36 instances from four RWS domains: \textit{Cohort}, \textit{Safety}, \textit{Efficacy}, and \textit{Economics}. Each domain includes three research questions, and each question is instantiated under three prompt-specificity levels: \textsc{Goal}, \textsc{Goal+Approach}, and \textsc{Goal+Method}. All instances use the same longitudinal HCC cohort described below. Agents receive raw relational tables and natural-language prompts under a unified protocol.

\noindent\textbf{Output contract.}
Each prompt contains \textbf{Task}, optional \textbf{Guidance}, and \textbf{Deliverables}. The expected bundle includes: (i) a study specification, or equivalent intermediate description, covering tables, keys, variables, temporal anchors, methods, artifacts, and claim policy; (ii) an executable entry point, such as \texttt{analysis.py} or \texttt{main.py}, with artifacts; and (iii) a structured report with assumptions, limitations, and evidence links for factual claims. ProCARE instantiates this contract with typed RWS-IRs and evidence certificates; non-ProCARE systems are evaluated via submitted or post-hoc parsed specifications, manifests, and report-level evidence links. Hallucinated tables or columns are disallowed. \textsc{Goal+Method} must follow specified variables and terminate when required inputs are missing; \textsc{Goal} and \textsc{Goal+Approach} permit the closest feasible analysis with explicit limitations.

\noindent\textbf{Scoring protocol.}
Deterministic validators assess code executability, schema use, artifact completeness, temporal/grain consistency, and claim support. Study-specification validity is scored after mapping each system's submitted or parsed specification into the common benchmark schema; native ProCARE IR objects are not treated as ground truth. Residual semantic criteria are scored by Gemini-2.5-Pro and Qwen2.5-1M at temperature 0; judges are blinded to method identity and evaluate full bundles under fixed rubrics. The same validators, rubrics, and post-hoc diagnostics are applied to ProCARE, non-ProCARE baselines, and Public-RWS; Public-RWS omits repeated board-certified review. All scores are normalized to $[0,1]$, and automated diagnostics are compared with expert review on the primary benchmark.
Artifact-level metrics support cross-system comparisons. Contract-native metrics, such as explicit IR validity, native certificate coverage, and report-DAG consistency, are reported as mechanism diagnostics and are not interpreted alone as evidence of clinical, causal, or statistical correctness.

\begin{table}[t]
\centering
\footnotesize
\setlength{\tabcolsep}{4pt}
\renewcommand{\arraystretch}{0.96}
\caption{Evaluation metrics for end-to-end RWS automation.}
\label{tab:quality_metrics}
\begin{tabularx}{\columnwidth}{@{}>{\raggedright\arraybackslash}p{0.14\columnwidth}
>{\raggedright\arraybackslash}X@{}}
\toprule
\textbf{Dimension} & \textbf{Metrics and scoring} \\
\midrule
\textbf{IR} &
\textbf{GA}: LLM-rubric goal alignment;
\textbf{PF}: schema/IR feasibility with LLM adjudication. \\
\textbf{Execution} &
\textbf{XE}: sandbox pass rate $n_{\mathrm{pass}}/N$;
\textbf{CQ}: LLM-rubric code quality;
\textbf{VRR}: resolved/detected contract violations. \\
\textbf{Visualization} &
\textbf{VC}: LLM-rubric visual clarity;
\textbf{CF}: chart--result consistency. \\
\textbf{Reporting} &
\textbf{RS}: report schema and section coverage;
\textbf{SL}: LLM-rubric scientific language. \\
\textbf{Claims} &
\textbf{CTC}: post-hoc claim trace coverage;
\textbf{UCR}: unsupported factual-claim rate;
\textbf{CLV}: language-support violation rate. \\
\bottomrule
\end{tabularx}
\end{table}

Let $N_v,R_v$ be detected and resolved execution-time violations, and $N_q,C_q,U_q,L_q$ be factual, certified, unsupported, and language-violating claims. We compute
\begin{equation}
    \mathrm{VRR}=\frac{R_v}{N_v}, \qquad
    \mathrm{CTC}=\frac{C_q}{N_q}, \qquad
    \mathrm{UCR}=\frac{U_q}{N_q}, \qquad
    \mathrm{CLV}=\frac{L_q}{N_q}.
\end{equation}
VRR is measured over execution-time violations before the accepted run; CTC, UCR, and CLV are measured over final-report factual claims.
To reduce evaluator coupling, native contract artifacts are separated from artifact-level scoring where possible. ProCARE's RWS-IRs, manifests, verifier traces, and evidence certificates are treated as system outputs rather than ground truth. For non-ProCARE systems, the checker reconstructs the closest available study specification, artifact manifest, and claim-evidence links from code, logs, tables, figures, reports, profile entries, and recorded assumptions. CTC/UCR/CLV are then scored post hoc against executed artifacts, rather than by accepting native certificates. Native certificate and DAG diagnostics are retained only for mechanism audit. This mitigates, but does not eliminate, coupling because validity is still operationalized through profile-derived constraints.

\noindent\textbf{Primary HCC benchmark.}
The primary benchmark is derived from IRB-approved longitudinal hepatocellular carcinoma data collected at a tertiary medical center. The released version is a fully de-identified record-level benchmark: direct identifiers and fields that could identify individual participants are removed before publication, while random benchmark subject and visit identifiers preserve the relational structure needed for reproduction. The cohort contains 1{,}295 subjects, 12{,}847 visits, and 54 interlinked electronic case-report-form sheets covering baseline intake, follow-up, treatments, laboratory measurements, outcomes, and quality-of-life assessments. The dataset captures practical RWS challenges: subject- and visit-level linkage, sparse high-dimensional forms, form-specific clinical conventions, redundant administrative fields, and longitudinal structure for repeated-measures and time-to-event analyses. The public benchmark includes the de-identified HCC tables, task specifications, schema/profile artifacts, prompts, data dictionary, checksums, and evaluation scripts; a complete sheet inventory and release description are provided in the appendix.

\noindent\textbf{Public-RWS external validation suite.}
Public-RWS-18 evaluates cross-dataset transfer on three public health datasets: UCI Diabetes 130-US Hospitals, an EHR-derived encounter-level readmission dataset~\citep{uci_diabetes_2014,strack2014impact}; a NHANES 2017--2018 subset linking demographics, examination, questionnaire, laboratory, medical-condition, and prescription-medication tables by \texttt{SEQN}~\citep{cdc_nhanes_2017_2018}; and TCGA-LIHC, HCC-aligned oncology metadata with demographic, diagnosis, treatment, vital-status, and follow-up-related fields~\citep{gdc_tcga_lihc}. Each dataset contributes two task objectives under three prompt-specificity levels, yielding 18 external validation tasks. The suite tests schema shift, multi-table grounding, proxy-outcome handling, and claim traceability, complementing the primary HCC benchmark.

\section{Experiments}
\label{sec:experiments}
We evaluate ProCARE along five research questions: \textbf{RQ1} end-to-end artifact validity against LLM-agent and table-analysis baselines; \textbf{RQ2} contributions of individual contract mechanisms; \textbf{RQ3} claim-level traceability and agreement with a five-member clinical/statistical expert panel; \textbf{RQ4} robustness across RWS objectives, prompt specificity, and public-data schema shifts; and \textbf{RQ5} sensitivity to the base LLM backbone and profiling cost amortization. The evaluation spans 36 primary HCC RWS tasks, 18 Public-RWS external validation tasks, baseline comparisons, ablations, claim diagnostics, expert review, backbone sensitivity, and cost analysis.

\subsection{Main Results}
\label{subsec:main_results}

\begin{table*}[t]
\centering
\scriptsize
\setlength{\tabcolsep}{3.2pt}
\renewcommand{\arraystretch}{1.06}
\caption{Baseline comparison on 36 HCC RWS tasks. GA--SL report task performance, while VRR--CLV report contract diagnostics following Table~\ref{tab:quality_metrics}; N/A indicates no executable artifact bundle. \textsuperscript{\textdagger}Proprietary; \textsuperscript{\textdaggerdbl}Open-source; \textsuperscript{\textsection}Agent-based.}
\label{tab:baseline_comparison}
\begin{tabularx}{\textwidth}{@{}>{\raggedright\arraybackslash}p{0.15\textwidth}
*{8}{>{\centering\arraybackslash}p{0.052\textwidth}}
*{4}{>{\centering\arraybackslash}p{0.055\textwidth}}@{}}
\toprule
\textbf{Method}
& \textbf{GA}$\uparrow$
& \textbf{PF}$\uparrow$
& \textbf{XE}$\uparrow$
& \textbf{CQ}$\uparrow$
& \textbf{VC}$\uparrow$
& \textbf{CF}$\uparrow$
& \textbf{RS}$\uparrow$
& \textbf{SL}$\uparrow$
& \textbf{VRR}$\uparrow$
& \textbf{CTC}$\uparrow$
& \textbf{UCR}$\downarrow$
& \textbf{CLV}$\downarrow$ \\
\midrule
ReAct$^{\ddagger\S}$              & 0.78 & 0.62 & 0.38 & 0.47 & 0.28 & 0.48 & 0.63 & 0.59 & 0.34 & 0.48 & 0.24 & 0.17 \\
ToolLLM$^{\ddagger\S}$            & 0.73 & 0.68 & 0.00 & N/A  & N/A  & N/A  & 0.56 & 0.24 & 0.00 & 0.21 & 0.33 & 0.24 \\
TableGPT-Agent$^{\ddagger\S}$     & 0.61 & 0.59 & N/A  & N/A  & N/A  & N/A  & 0.40 & 0.43 & N/A  & 0.18 & 0.36 & 0.22 \\
ChatExcel-Max$^{\dagger\S}$       & N/A  & N/A  & N/A  & N/A  & N/A  & N/A  & 0.52 & 0.27 & N/A  & 0.12 & 0.39 & 0.28 \\
Gemini-3-Pro direct$^{\dagger}$   & 0.75 & 0.51 & 0.13 & 0.45 & 0.28 & 0.48 & 0.61 & 0.53 & 0.18 & 0.36 & 0.27 & 0.15 \\
TableGPT2$^{\ddagger}$            & 0.41 & 0.45 & 0.00 & 0.25 & N/A  & N/A  & N/A  & N/A  & 0.00 & N/A  & N/A  & N/A  \\
Claude Code$^{\dagger\S}$         & 0.82 & 0.75 & 0.86 & 0.80 & 0.52 & 0.58 & 0.71 & 0.66 & 0.70 & 0.66 & 0.16 & 0.12 \\
Codex$^{\dagger\S}$               & 0.83 & 0.77 & 0.89 & 0.82 & 0.55 & 0.61 & 0.73 & 0.68 & 0.73 & 0.69 & 0.14 & 0.10 \\
\midrule
\multicolumn{13}{l}{\textbf{Methods without valid artifact bundles:} LLM (DeepSeek-R1-0528); ChatExcel-Pro$^{\dagger\S}$.} \\
\midrule
\textbf{ProCARE (Ours)}
& \textbf{0.86}
& \textbf{0.84}
& \textbf{0.92}
& \textbf{0.88}
& \textbf{0.72}
& \textbf{0.74}
& \textbf{0.80}
& \textbf{0.79}
& \textbf{0.81}
& \textbf{0.90}
& \textbf{0.06}
& \textbf{0.03} \\
\bottomrule
\end{tabularx}
\end{table*}

\textbf{Baseline comparison (RQ1).}
\label{sec:baseline_comparison}
We first compare ProCARE with LLM-agent, table-analysis, and code-agent baselines under matched prompts, inference budgets, and artifact contracts. Table~\ref{tab:baseline_comparison} shows that existing methods often either fail to produce executable artifact bundles or emit factual claims unsupported by admissible artifacts, code, or claim policies. Claude Code and Codex substantially improve executability and code quality over generic agents, but remain weaker than ProCARE on clinically grounded profile fit, variable-role consistency, statistical language, and claim traceability. To partially control for evaluator bias toward systems that natively emit IRs, manifests, and evidence certificates, we additionally evaluate scaffolded baselines. These RQ1 controls keep the tested system as an external baseline or direct Gemini-3-Pro run, while only adding evaluator-facing information, output schemas, validator feedback, and repair budget. They do not expose ProCARE's typed IR compiler, certificate admission gate, certified reporting DAG, or internal planner state.
The scaffolds improve generic baselines, especially in executability and artifact completeness. ProCARE still achieves the strongest violation resolution and post-hoc claim traceability (Table~\ref{tab:scaffolded_baselines}), suggesting that its advantage is not explained by schema exposure or manifest formatting alone.

\begin{table*}[t]
\centering
\captionsetup{skip=2pt}

\begin{threeparttable}
\caption{Scaffolded RQ1 controls. Direct rows use Gemini-3-Pro with the same decoding settings as the default ProCARE backbone. Non-ProCARE rows are independent baseline adaptations; ProCARE modules are not transplanted into them.}
\label{tab:scaffolded_baselines}

{\scriptsize
\setlength{\tabcolsep}{2.5pt}
\renewcommand{\arraystretch}{0.92}

\begin{tabularx}{\textwidth}{@{}p{0.20\textwidth}Y *{7}{N}@{}}
\toprule
\textbf{Baseline} & \textbf{Fairness scaffold} &
{\textbf{PF}$\uparrow$} & {\textbf{XE}$\uparrow$} & {\textbf{CQ}$\uparrow$} & {\textbf{VRR}$\uparrow$} &
{\textbf{CTC}$\uparrow$} & {\textbf{UCR}$\downarrow$} & {\textbf{CLV}$\downarrow$} \\
\midrule
ReAct + scaffold
& $\Phi$ summary + bundle tmpl. + val. feedback + 3 repairs
& 0.79 & 0.74 & 0.69 & 0.62 & 0.70 & 0.13 & 0.10 \\

ToolLLM + scaffold
& Same scaffold on the tool-use baseline
& 0.80 & 0.68 & 0.65 & 0.58 & 0.66 & 0.15 & 0.12 \\

G3P + profile/template
& $\Phi$ + manifests; no validator loop
& 0.73 & 0.66 & 0.62 & 0.31 & 0.55 & 0.20 & 0.15 \\

G3P + validator repair
& $\Phi$ + manifests + val. feedback + repairs
& 0.78 & 0.79 & 0.72 & 0.68 & 0.65 & 0.15 & 0.11 \\

G3P + study-spec schema
& Emits study-spec artifact; no IR compiler
& 0.82 & 0.70 & 0.68 & 0.43 & 0.52 & 0.21 & 0.16 \\

G3P + claim-trace schema
& Emits claim-trace artifact; no cert. gate
& 0.74 & 0.63 & 0.61 & 0.35 & 0.74 & 0.11 & 0.08 \\

\textbf{ProCARE reference}
& \textbf{Profile--IR--validator--certificate--DAG workflow}
& \bfseries 0.84 & \bfseries 0.92 & \bfseries 0.88
& \bfseries 0.81 & \bfseries 0.90 & \bfseries 0.06
& \bfseries 0.03 \\
\bottomrule
\end{tabularx}
}

\end{threeparttable}
\end{table*}

\textbf{Contract mechanism ablation (RQ2).}
\label{subsec:ablation}
To identify the sources of improvement, we ablate each contract object in Table~\ref{tab:ablation}. The profile yields the largest contribution, suggesting that reliable schema, field, and join constraints are the basis for valid execution and downstream reasoning. RWS-IR supplies the typed study contract and strengthens claim-policy control. Violation-guided repair mainly improves executable constraint satisfaction, whereas evidence certificates and certified DAG ordering primarily affect reporting: the former enforces claim traceability, and the latter preserves dependency-aware structure. These results indicate that the framework’s gains come from complementary contract objects: structural contracts support computation, repair improves execution robustness, and evidence/order contracts regulate faithful reporting.

\begin{table*}[t]
\centering
\caption{Mechanism ablation over 36 RWS instances. N/A indicates no typed RWS-IR.}
\label{tab:ablation}
\footnotesize
\setlength{\tabcolsep}{3.8pt}
\renewcommand{\arraystretch}{1.04}
\begin{tabularx}{\textwidth}{@{}lcccccc>{\raggedright\arraybackslash}X@{}}
\toprule
\textbf{Variant} &
\textbf{PF}$\uparrow$ &
\textbf{XE}$\uparrow$ &
\textbf{VRR}$\uparrow$ &
\textbf{CTC}$\uparrow$ &
\textbf{UCR}$\downarrow$ &
\textbf{SL}$\uparrow$ &
\textbf{Dominant failure} \\
\midrule
No profile $\Phi$
& 0.49 & 0.18 & 0.35 & 0.58 & 0.28 & 0.42
& Field hallucination; invalid joins. \\

No RWS-IR
& N/A & 0.34 & 0.44 & 0.63 & 0.22 & 0.69
& Missing grain, anchors, and claim policy. \\

No repair
& 0.84 & 0.69 & 0.43 & 0.80 & 0.13 & 0.70
& Unresolved runtime and artifact violations. \\

No certificates
& 0.84 & 0.92 & 0.81 & N/A & 0.26 & 0.62
& Unsupported causal or comparative claims. \\

No DAG ordering
& 0.84 & 0.92 & 0.81 & 0.84 & 0.11 & 0.64
& Dependency-agnostic report ordering. \\
\midrule
\textbf{Full ProCARE}
& \textbf{0.84} & \textbf{0.92} & \textbf{0.81}
& \textbf{0.90} & \textbf{0.06} & \textbf{0.79}
& Localized repair; traceable final claims. \\
\bottomrule
\end{tabularx}
\end{table*}

\textbf{Claim traceability and expert validation (RQ3).}
\label{subsec:claim_traceability}
Table~\ref{tab:claim_traceability} reports post-hoc claim diagnostics for ProCARE. More specific prompts improve CTC and reduce UCR/CLV, mainly by constraining variable bindings, temporal anchors, and methods. Remaining downgrades mostly involve causal or high-stakes wording unsupported by proxy artifacts, sparse endpoints, or subgroup evidence.
\begin{table}[t]
\centering
\small
\setlength{\tabcolsep}{4.5pt}
\renewcommand{\arraystretch}{1.08}
\caption{ProCARE claim diagnostics. Values are averaged over 12 instances per prompt level; Overall aggregates all 36 instances.}
\label{tab:claim_traceability}

\begin{tabularx}{\linewidth}{@{}l
    S[mode=text,detect-family=true,detect-weight=true,table-format=1.1]
    S[mode=text,detect-family=true,detect-weight=true,table-format=1.2]
    S[mode=text,detect-family=true,detect-weight=true,table-format=1.2]
    S[mode=text,detect-family=true,detect-weight=true,table-format=1.2]
    >{\raggedright\arraybackslash}X@{}}
\toprule
\textbf{Prompt} &
{\textbf{Claims}} &
{\textbf{CTC}$\uparrow$} &
{\textbf{UCR}$\downarrow$} &
{\textbf{CLV}$\downarrow$} &
\textbf{Main downgrade reason} \\
\midrule
\textsc{Goal}
& 5.8 & 0.87 & 0.09 & 0.05
& Unsupported causal/comparative wording. \\

\textsc{Goal+Approach}
& 6.4 & 0.90 & 0.05 & 0.03
& Proxy endpoint beyond evidence level. \\

\textsc{Goal+Method}
& 6.9 & 0.94 & 0.03 & 0.02
& Sparse endpoint or subgroup evidence. \\
\midrule
\textbf{Overall}
& {\bfseries 6.4} & {\bfseries 0.90} & {\bfseries 0.06} & {\bfseries 0.03}
& Downgrades mainly concern causal strength. \\
\bottomrule
\end{tabularx}
\end{table}
We further compare automated diagnostics with blinded review by three clinical and two statistics/data experts. Table~\ref{tab:human_ai} shows moderate-to-strong Spearman agreement across stages, strongest for RWS-IR validity ($\rho=0.82$) and significant for code, visualization, and reporting. The appendix provides reviewer rubrics, aggregation rules, confidence intervals, and inter-rater reliability.

\begin{table}[!t]
\centering
\small
\setlength{\tabcolsep}{3.6pt}
\caption{Expert consistency of automated metrics. Expert scores are mean$\pm$SD on a 5-point scale; CI and reliability details are reported in the appendix.}
\label{tab:human_ai}
\begin{tabularx}{\linewidth}{@{}l l >{\centering\arraybackslash}X c c c@{}}
\toprule
\textbf{Stage} & \textbf{Reviewers} & \textbf{Metric(s)} & \textbf{Auto} & \textbf{Expert} & \textbf{$\rho$} \\
\midrule
RWS-IR        & 3 clinical + 2 stats/data & $(GA+PF)/2$ & 0.82 & 4.38$\pm$0.48 & 0.82 \\
Code          & 2 stats/data              & $CQ$        & 0.88 & 4.42$\pm$0.21 & 0.74 \\
Visualization & 3 clinical + 2 stats/data & $(VC+CF)/2$ & 0.66 & 4.50$\pm$0.53 & 0.75 \\
Reporting     & 3 clinical + 2 stats/data & $SL$        & 0.76 & 4.13$\pm$0.64 & 0.71 \\
\bottomrule
\multicolumn{6}{@{}l}{\footnotesize All Spearman correlations are computed across the 36 primary HCC instances; all $p<0.001$.}
\end{tabularx}
\end{table}

\subsection{Robustness and Deployment Analyses}
\label{subsec:robustness_deployment}

\begin{figure}[!t]
    \centering
    \includegraphics[width=1\linewidth]{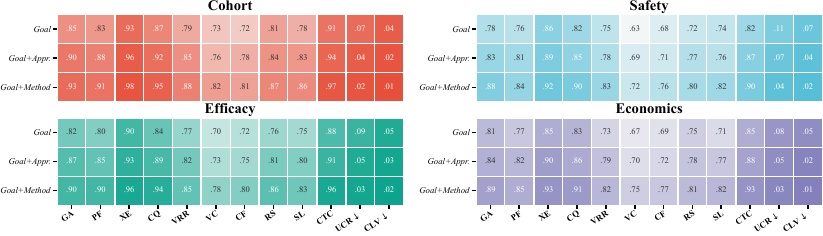}
    \caption{Performance across four RWS tasks and prompt-specificity levels. Cell labels report raw metric values; UCR and CLV are lower-is-better rates with reversed color intensity.}
    \label{fig:cross_task}
\end{figure}

\textbf{Task, prompt, and public-data robustness (RQ4).}
\label{subsec:cross_task}
Figure~\ref{fig:cross_task} evaluates ProCARE across four RWS objectives and three prompt-specificity levels on the HCC benchmark. Metrics generally improve with prompt specificity, suggesting that methodological context helps specification, repair, and claim control. Performance is stable across task families; residual variation is mainly due to sparse endpoints, limited subgroup support, and claim-scope restrictions.
Table~\ref{tab:public_rws_results} evaluates Public-RWS-18 schema transfer without dataset-specific changes. Claim policies restrict outputs to descriptive/readmission-proxy claims for UCI Diabetes, descriptive/association claims for NHANES, and descriptive oncology-metadata/follow-up-proxy claims for TCGA-LIHC. ProCARE outperforms Claude Code and Scaffolded G3P in violation resolution, post-hoc claim traceability, UCR, and CLV; NHANES remains hardest due to multi-table survey joins.

\begin{table*}[t]
\centering
\small
\setlength{\tabcolsep}{6pt}
\renewcommand{\arraystretch}{1.08}
\caption{Public-RWS-18 transfer results. Values average six tasks per dataset; baseline claim traces are computed post hoc.}
\label{tab:public_rws_results}
\begin{tabularx}{\textwidth}{@{}>{\raggedright\arraybackslash}p{0.17\textwidth}
>{\raggedright\arraybackslash}p{0.20\textwidth}
*{5}{>{\centering\arraybackslash}X}@{}}
\toprule
\textbf{Dataset} & \textbf{Method} & \textbf{XE}$\uparrow$ & \textbf{VRR}$\uparrow$ & \textbf{CTC}$\uparrow$ & \textbf{UCR}$\downarrow$ & \textbf{CLV}$\downarrow$ \\
\midrule
\textbf{UCI Diabetes} 
 & Claude Code & 0.60 & 0.61 & 0.69 & 0.12 & 0.09 \\
 & Scaffolded G3P & 0.57 & 0.59 & 0.66 & 0.13 & 0.10 \\
 & \textbf{ProCARE} & \textbf{0.64} & \textbf{0.72} & \textbf{0.87} & \textbf{0.07} & \textbf{0.04} \\
\textbf{NHANES} 
 & Claude Code & 0.52 & 0.54 & 0.63 & 0.15 & 0.11 \\
 & Scaffolded G3P & 0.50 & 0.52 & 0.60 & 0.16 & 0.12 \\
 & \textbf{ProCARE} & \textbf{0.58} & \textbf{0.66} & \textbf{0.84} & \textbf{0.09} & \textbf{0.06} \\
\textbf{TCGA-LIHC} 
 & Claude Code & 0.57 & 0.58 & 0.66 & 0.14 & 0.10 \\
 & Scaffolded G3P & 0.54 & 0.56 & 0.63 & 0.15 & 0.11 \\
 & \textbf{ProCARE} & \textbf{0.61} & \textbf{0.69} & \textbf{0.86} & \textbf{0.08} & \textbf{0.05} \\
\bottomrule
\end{tabularx}
\end{table*}

\textbf{Backbone sensitivity and cost (RQ5).}
\label{subsec:cross_model}
Holding the profile, RWS-IR, validators, and certificate gate fixed, ProCARE remained contract-consistent across tested backbones; stronger planners mainly improved GA, PF, VRR, and CTC, while execution metrics were comparatively stable. Profiling is a one-time dataset cost ($T_{\rm prof}\approx65$k tokens for the HCC schema); per-query inference costs about $T_{\rm inf}\approx38$k tokens, giving an amortized cost $\bar T(N)=T_{\rm inf}+T_{\rm prof}/N$ and $\bar T(20)\approx41.3$k tokens. Full backbone results are reported in Appendix.

\begin{table*}[t]
\centering
\scriptsize
\setlength{\tabcolsep}{2.0pt}
\renewcommand{\arraystretch}{1.06}
\caption[Evaluation results across base models]{Backbone sensitivity with fixed ProCARE workflow. Metrics follow Table~\ref{tab:quality_metrics}; model access details and version identifiers are reported in the appendix. 
\textsuperscript{\textdagger}Proprietary models; \textsuperscript{\textdaggerdbl}Open-source models.}
\label{tab:cross_model}
\begin{tabularx}{\textwidth}{@{}>{\bfseries\raggedright\arraybackslash}p{0.080\textwidth}
>{\raggedright\arraybackslash}X
*{8}{>{\centering\arraybackslash}p{0.045\textwidth}}
*{4}{>{\centering\arraybackslash}p{0.052\textwidth}}@{}}
\toprule
\textbf{Family} & \textbf{Model} & \multicolumn{8}{c}{\textbf{Performance}} & \multicolumn{4}{c}{\textbf{Contract diagnostics}} \\
\cmidrule(lr){3-10} \cmidrule(lr){11-14}
& & \textbf{GA}$\uparrow$ & \textbf{PF}$\uparrow$ & \textbf{XE}$\uparrow$ & \textbf{CQ}$\uparrow$ 
& \textbf{VC}$\uparrow$ & \textbf{CF}$\uparrow$ & \textbf{RS}$\uparrow$ & \textbf{SL}$\uparrow$
& \textbf{VRR}$\uparrow$ & \textbf{CTC}$\uparrow$ & \textbf{UCR}$\downarrow$ & \textbf{CLV}$\downarrow$ \\
\midrule

\multirow{2}{*}{GPT} 
& GPT-5.2\textsuperscript{\textdagger} 
& 0.84 & 0.82 & \underline{0.94} & 0.84 & 0.68 & 0.70 & \underline{0.78} & \underline{0.77} & 0.78 & \underline{0.88} & \textbf{0.06} & \underline{0.04} \\
& GPT-4 Turbo\textsuperscript{\textdagger} 
& 0.81 & 0.78 & 0.89 & 0.80 & 0.64 & 0.66 & 0.75 & 0.74 & 0.74 & 0.85 & 0.08 & 0.06 \\
\midrule

\multirow{2}{*}{Claude}
& Claude Opus 4.5\textsuperscript{\textdagger} 
& \underline{0.85} & \underline{0.83} & \textbf{0.97} & \underline{0.86} & \textbf{0.72} & \textbf{0.74} & 0.78 & 0.77 & \underline{0.80} & 0.89 & \textbf{0.06} & \underline{0.04} \\
& Claude Sonnet 4.5\textsuperscript{\textdagger} 
& 0.83 & 0.81 & \underline{0.94} & 0.84 & \underline{0.70} & \underline{0.72} & 0.76 & 0.75 & 0.78 & 0.87 & 0.07 & 0.05 \\
\midrule

\multirow{2}{*}{Gemini}
& Gemini-3-Pro\textsuperscript{\textdagger} 
& \textbf{0.86} & \textbf{0.84} & 0.92 & \textbf{0.88} & \textbf{0.72} & \textbf{0.74} & \textbf{0.80} & \textbf{0.79} & \textbf{0.81} & \textbf{0.90} & \textbf{0.06} & \textbf{0.03} \\
& Gemini-2.5-Pro\textsuperscript{\textdagger} 
& 0.82 & 0.78 & 0.91 & 0.82 & 0.66 & 0.68 & 0.75 & 0.74 & 0.76 & 0.86 & 0.08 & 0.06 \\
\midrule

\multirow{2}{*}{Qwen}
& Qwen3-Max\textsuperscript{\textdagger} 
& 0.77 & 0.70 & 0.84 & 0.76 & 0.58 & 0.60 & 0.68 & 0.68 & 0.68 & 0.80 & 0.12 & 0.09 \\
& Qwen3-235B-Instruct\textsuperscript{\textdaggerdbl}
& 0.74 & 0.66 & 0.80 & 0.72 & 0.55 & 0.57 & 0.64 & 0.65 & 0.64 & 0.76 & 0.15 & 0.11 \\
\midrule

\multirow{2}{*}{DeepSeek}
& DeepSeek-V3.2\textsuperscript{\textdaggerdbl} 
& 0.76 & 0.68 & 0.86 & 0.78 & 0.60 & 0.62 & 0.66 & 0.67 & 0.70 & 0.80 & 0.11 & 0.08 \\
& DeepSeek-R1-0528\textsuperscript{\textdaggerdbl} 
& 0.72 & 0.64 & 0.82 & 0.74 & 0.56 & 0.58 & 0.62 & 0.63 & 0.66 & 0.77 & 0.13 & 0.10 \\
\bottomrule
\end{tabularx}
\end{table*}

Token budgets are not directly comparable across agents because each baseline uses a different pipeline, so we report token use for \textsc{ProCARE}. Profiling is a one-time dataset cost, about $T_{\mathrm{prof}}{\approx}65$k tokens for the rich schema; each query then uses about $T_{\mathrm{inf}}{\approx}38$k tokens. For $N$ queries on the same dataset, the amortized cost is $\bar{T}(N)=T_{\mathrm{inf}}+T_{\mathrm{prof}}/N$. A single query costs $\bar{T}(1){\approx}103$k tokens; with $N{=}20$, the cost falls to $\bar{T}(20){\approx}41.3$k tokens, close to the inference-only cost.

\textbf{Discussion and Limitations.}
Overall, ProCARE should be interpreted as a contract layer for making RWS assumptions, artifacts, and claim support auditable under human oversight. Its reliability depends on the correctness of the constructed profile, the coverage of deterministic validators, and the assumptions encoded in the RWS-IR. A successful run shows that the generated bundle satisfies the implemented schema, method, temporal, and artifact checks; it does not establish clinical optimality, adequate confounding control, causal validity, or evidentiary sufficiency beyond the recorded assumptions. The primary benchmark is single-center and disease-specific, and Public-RWS-18 mainly tests schema transfer rather than multi-institutional longitudinal generalization. Deployments on OMOP, FHIR, or institutional schemas therefore require re-instantiating the profile and revalidating the IR, execution gates, and certificate checks.

\section{Conclusion}
\label{sec:conclusion}

We introduced ProCARE, a profile-grounded contract architecture for RWS automation. ProCARE compiles study goals into typed RWS-IR, generates executable artifact bundles through a profile-constrained coding harness, repairs localized contract violations, and admits report claims through evidence certificates. On a 36-instance HCC RWS benchmark and Public-RWS-18 external validation suite, these mechanisms improve executable validity, code quality, and claim traceability relative to LLM-agent and table-analysis baselines. The public release includes the fully de-identified HCC benchmark tables and scoring assets to support independent reproduction. Future work should test the same contract mechanisms on multi-institutional longitudinal RWS data and standardized OMOP/FHIR deployments.

\bibliographystyle{unsrtnat}
\bibliography{neurips_2026}

\clearpage
\appendix
\newcommand{\MapCoverageOverall}{0.83}      %
\newcommand{\MapCoverageCritical}{0.91}     %
\newcommand{\NoConceptRate}{0.07}           %
\newcommand{\TypeIncompatRate}{0.05}        %
\newcommand{\AmbiguityRate}{0.03}           %
\newcommand{\OtherFailRate}{0.02}           %

\newcommand{\AlphaSupport}{0.0}            %
\newcommand{\BetaSupport}{3.0}             %
\newcommand{\WVar}{0.35}                    %
\newcommand{\WTemp}{0.25}                   %
\newcommand{\WSize}{0.25}                   %
\newcommand{\WCons}{0.15}                   %
\newcommand{\NMin}{30}                    %

\section{Appendix Roadmap, Navigation, and Main-Paper Coverage}
\label{app:roadmap}

The appendix contains both claim-support material for the main paper and standalone audit material that is not fully described in the main text. Table~\ref{tab:appendix_navigation} gives a complete section-level map of the supplement. Table~\ref{tab:appendix_crosswalk} then maps main-paper claims to their most relevant supporting material.

\begin{table*}[htbp]
\centering
\scriptsize
\setlength{\tabcolsep}{4pt}
\renewcommand{\arraystretch}{1.08}
\caption{Complete appendix navigation. The map covers all supplement sections, including audit and interface material that is not tied to a specific main-paper table or figure.}
\label{tab:appendix_navigation}
\begin{tabularx}{\textwidth}{@{}p{0.16\textwidth}p{0.34\textwidth}X@{}}
\toprule
\textbf{Appendix section} & \textbf{Contents} & \textbf{Purpose in the supplement} \\
\midrule
App.~\ref{app:roadmap} & Roadmap, complete navigation, and main-paper crosswalk. & Orients readers to both the full supplement and the subset of material that directly supports main-paper claims. \\
App.~\ref{app:benchmark_data} & HCC benchmark scope, release boundary, task construction, Public-RWS-18, claim policies, and worksheet inventory. & Documents the datasets, task instances, and schema constraints needed to reproduce or audit the benchmark. \\
App.~\ref{app:method_contract} & Profile schema, concrete profile instance, RWS-IR checks, evidence certificates, validators, artifact bundle, coding harness, and DAG reporting. & Specifies the executable contract objects and implementation checks behind ProCARE. \\
App.~\ref{app:experimental_protocol} & Execution environment, cost accounting, baseline adaptation, scaffolded fairness controls, ablations, model access, and repository commands. & Records experimental controls and reproducibility details. \\
App.~\ref{app:additional_results} & Claim-diagnostic interpretation and backbone-sensitivity interpretation. & Explains claim-policy downgrades and performance differences across backbones. \\
App.~\ref{app:system_interface} & Dashboard, profile panel, RWS-IR view, coding loop, report review, and section-wise report generation screenshots. & Supplies workflow/interface evidence showing where artifacts, traces, logs, and evidence-linked report sections are inspected. \\
App.~\ref{app:rubrics} & Metric definitions, deterministic validator rules, LLM judge protocol, expert-review protocol, raw expert-data schema, reliability, and rubric anchors. & Defines how automated and human evaluation scores are produced and audited. \\
App.~\ref{app:prompts} & Benchmark, IR compiler, coding harness, report/certificate, and judge prompt templates. & Makes the generation and scoring prompts inspectable. \\
App.~\ref{app:responsible_release} & Data release, privacy, responsible-use caveats, and deployment limits. & Clarifies the de-identified HCC release, public-data sources, and how the system should be interpreted. \\
\bottomrule
\end{tabularx}
\end{table*}

\begin{table*}[htbp]
\centering
\scriptsize
\setlength{\tabcolsep}{4pt}
\renewcommand{\arraystretch}{1.08}
\caption{Main-paper claims and corresponding supplementary support. This crosswalk is not the full appendix map; complete section navigation is provided in Table~\ref{tab:appendix_navigation}.}
\label{tab:appendix_crosswalk}
\begin{tabularx}{\textwidth}{@{}p{0.20\textwidth}p{0.36\textwidth}X@{}}
\toprule
\textbf{Main-paper location} & \textbf{Claim or reference} & \textbf{Supplementary support} \\
\midrule
Sec.~\mainref{sec:profiling} & The heterogeneous profile records structural and knowledge-driven constraints, with a concrete serialization and de-identified profile instance. & App.~\ref{app:profile_schema} and App.~\ref{app:phi_instance}. \\
Sec.~\mainref{sec:planning-execution} & RWS-IR compilation and execution enforce schema, semantic-role, temporal, grain, method, and artifact constraints. & App.~\ref{app:ir_static_checks}, App.~\ref{app:execution_validators}, and App.~\ref{app:coding-harness}. \\
Sec.~\mainref{sec:certified_reporting} & Evidence certificates and claim policies gate factual report claims, while a DAG orders supported report sections. & App.~\ref{app:claim_admission}, App.~\ref{app:artifact_bundle}, and App.~\ref{app:scoring}. \\
Sec.~\mainref{sec:benchmark} & The primary benchmark contains 36 HCC tasks over four objectives and a 54-sheet longitudinal eCRF workbook. & App.~\ref{app:benchmark_data} and Table~\ref{tab:sheet-inventory}. \\
Sec.~\mainref{sec:benchmark}--\mainref{sec:experiments} & Public-RWS-18 uses three public datasets, dataset-specific claim policies, and the same validators and rubrics. & App.~\ref{app:public_rws}. \\
Main Table~\mainref{tab:baseline_comparison} & LLM-agent, table-analysis, and code-agent baselines are evaluated under a unified prompt, budget, artifact contract, and post-hoc checker. & App.~\ref{app:baselines}. \\
Main Table~\mainref{tab:ablation} & Ablations remove one contract mechanism at a time while keeping the rest of the workflow fixed when defined. & App.~\ref{app:ablation_protocol}. \\
Main Table~\mainref{tab:claim_traceability} & Claim diagnostics are computed over factual report claims with trace and policy checks. & App.~\ref{app:claim_metrics}. \\
Main Table~\mainref{tab:human_ai} & Automated metrics are compared with expert review. & App.~\ref{app:expert_review}. \\
Main Figure~\mainref{fig:cross_task} & Stage-wise performance varies by objective and prompt specificity. & App.~\ref{app:benchmark_data} and App.~\ref{app:rubrics}. \\
Main Table~\mainref{tab:cross_model} and Sec.~\mainref{subsec:robustness_deployment} & Backbone sensitivity uses fixed workflow components; model metadata and amortized token cost are reported. & App.~\ref{app:model_access} and App.~\ref{app:cost_accounting}. \\
\bottomrule
\end{tabularx}
\end{table*}

\section{Benchmark Datasets and Tasks}
\label{app:benchmark_data}
\label{app:dataset}

This section gives the benchmark details not repeated in the main paper: release contents, worksheet organization, task instantiation, public-data transfer tasks, and the worksheet inventory used by the validators.

\subsection{Supplemental schema scope}
The main paper summarizes the cohort size and high-level eCRF scope. Here we record the schema properties needed to reproduce profiling and task construction. The workbook is multi-table by design, spanning enrollment, baseline characterization, time-varying interventions, follow-up, and survival outcomes; this structure is what drives the join, grain, temporal-anchor, and claim-policy checks used throughout the benchmark.

\subsection{Release boundary and de-identification}
\label{app:hcc_release_boundary}
The public benchmark release contains the fully de-identified HCC record-level tables used in the 36 primary tasks, together with task specifications, prompt files, schema and profile artifacts, evaluation scripts, aggregate profile examples, checksums, and a data dictionary. Before release, direct identifiers and fields that could identify individual participants are removed from the benchmark copy. Random benchmark subject and visit identifiers are retained only to preserve within-dataset linkage, visit order, and longitudinal analysis structure. The released tables are intended for reproducibility of the benchmark and do not constitute a clinical deployment dataset.

\begin{table*}[t]
\centering
\small
\setlength{\tabcolsep}{4pt}
\renewcommand{\arraystretch}{1.08}
\caption{HCC benchmark release contents and privacy treatment.}
\label{tab:hcc_release_contents}
\begin{tabularx}{\textwidth}{@{}p{0.22\textwidth}p{0.26\textwidth}X@{}}
\toprule
\textbf{Release item} & \textbf{Included material} & \textbf{Privacy and reproducibility treatment} \\
\midrule
De-identified HCC tables & 54 linked eCRF sheets covering the 1{,}294-subject benchmark cohort. & Direct participant identifiers are removed; random benchmark IDs preserve subject-level and visit-level linkage required for rerunning all tasks. \\
Task and prompt files & The 36 primary HCC benchmark prompts and task metadata. & Prompts refer only to de-identified table and field names and do not expose participant identities. \\
Schema/profile artifacts & Field dictionaries, join keys, temporal roles, missingness summaries, and profile objects. & Contains schema and aggregate metadata used by validators; no direct identity fields are present. \\
Evaluation assets & Validators, judge prompts, artifact templates, scoring scripts, and expected output schema. & Enables independent reruns against the released de-identified tables. \\
Release metadata & Data dictionary, checksums, and responsible-use terms. & Documents provenance, file integrity, de-identification scope, and restrictions against re-identification or clinical use. \\
\bottomrule
\end{tabularx}
\end{table*}

\subsection{Functional groupings}
Table~\ref{tab:functional-groups} groups worksheets by their role in profiling and benchmark construction.

\begin{table*}[htbp]
\centering
\scriptsize
\setlength{\tabcolsep}{4pt}
\renewcommand{\arraystretch}{1.08}
\caption{Functional worksheet groups used by profiling and benchmark construction.}
\label{tab:functional-groups}
\begin{tabularx}{\textwidth}{@{}l l X X@{}}
\toprule
\textbf{Group} & \textbf{Representative sheets} & \textbf{Main contents} & \textbf{Use in benchmark/profile} \\
\midrule
System metadata & SYSVAR, ECRF, DATA\_RANGE & Form catalog, site allocation, variable descriptions, and codebook fields. & Schema parsing, worksheet indexing, and field-name grounding. \\
Baseline & DM, PH, CP, BCLC, ECOG & Demographics, history, liver function, staging, and performance status. & Cohort definition, baseline table construction, and confounder checks. \\
Longitudinal labs & LB\_AFP, LB\_BB, LB\_BR, LB\_TT & Tumor markers, biochemistry, complete blood count, and coagulation tests. & Safety proxy signals and follow-up burden measures. \\
Treatment & TACE, TART, IMU, PATE, CM & Procedures, systemic therapy, retreatment, and concomitant medication records. & Exposure definition, treatment intensity, and resource-utilization tasks. \\
Outcomes & VT, AE\_HQ, SAE\_HQ, SF & Recurrence and survival proxies, AE/SAE screening, and quality-of-life forms. & Endpoint derivation and limitation checks. \\
\bottomrule
\end{tabularx}
\end{table*}

\subsection{Dataset limitations for benchmark design}
Several data properties directly shape the benchmark tasks. Adverse-event screening pages are available, but detailed AE/SAE records are sparse in the released workbook; safety tasks therefore distinguish screening-level summaries from detailed event-term analyses. The dataset does not contain direct billing or price fields, so economics-oriented tasks are framed as resource utilization, follow-up burden, or quality-of-life proxy analyses rather than cost-effectiveness studies.

\subsection{Benchmark task construction}
The benchmark crosses four RWS objectives, three core questions per objective, and three prompt-specificity levels:
\[
|\mathcal{B}|=|\mathcal{O}|\times|\mathcal{Q}|\times|\mathcal{L}|=4\times3\times3=36.
\]
Table~\ref{tab:prompt-levels} defines the prompt levels used across all objectives.

For each metric in Main Figure~\mainref{fig:cross_task}, an objective--prompt cell averages the three benchmark instances for that objective at the specified prompt level. A fixed prompt level therefore contains twelve instances across the four objectives; the full benchmark contains all 36 instances across the three levels.

\begin{table}[htbp]
\centering
\small
\setlength{\tabcolsep}{4pt}
\renewcommand{\arraystretch}{1.08}
\caption{Prompt specificity levels in the benchmark.}
\label{tab:prompt-levels}
\begin{tabularx}{\linewidth}{@{}l X X@{}}
\toprule
\textbf{Level} & \textbf{Information provided} & \textbf{Expected behavior} \\
\midrule
GOAL & High-level study intent only. & Select feasible variables and disclose assumptions. \\
GOAL+APPROACH & Study intent plus an analysis strategy. & Follow the strategy when the data support it; otherwise report the deviation. \\
GOAL+METHOD & Specific variables, time windows, or statistical method. & Stop or report infeasibility when required data are missing. \\
\bottomrule
\end{tabularx}
\end{table}

All instances use the same output contract: a verification plan, executable analysis entry point, generated artifacts, and a structured RWS report. Table~\ref{tab:core-questions} lists the twelve core questions, and Tables~\ref{tab:cohort-instances}--\ref{tab:econ-instances} give the corresponding 36 prompts.

\begin{table*}[htbp]
\centering
\scriptsize
\setlength{\tabcolsep}{3pt}
\renewcommand{\arraystretch}{1.08}
\caption{Core research questions used to instantiate benchmark prompts.}
\label{tab:core-questions}
\begin{tabularx}{\textwidth}{@{}p{0.12\textwidth}p{0.07\textwidth}X p{0.30\textwidth}@{}}
\toprule
\textbf{Objective} & \textbf{ID} & \textbf{Core research question} & \textbf{Primary data sources} \\
\midrule
Cohort & C1 & Characterize the analyzable HCC cohort, including enrollment, baseline characteristics, and clinical staging. & \texttt{CO}, \texttt{DM}, \texttt{IE}, \texttt{SG}, \texttt{BCLC}, \texttt{CP}, \texttt{ECOG} \\
Cohort & C2 & Assess longitudinal follow-up completeness and data density. & \texttt{CPH\_FT}, \texttt{VT}, \texttt{LB\_AFP}, \texttt{LB\_BB}, \texttt{LB\_BR}, \texttt{ECOG} \\
Cohort & C3 & Assess data readiness for downstream endpoint analyses. & \texttt{DM}, \texttt{SG}, \texttt{BCLC}, \texttt{CP}, \texttt{ECOG}, \texttt{VT}, treatment sheets \\
Safety & S1 & Summarize AE/SAE screening rates and the availability of detailed event records. & \texttt{AE\_HQ}, \texttt{SAE\_HQ}, \texttt{AE}, \texttt{SAE} \\
Safety & S2 & Describe postoperative laboratory safety signals from liver function, hematology, and coagulation panels. & \texttt{LB\_BB}, \texttt{LB\_BR}, \texttt{LB\_TT}, \texttt{SG} \\
Safety & S3 & Compare safety proxy signals after major treatment exposures. & \texttt{TACE}, \texttt{TART}, \texttt{IMU}, \texttt{PATE}, \texttt{AE\_HQ}, \texttt{SAE\_HQ}, lab sheets \\
Efficacy & E1 & Analyze recurrence and recurrence-free survival proxies after surgery. & \texttt{SG}, \texttt{VT}, \texttt{BCLC}, \texttt{CP}, \texttt{ECOG}, treatment sheets \\
Efficacy & E2 & Analyze overall survival proxies and their relationship to treatment exposure. & \texttt{SG}, \texttt{VT}, \texttt{DM}, \texttt{BCLC}, \texttt{CP}, \texttt{ECOG}, treatment sheets \\
Efficacy & E3 & Analyze response assessment and longitudinal tumor-marker changes. & \texttt{VT}, \texttt{LB\_AFP}, \texttt{AFP2}, \texttt{BCLC}, treatment sheets \\
Health economics/resource use & H1 & Describe postoperative treatment resource utilization and treatment intensity. & \texttt{TACE}, \texttt{TART}, \texttt{IMU}, \texttt{PATE}, \texttt{CM}, \texttt{SG} \\
Health economics/resource use & H2 & Estimate follow-up and testing burden as a resource-use proxy. & \texttt{CPH\_FT}, \texttt{VT}, \texttt{LB\_AFP}, \texttt{LB\_BB}, \texttt{LB\_BR}, \texttt{LB\_TT}, \texttt{ECOG}, \texttt{SF} \\
Health economics/resource use & H3 & Use SF-12 availability and treatment intensity to assess QoL/resource-use feasibility. & \texttt{SF}, \texttt{TACE}, \texttt{TART}, \texttt{IMU}, \texttt{PATE}, \texttt{VT} \\
\bottomrule
\end{tabularx}
\end{table*}

\begin{table*}[htbp]
\centering
\scriptsize
\setlength{\tabcolsep}{3pt}
\renewcommand{\arraystretch}{1.08}
\caption{Cohort benchmark instances. Suffixes denote prompt specificity: G = GOAL, A = GOAL+APPROACH, and M = GOAL+METHOD.}
\label{tab:cohort-instances}
\begin{tabularx}{\textwidth}{@{}p{0.07\textwidth}X X@{}}
\toprule
\textbf{ID} & \textbf{Prompt text} & \textbf{Required or expected evidence} \\
\midrule
C1-G & Characterize the baseline hepatocellular carcinoma cohort available in the workbook. Deliver a cohort summary suitable for a real-world study report. & Discover \texttt{DM}, \texttt{SG}, \texttt{BCLC}, \texttt{CP}, \texttt{ECOG}, and subject identifiers. \\
C1-A & Characterize the baseline HCC cohort. Build a STROBE-style cohort table covering demographics, surgery, liver function, tumor stage, and performance status. & Use patient-level baseline records, report missingness, and avoid duplicate subject counting. \\
C1-M & Produce a baseline cohort table for subjects with valid \texttt{Subject\_Id}. Join \texttt{DM}, \texttt{SG}, \texttt{BCLC}, \texttt{CP}, and baseline \texttt{ECOG}; summarize \texttt{AGE}, \texttt{SEX}, \texttt{HEIGHT}, \texttt{WEIGHT}, \texttt{SGDAT}, \texttt{SGTYP}, \texttt{BCLC}, \texttt{CPRES}, \texttt{CPGRA}, and \texttt{ECOG}. & Use one baseline row per subject; report variable-level missingness; do not impute incomplete \texttt{AGE}. \\
C2-G & Assess whether the dataset has sufficient longitudinal follow-up for real-world evidence generation. & Identify visit-level forms and repeated measurements. \\
C2-A & Assess longitudinal follow-up completeness across follow-up forms, laboratory panels, tumor-marker tests, ECOG assessments, and efficacy follow-up records. & Use \texttt{CPH\_FT}, \texttt{VT}, laboratory sheets, and \texttt{ECOG}; produce visit-level summaries. \\
C2-M & Quantify longitudinal data completeness using \texttt{CPH\_FT}, \texttt{VT}, \texttt{LB\_AFP}, \texttt{LB\_BB}, \texttt{LB\_BR}, and \texttt{ECOG}; count distinct visits by \texttt{Subject\_Id}, tabulate availability by \texttt{Visit\_No}, and estimate follow-up duration from \texttt{SG.SGDAT} to the latest \texttt{VT.VTDAT1} or \texttt{VT.VTDAT2}. & Use \texttt{Subject\_Id} and \texttt{Visit\_No}; report subjects with no valid follow-up date. \\
C3-G & Evaluate whether the cohort is fit for downstream RWS endpoint analyses. & Reason about exposure, outcome, confounder, and time-anchor variables. \\
C3-A & Evaluate endpoint-analysis readiness through a feasibility matrix covering identifiers, surgery date, treatment exposure, recurrence/death endpoints, and baseline confounders. & Include data limitations, not only descriptive counts. \\
C3-M & Build a data-readiness matrix by checking \texttt{Subject\_Id}, \texttt{SG.SGDAT}, recurrence dates \texttt{VT.PADHDAT}/\texttt{VT.PADRDAT}, death dates \texttt{VT.DEDAT1}/\texttt{VT.DEDAT2}, treatments from \texttt{TACE}, \texttt{TART}, \texttt{IMU}, \texttt{PATE}, and confounders \texttt{DM.SEX}, \texttt{DM.AGE}, \texttt{BCLC.BCLC}, \texttt{CP.CPGRA}, \texttt{ECOG.ECOG}. & Flag sparse or unavailable fields, especially \texttt{VT.DEDAT2}; do not invent endpoint fields. \\
\bottomrule
\end{tabularx}
\end{table*}

\begin{table*}[htbp]
\centering
\scriptsize
\setlength{\tabcolsep}{3pt}
\renewcommand{\arraystretch}{1.08}
\caption{Safety benchmark instances.}
\label{tab:safety-instances}
\begin{tabularx}{\textwidth}{@{}p{0.07\textwidth}X X@{}}
\toprule
\textbf{ID} & \textbf{Prompt text} & \textbf{Required or expected evidence} \\
\midrule
S1-G & Analyze the adverse-event and serious-adverse-event safety profile in the cohort. & Discover that AE/SAE screening pages exist but detailed event tables are empty in the current workbook. \\
S1-A & Analyze AE and SAE safety reporting. Start with screening pages, then check whether detailed AE/SAE records are available for term, onset date, severity, causality, outcome, and seriousness. & Document infeasibility of detailed AE/SAE term analysis when detail tables have no valid records. \\
S1-M & Summarize AE/SAE reporting completeness using \texttt{AE\_HQ.AEYN} and \texttt{SAE\_HQ.SAEYN}; inspect \texttt{AE.AETERM}, \texttt{AE.AESTDAT}, \texttt{AE.AESEV}, \texttt{AE.AEREL}, \texttt{AE.AESAE}, \texttt{SAE.SAESTDAT}, and \texttt{SAE.SAESEV}. & Compute screening prevalence by subject; terminate detail-level analysis if \texttt{AE} and \texttt{SAE} contain zero valid records. \\
S2-G & Identify laboratory safety signals after surgery. & Use longitudinal laboratory sheets rather than relying only on AE tables. \\
S2-A & Identify post-surgical laboratory safety signals by summarizing liver function, hematology, and coagulation markers over follow-up visits. & Use lab value/result-judgment pairs and visit dates; report abnormal-result flags when available. \\
S2-M & Generate a post-surgical laboratory safety profile anchored at \texttt{SG.SGDAT}; use \texttt{LB\_BB.LBBBDAT}, \texttt{ALTRES}, \texttt{ASTRES}, \texttt{TBILRES}, \texttt{ALBRES}; \texttt{LB\_BR.BRDAT}, \texttt{WBCRES}, \texttt{HBRES}, \texttt{PLTRES}; and \texttt{LB\_TT.LBTTDAT}, \texttt{PTRES}, \texttt{INRRES}, \texttt{DDIMRES}. & Summarize baseline and postoperative windows; use available-case denominators and \texttt{\_SIG} flags when numeric thresholds are unavailable. \\
S3-G & Compare safety patterns across major postoperative treatment exposures. & Identify TACE, targeted therapy, immunotherapy, and retreatment exposure sheets. \\
S3-A & Compare safety proxy signals by treatment exposure using TACE, targeted therapy, immunotherapy, and postoperative retreatment groups. & Compare AE/SAE screening and laboratory abnormalities descriptively; avoid causal claims. \\
S3-M & Compare treatment-associated safety proxies using exposure indicators from \texttt{TACE.TACEDAT}, \texttt{TART.TARSDAT}, \texttt{IMU.IMUSDAT}, and \texttt{PATE.PATEYN}; calculate screening rates from \texttt{AE\_HQ.AEYN} and \texttt{SAE\_HQ.SAEYN}; summarize post-exposure laboratory signals from \texttt{LB\_BB}, \texttt{LB\_BR}, and \texttt{LB\_TT}. & Report confounder imbalance using \texttt{BCLC.BCLC}, \texttt{CP.CPGRA}, and \texttt{ECOG.ECOG}; frame results as descriptive safety proxies. \\
\bottomrule
\end{tabularx}
\end{table*}

\begin{table*}[htbp]
\centering
\scriptsize
\setlength{\tabcolsep}{3pt}
\renewcommand{\arraystretch}{1.08}
\caption{Efficacy benchmark instances.}
\label{tab:efficacy-instances}
\begin{tabularx}{\textwidth}{@{}p{0.07\textwidth}X X@{}}
\toprule
\textbf{ID} & \textbf{Prompt text} & \textbf{Required or expected evidence} \\
\midrule
E1-G & Analyze recurrence-related outcomes after surgery. & Locate surgery dates and recurrence dates in \texttt{VT}. \\
E1-A & Analyze recurrence-free outcomes using surgery date as time origin, recurrence events from follow-up records, and clinically relevant baseline risk strata. & Handle multiple follow-up visits and competing or missing dates. \\
E1-M & Estimate recurrence-free survival proxies using \texttt{SG.SGDAT} as index date; define recurrence as the earliest non-missing \texttt{VT.PADHDAT} or \texttt{VT.PADRDAT}; censor at latest \texttt{VT.VTDAT1} or \texttt{VT.VTDAT2}; stratify by \texttt{BCLC.BCLC}, \texttt{CP.CPGRA}, and exposure from \texttt{TACE}, \texttt{TART}, \texttt{IMU}, or \texttt{PATE}. & Use earliest valid recurrence date and valid time ordering; generate Kaplan--Meier curves only when dates are sufficient. \\
E2-G & Analyze overall survival patterns in the cohort. & Identify death dates and follow-up dates. \\
E2-A & Analyze overall survival by deriving time from surgery to death or last follow-up and comparing clinically meaningful strata. & Use \texttt{DEDAT1}/\texttt{DEDAT2} and \texttt{VT} follow-up dates; disclose death-date limitations. \\
E2-M & Estimate overall-survival proxies using \texttt{SG.SGDAT} as index date; define death date as earliest non-missing \texttt{VT.DEDAT1} or \texttt{VT.DEDAT2}; censor at latest valid \texttt{VT.VTDAT1} or \texttt{VT.VTDAT2}; compare treatment exposures and adjust or stratify descriptively by \texttt{DM.SEX}, \texttt{DM.AGE}, \texttt{BCLC.BCLC}, \texttt{CP.CPGRA}, and \texttt{ECOG.ECOG} when available. & Note that \texttt{DEDAT2} is unavailable in the current data; avoid over-adjusted models when events are limited. \\
E3-G & Evaluate treatment response and tumor-marker dynamics over follow-up. & Combine efficacy assessment and biomarker sheets. \\
E3-A & Evaluate response and tumor-marker dynamics by summarizing hospital/telephone efficacy evaluations and longitudinal AFP, CA199, and DCP trajectories by visit and exposure. & Combine categorical response information with biomarker trends. \\
E3-M & Analyze response and tumor-marker trends using \texttt{VT.CEVYN}, \texttt{VT.CUEV}, \texttt{VT.VTDAT1}, \texttt{VT.VTDAT2}, plus \texttt{LB\_AFP.LBAFPDAT}, \texttt{AFPRES}, \texttt{CA199RES}, and \texttt{DCPRES}; align measurements by \texttt{Subject\_Id} and date relative to \texttt{SG.SGDAT}; summarize trajectories by exposure to \texttt{TACE}, \texttt{TART}, or \texttt{IMU}. & If formal response criteria are not encoded, report observed categories and marker changes without RECIST claims. \\
\bottomrule
\end{tabularx}
\end{table*}

\begin{table*}[htbp]
\centering
\scriptsize
\setlength{\tabcolsep}{3pt}
\renewcommand{\arraystretch}{1.08}
\caption{Health economics and resource-use benchmark instances.}
\label{tab:econ-instances}
\begin{tabularx}{\textwidth}{@{}p{0.07\textwidth}X X@{}}
\toprule
\textbf{ID} & \textbf{Prompt text} & \textbf{Required or expected evidence} \\
\midrule
H1-G & Analyze postoperative treatment resource utilization in the cohort. & Recognize that the task concerns utilization rather than direct cost. \\
H1-A & Analyze postoperative treatment resource utilization by quantifying use and timing of TACE, targeted therapy, immunotherapy, postoperative retreatment, and concomitant medications. & Use procedure and drug counts with dates as utilization proxies. \\
H1-M & Quantify treatment-utilization intensity by deriving per-subject counts and time-to-first-use for \texttt{TACE.TACEDAT}, \texttt{TART.TARSDAT}, \texttt{IMU.IMUSDAT}, \texttt{PATE.PATEYN}/\texttt{PATETR*}, and \texttt{CM.CMCAT}/\texttt{CMSTDAT}; summarize utilization distributions and stratify by \texttt{BCLC.BCLC} and recurrence status when feasible. & Explicitly state that no direct cost or billing variables are present. \\
H2-G & Estimate follow-up and testing burden associated with the RWS cohort. & Use visit and form/test counts as burden proxies. \\
H2-A & Estimate follow-up and testing burden by counting follow-up visits, efficacy assessments, laboratory panels, tumor-marker checks, coagulation tests, ECOG assessments, and QoL forms per patient. & Summarize burden by patient and visit. \\
H2-M & Build a follow-up resource-burden profile by counting records in \texttt{CPH\_FT}, \texttt{VT}, \texttt{LB\_AFP}, \texttt{LB\_BB}, \texttt{LB\_BR}, \texttt{LB\_TT}, \texttt{ECOG}, and \texttt{SF}; compute total follow-up forms/tests, median counts per subject, and distribution by \texttt{Visit\_No}; use \texttt{SG.SGDAT} and latest \texttt{VT.VTDAT1}/\texttt{VT.VTDAT2} to contextualize burden over time. & Frame the result as utilization burden, not economic cost. \\
H3-G & Explore quality-of-life evidence and its relationship to treatment burden. & Discover SF-12 sparsity and avoid overclaiming. \\
H3-A & Explore QoL and treatment burden using SF-12 items when available; summarize missingness and compare available QoL responses across treatment-utilization groups. & Emphasize sparse SF item completeness. \\
H3-M & Analyze QoL/resource-use proxy feasibility by inspecting \texttt{SF.SFYN} and \texttt{SF1}--\texttt{SF12}; compute complete-item availability; when enough records exist, summarize item-level or total scores by exposure groups from \texttt{TACE}, \texttt{TART}, \texttt{IMU}, and \texttt{PATE}; otherwise report feasibility and missingness only. & Terminate formal QoL comparison when usable SF-12 item records are too sparse. \\
\bottomrule
\end{tabularx}
\end{table*}

\subsection{Public-RWS external validation suite}
\label{app:public_rws}

Public-RWS-18 is an external validation suite for testing whether the same profile--IR--certificate pipeline transfers to public health datasets with different schemas and study designs. It is not merged with the 36-instance HCC benchmark and is not treated as another full longitudinal RWS benchmark. Each public dataset contributes two task objectives, and each objective is evaluated under low-, medium-, and high-specificity prompts.

\begin{table*}[t]
\centering
\small
\setlength{\tabcolsep}{4pt}
\renewcommand{\arraystretch}{1.08}
\caption{Public-RWS external validation datasets. These datasets test cross-dataset transfer and contract diagnostics, not additional full HCC-style longitudinal RWS validity.}
\label{tab:public_rws_datasets}
\begin{tabularx}{\textwidth}{@{}p{0.18\textwidth}p{0.24\textwidth}p{0.30\textwidth}X@{}}
\toprule
\textbf{Dataset} & \textbf{Structure} & \textbf{Task focus} & \textbf{Boundary} \\
\midrule
UCI Diabetes 130-US Hospitals & 101{,}766 encounters, 50 columns; the department-split workbook is used only as a sheet-navigation variant. & Readmission proxy, medication process, utilization, coded variables, missingness. & Encounter-level data with limited longitudinal structure. \\
NHANES 2017--2018 subset & 10 linked XPT/CSV tables with 3{,}036--19{,}643 rows per table, linked by \texttt{SEQN}. & Survey cohort construction, labs, medications, conditions, demographic covariates. & Cross-sectional survey data, not EHR. \\
TCGA-LIHC clinical metadata & 377 cases, 137 clinical metadata columns. & HCC metadata, vital status, treatment fields, follow-up proxy fields. & Cancer registry-like clinical metadata, not routine-care EHR. \\
\bottomrule
\end{tabularx}
\end{table*}

\begin{table*}[t]
\centering
\small
\setlength{\tabcolsep}{4pt}
\renewcommand{\arraystretch}{1.08}
\caption{Task objectives in Public-RWS-18. Each objective is evaluated under low-, medium-, and high-specificity prompts.}
\label{tab:public_rws_tasks}
\begin{tabularx}{\textwidth}{@{}p{0.18\textwidth}p{0.30\textwidth}X@{}}
\toprule
\textbf{Dataset} & \textbf{Objective} & \textbf{Contract stressor} \\
\midrule
UCI Diabetes & Readmission proxy profiling & Outcome binding, coded variables, missingness, conservative claim scope. \\
UCI Diabetes & Medication and utilization analysis & Exposure/proxy roles, subgroup aggregation, artifact traceability. \\
NHANES & Diabetes/cardiometabolic cohort analysis & Multi-table joins by \texttt{SEQN}, lab-defined variables, survey-table grounding. \\
NHANES & Medication and condition association analysis & Questionnaire, laboratory, and prescription-medication grounding under a cross-sectional claim policy. \\
TCGA-LIHC & HCC case and vital-status profiling & Domain transfer, oncology metadata grounding, missingness disclosure. \\
TCGA-LIHC & Treatment/follow-up metadata analysis & Sparse fields, proxy endpoint handling, limitation reporting. \\
\bottomrule
\end{tabularx}
\end{table*}

\begin{table*}[t]
\centering
\small
\setlength{\tabcolsep}{4pt}
\renewcommand{\arraystretch}{1.08}
\caption{Dataset-specific claim policies for Public-RWS-18.}
\label{tab:public_rws_claim_policy}
\begin{tabularx}{\textwidth}{@{}>{\raggedright\arraybackslash}p{0.18\textwidth}X X@{}}
\toprule
\textbf{Dataset} & \textbf{Allowed claims} & \textbf{Forbidden claims} \\
\midrule
UCI Diabetes & Descriptive and readmission-proxy association claims. & Causal treatment effect, safety conclusion, longitudinal endpoint claim. \\
NHANES & Descriptive and cross-sectional association claims. & Temporal effect, incidence claim, causal effect. \\
TCGA-LIHC & Descriptive oncology metadata and follow-up-proxy claims. & Treatment efficacy, causal survival effect, routine-care EHR claim. \\
\bottomrule
\end{tabularx}
\end{table*}

Public-RWS-18 uses the same profile construction, RWS-IR schema, validators, and claim checker as the HCC benchmark. Dataset-specific preprocessing is limited to loading the public tables, normalizing file and column names for stable tool access, preserving original public codebook variables, and constructing profile metadata. No public dataset receives a custom validator, claim checker, or reporting rule. The claim policy in Table~\ref{tab:public_rws_claim_policy} is the only dataset-specific adaptation, because each public source supports a different study-design boundary.

Each public dataset contributes two objectives, and each objective is instantiated under three prompt-specificity levels. Dataset-level results are reported once in the main paper. Public-RWS-18 uses ReAct as a generic agent baseline and adds two stronger controls aligned with the primary baseline table: Claude Code as a native code-agent baseline, and Scaffolded G3P as the direct Gemini-3-Pro profile/template plus validator-repair control. Scaffolded G3P receives evaluator-facing schema information, artifact templates, validator feedback, and the same repair budget, but not ProCARE's IR compiler, certificate admission gate, certified reporting DAG, or internal planner state. This appendix records the dataset boundaries, task objectives, claim policies, and prompt text needed to rerun the transfer suite. For baselines without explicit evidence certificates, CTC, UCR, and CLV are computed by post-hoc evidence reconstruction over submitted artifacts and reports.

\begin{lstlisting}[basicstyle=\ttfamily\footnotesize,breaklines=true,breakatwhitespace=true,columns=fullflexible,caption={Prompt set for Public-RWS-18.},label={lst:public-rws-prompts}]
UCI Diabetes -- Objective U1: Readmission proxy profiling
Low: Analyze the diabetes hospital encounter data and identify which patient groups appear to have higher readmission burden.
Medium: Using the readmitted field as a proxy outcome, summarize early-readmission patterns by age, admission type, discharge disposition, and prior utilization variables.
High: Using readmitted = "<30" as the early-readmission proxy outcome, compute subgroup-level rates by age band, admission_type_id, discharge_disposition_id, number_inpatient, and number_emergency. Produce a table, one figure, and a report restricted to descriptive or proxy-association claims with evidence certificates for all factual statements.

UCI Diabetes -- Objective U2: Medication and utilization analysis
Low: Analyze whether medication and hospital-utilization patterns differ across diabetes encounters.
Medium: Summarize insulin, change, diabetesMed, A1Cresult, time_in_hospital, num_lab_procedures, num_medications, and num_procedures, and describe any supported association with readmitted.
High: Compute utilization summaries for time_in_hospital, num_lab_procedures, num_medications, and num_procedures stratified by insulin, change, and diabetesMed. Compare early-readmission proxy rates across medication-process groups, and avoid causal treatment-effect language.

NHANES -- Objective N1: Diabetes/cardiometabolic cohort analysis
Low: Use the NHANES tables to profile diabetes and cardiometabolic risk patterns in the population sample.
Medium: Join demographics, examination, blood pressure, diabetes questionnaire, laboratory, and medical-condition tables by SEQN. Summarize diabetes-related variables and cardiometabolic measurements by age and sex.
High: Construct a cross-sectional NHANES cohort using SEQN-linked tables. Define diabetes-related status using supported questionnaire or laboratory fields, summarize blood pressure and relevant laboratory measures by demographic subgroup, and report only descriptive or cross-sectional association claims.

NHANES -- Objective N2: Medication and condition association analysis
Low: Analyze how prescription medication records relate to self-reported medical conditions in NHANES.
Medium: Join prescription medication, medical conditions, demographics, and laboratory tables by SEQN. Summarize medication exposure groups and compare supported condition or laboratory profiles.
High: Using SEQN-linked medication and condition tables, construct medication-exposure groups when supported by the codebooks, summarize condition prevalence and laboratory measurements, and produce a report that explicitly states cross-sectional design limitations.

TCGA-LIHC -- Objective T1: HCC case and vital-status profiling
Low: Profile the TCGA-LIHC clinical metadata and summarize major patient, diagnosis, and vital-status patterns.
Medium: Summarize demographics, diagnosis-related fields, vital status, and available follow-up-related fields in TCGA-LIHC. Identify missingness and fields that can support survival-proxy reporting.
High: Using TCGA-LIHC clinical metadata, produce a case-profile table covering age, sex, diagnosis-related fields, vital status, days-to-death or follow-up-related fields when available, and missingness. Report only descriptive oncology-metadata claims with evidence certificates.

TCGA-LIHC -- Objective T2: Treatment/follow-up metadata analysis
Low: Analyze whether available treatment and follow-up metadata in TCGA-LIHC support subgroup comparisons.
Medium: Identify treatment-related, diagnosis-related, and follow-up-related fields in TCGA-LIHC, assess their completeness, and perform only comparisons supported by valid profile entries.
High: Using TCGA-LIHC clinical metadata, identify valid treatment-related and follow-up-related fields, summarize completeness by subgroup, and generate a conservative report that distinguishes observed metadata patterns from treatment-effect or causal claims.
\end{lstlisting}

\subsection{Worksheet inventory}

\begingroup
\small
\setlength{\tabcolsep}{4pt}
\renewcommand{\arraystretch}{1.0}
\begin{longtable}{@{}>{\ttfamily}p{0.13\textwidth} r r p{0.36\textwidth} p{0.29\textwidth}@{}}
\caption{Standardized worksheet inventory in the Excel dataset.}
\label{tab:sheet-inventory}\\
\toprule
\textbf{Prefix} & \textbf{\#Fields} & \textbf{\#Rows} & \textbf{Sheet Description} & \textbf{Differential field examples} \\
\midrule
\endfirsthead
\caption[]{Standardized worksheet inventory in the Excel dataset (continued).}\\
\toprule
\textbf{Prefix} & \textbf{\#Fields} & \textbf{\#Rows} & \textbf{Sheet Description} & \textbf{Differential field examples} \\
\midrule
\endhead
\bottomrule
\endfoot
SYSVAR      & 2   & 27    & System variables and metadata. & \texttt{Variable}, \texttt{Description} \\
DATA\_RANGE & 2   & 4     & Data range / center mapping and allocation notes. & \texttt{Center}, \texttt{RangeCode} \\
ECRF        & 8   & 54    & eCRF form catalog and design attributes. & \texttt{FormCode}, \texttt{FormName}, \texttt{FormType} \\
CPH\_FT     & 22  & 11614 & Header/footer records for visit-/form-level. & \texttt{VISTNM}, \texttt{SITNO}, \texttt{SUBNO} \\
CO          & 21  & 1296  & Study cover page and subject identification fields. & \texttt{SITENO}, \texttt{SUBJNO}, \texttt{SUBINI} \\
DS          & 19  & 1296  & Informed consent date and related tracking fields. & \texttt{DSDAT} \\
DM          & 27  & 1295  & Demographics and baseline characteristics. & \texttt{SEX}, \texttt{BRTHDAT}, \texttt{AGE} \\
PH          & 123 & 1262  & Medical history and prior conditions. & \texttt{PHSMYN}, \texttt{SMNUM}, \texttt{SMYER} \\
PH1         & 123 & 1     & Medical history (continuation / overflow record). & \texttt{PH1SMYN}, \texttt{SMNUM1}, \texttt{SMYER1} \\
HBCV        & 49  & 4282  & HBV/HCV-related laboratory assessments. & \texttt{HBVYN}, \texttt{HBVDAT}, \texttt{HBSAGRES} \\
HBCV1       & 49  & 1     & HBV/HCV laboratory assessments. & \texttt{HBV1YN}, \texttt{HBV1DAT}, \texttt{HBSAGRE1} \\
VS          & 25  & 1233  & Vital signs measurements. & \texttt{VSRES}, \texttt{VSTEMP}, \texttt{VSSBP} \\
CP          & 32  & 1262  & Child-Pugh assessment and component scoring. & \texttt{CPYN}, \texttt{CPRES}, \texttt{CPGRA} \\
IE          & 31  & 549   & Eligibility criteria assessment. & \texttt{IEDAT}, \texttt{INCL1}, \texttt{INCL2} \\
IE1         & 30  & 714   & Eligibility criteria (alternate set).. & \texttt{INCL01}, \texttt{INCL02}, \texttt{EXCL01} \\
TUBA        & 94  & 1295  & Intraoperative tumor pathology / surgical findings. & \texttt{TUBADSC}, \texttt{TUBATYPE}, \texttt{TUBAGRA} \\
BCLC        & 28  & 1261  & BCLC staging for primary liver cancer. & \texttt{BCLC}, \texttt{PLICA}, \texttt{PLICA1} \\
BCLC1       & 28  & 1     & BCLC staging (continuation record). & \texttt{BCLC1}, \texttt{BCPLICA}, \texttt{BCPLICA1} \\
VT          & 96  & 5321  & Efficacy assessment interview and follow-up. & \texttt{VTTYP}, \texttt{VTDAT1}, \texttt{VTDAT2} \\
VT1         & 96  & 2     & Efficacy assessment (continuation record). & \texttt{VT1TYP}, \texttt{VT1DAT1}, \texttt{VT1DAT2} \\
LB\_AFP     & 57  & 4613  & Tumor marker laboratory tests. & \texttt{AFPYN}, \texttt{LBAFPDAT}, \texttt{AFPRES} \\
AFP2        & 26  & 4166  & Other tumor marker laboratory tests. & \texttt{AFP2DAT}, \texttt{AFP2NM}, \texttt{AFP2RES} \\
LB\_CD      & 87  & 4392  & Lymphocyte subset testing panel. & \texttt{CDYN}, \texttt{CDDAT}, \texttt{CD3PRES} \\
LB\_BR      & 93  & 4474  & Complete blood count (CBC). & \texttt{BRYN}, \texttt{BRDAT}, \texttt{WBCRES} \\
LB\_BB      & 123 & 4583  & Serum biochemistry and hepatic function panel. & \texttt{LBBBYN}, \texttt{LBBBDAT}, \texttt{ALTRES} \\
LB\_BB1     & 105 & 1     & Serum biochemistry. & \texttt{LBBB1YN}, \texttt{LBBB1DAT}, \texttt{ALTRES1} \\
LB\_CYT     & 75  & 4325  & Cytokine testing panel. & \texttt{LBCYTYN}, \texttt{LBCYTDAT}, \texttt{IL1RES} \\
LB\_TT      & 55  & 4466  & Coagulation function testing panel. & \texttt{LBTTYN}, \texttt{LBTTDAT}, \texttt{PTRES} \\
ECOG        & 22  & 4459  & ECOG performance status assessment. & \texttt{ECOGYN}, \texttt{ECOG} \\
PATE        & 39  & 3265  & Postoperative retreatment information. & \texttt{PATEYN}, \texttt{PATETR1}, \texttt{PATETR2} \\
SG          & 21  & 1325  & Surgery-related records and key attributes. & \texttt{SGDAT}, \texttt{SGTYP} \\
RFT         & 75  & 1323  & Combined radiofrequency therapy records. & \texttt{UEXYN}, \texttt{UEXLO1}, \texttt{UEXLO2} \\
FIN         & 20  & 1325  & Number of lesions and summary lesion count. & \texttt{FINM} \\
FI          & 74  & 1489  & Lesion-level details (location and characteristics). & \texttt{FILO1}, \texttt{FILO2}, \texttt{FILO3} \\
TRDAT       & 20  & 27    & Postoperative treatment start date/time fields. & \texttt{TRDAT}, \texttt{TRDATS} \\
THET\_HQ    & 23  & 123   & Chemotherapy screening / question page. & \texttt{THETYN}, \texttt{THETDSC}, \texttt{THETUK} \\
THET        & 30  & 154   & Chemotherapy administration records. & \texttt{THETPHA}, \texttt{THESDAT}, \texttt{THETONGO} \\
TART\_HQ    & 20  & 453   & Targeted therapy screening / question page. & \texttt{TARTYN}, \texttt{TARTYN\_DEC} \\
TART        & 35  & 303   & Targeted therapy administration records. & \texttt{TARTPHA}, \texttt{TARSDAT}, \texttt{TARTONGO} \\
IMU\_HQ     & 20  & 404   & Immunotherapy screening / question page. & \texttt{IMUYN}, \texttt{IMUYN\_DEC} \\
IMU         & 35  & 290   & Immunotherapy administration records. & \texttt{IMUPHA}, \texttt{IMUSDAT}, \texttt{IMUONGO} \\
TACE\_HQ    & 20  & 615   & TACE screening / question page. & \texttt{TACEYN}, \texttt{TACEYN\_DEC} \\
TACE        & 30  & 609   & TACE procedure and regimen details. & \texttt{TACEDAT}, \texttt{TACETYP1}, \texttt{TACETYP2} \\
CMED        & 20  & 283   & Traditional Chinese medicine treatment records. & \texttt{CMEDYN}, \texttt{CMEDYN\_DEC} \\
RADI\_HQ    & 52  & 15    & Radiotherapy screening / question page. & \texttt{RADIYN}, \texttt{RADIMD1}, \texttt{RADIMD2} \\
RADI        & 21  & 6     & Radiotherapy administration records. & \texttt{RADIGN}, \texttt{RADIDAT}, \texttt{RADIDSC} \\
SF          & 44  & 3183  & SF-12 quality-of-life questionnaire. & \texttt{SFYN}, \texttt{SF1}, \texttt{SF2} \\
CM\_HQ      & 20  & 1235  & Concomitant medication screening / question. & \texttt{CMYN}, \texttt{CMYN\_DEC} \\
CM          & 34  & 6632  & Concomitant medication records. & \texttt{CMCAT}, \texttt{CMINDC}, \texttt{CMDOSE} \\
LB\_BP      & 24  & 1232  & Blood typing and related laboratory annotations. & \texttt{LBBPYN}, \texttt{LBBPDAT}, \texttt{LBBPSIG} \\
AE\_HQ      & 20  & 1231  & Adverse event screening / question page. & \texttt{AEYN}, \texttt{AEYN\_DEC} \\
AE          & 34  & 1     & Adverse event records. & \texttt{AETERM}, \texttt{AESTDAT}, \texttt{AEENDAT} \\
SAE         & 74  & 1     & Serious adverse event records and site contact. & \texttt{SAECAT1}, \texttt{SAECAT2}, \texttt{SAERPDAT} \\
SAE\_HQ     & 20  & 1231  & Serious adverse event screening / question page. & \texttt{SAEYN}, \texttt{SAEYN\_DEC} \\
LOAB\_HQ    & 20  & 51    & Local ablation screening / question page. & \texttt{LOABYN}, \texttt{LOABYN\_DEC} \\
LOAB        & 21  & 51    & Local ablation procedure records. & \texttt{LOABDAT}, \texttt{LOABMEAR}, \texttt{LOABMEAR\_DEC} \\
\end{longtable}
\endgroup

\section{Method and Contract Implementation Details}
\label{app:method_contract}
This section records the implementation artifacts used by the profiling, RWS-IR compilation, violation-guided execution, and certified reporting stages.

\subsection{Profile schema \texorpdfstring{$\Phi$}{Phi}}
\label{app:profile_schema}
The main paper defines the four profile blocks. The implementation stores the same object as a JSON contract with stable keys for high-level structural summaries, high-level knowledge roles, field-level structural statistics, and field-level knowledge mappings. Downstream stages do not consume free-form profile prose; they read these keys directly when compiling RWS-IR fields, validating generated code, and admitting report claims.

\paragraph{Structural profile $\Phi_{\cdot,S}$.}
For each table/sheet and column, we record deterministic statistics, including: inferred type, missingness, cardinality, value ranges, common categories, and candidate join keys.

\paragraph{Knowledge profile $\Phi_{\cdot,K}$.}
We map raw fields to standardized biomedical concepts (e.g., UMLS/SNOMED CT/ICD/LOINC) by candidate generation followed by deterministic validation checks:
(i) \emph{semantic type compatibility}, and
(ii) \emph{value-range consistency}.
Unmappable or ambiguous fields are retained with flags to propagate uncertainty into RWS-IR compilation, execution, and reporting.

\paragraph{Implementation serialization and fallback.}
The implementation serializes $\Phi$ as a JSON object with four named blocks: \texttt{high\_structural}, \texttt{low\_structural}, \texttt{high\_knowledge}, and \texttt{low\_knowledge}. The high-structural block stores dataset-level table relationships, candidate subject and visit keys, and worksheet-level roles. The low-structural block stores field-level type inference, missingness, cardinality, range checks, common categorical values, date-parsing status, and candidate temporal anchors. The high-knowledge block stores study-level roles, including index-date schema, exposure, outcome, censoring, confounder, proxy, and claim-scope roles. The low-knowledge block stores ontology candidates, accepted mappings, validation status, units, value-range plausibility, temporal-role compatibility, and ambiguity or rejection causes.

For each biomedical mapping candidate, the profiler applies semantic-type compatibility, unit consistency, value-range plausibility, and temporal-role compatibility checks. A field that fails these checks is retained in $\Phi_{L,K}$ with a failure tag rather than removed. Such fields remain available for structure-grounded summaries, missingness reporting, and data-quality diagnostics, but cannot authorize clinical-role binding, endpoint derivation, safety conclusions, causal claims, or high-stakes report language unless an explicit proxy assumption is recorded in the RWS-IR and admitted by the claim policy.

\subsection{Concrete heterogeneous profile instance}
\label{app:phi_instance}
Listing~\ref{lst:phi-instance} shows a compact de-identified JSON instance of $\Phi$ from the HCC benchmark. The example follows the same schema used by the implementation and contains aggregate profile statistics and field-level metadata; record-level values are provided in the released de-identified HCC tables rather than repeated in the text.

\begin{lstlisting}[language=json,basicstyle=\ttfamily\scriptsize,caption={Compact JSON instance of the heterogeneous data profile $\Phi$.},label={lst:phi-instance}]
{
  "serialization": "JSON",
  "notation_map": {
    "Phi^{H,S}": "heterogeneous_profile.high_structural",
    "Phi^{L,S}": "heterogeneous_profile.low_structural",
    "Phi^{H,K}": "heterogeneous_profile.high_knowledge",
    "Phi^{L,K}": "heterogeneous_profile.low_knowledge"
  },
  "heterogeneous_profile": {
    "high_structural": {
      "dataset_summary": {
        "workbook_type": "multi-sheet eCRF workbook",
        "benchmark_subject_count": 1294,
        "benchmark_visit_count": 12847
      },
      "sheet_inventory_subset": [
        {
          "sheet": "DM--Demographics",
          "role": "baseline demographics",
          "estimated_data_rows": 1294,
          "columns": 27,
          "join_keys": ["Subject_Id"]
        },
        {
          "sheet": "SG--Surgery",
          "role": "index surgery",
          "estimated_data_rows": 1324,
          "columns": 21,
          "join_keys": ["Subject_Id"]
        },
        {
          "sheet": "VT--Follow-up",
          "role": "postoperative follow-up",
          "estimated_data_rows": 5320,
          "columns": 96,
          "join_keys": ["Subject_Id", "Visit_No"]
        }
      ],
      "relationship_hints": [
        "DM.Subject_Id joins SG.Subject_Id and VT.Subject_Id.",
        "SG.SGDAT is the preferred index date.",
        "VT.Visit_No encodes repeated follow-up windows."
      ]
    },
    "low_structural": {
      "field_profiles": {
        "DM.SEX_DEC": {
          "dtype": "categorical",
          "missing_rate": 0.0,
          "top_values": {"male": 165, "female": 34}
        },
        "SG.SGDAT": {
          "dtype": "date-like text",
          "missing_rate": 0.005,
          "semantic_hint": "index surgery date"
        },
        "VT.PADHDAT": {
          "dtype": "partial date-like categorical",
          "missing_rate": 0.915,
          "quality_note": "sparse recurrence date"
        },
        "VT.DEDAT2": {
          "dtype": "date-like categorical",
          "missing_rate": 0.995,
          "quality_note": "nearly unavailable"
        }
      }
    },
    "high_knowledge": {
      "variable_roles": {
        "id": ["DM.Subject_Id", "SG.Subject_Id", "VT.Subject_Id"],
        "time_anchor": ["SG.SGDAT"],
        "baseline_covariate": [
          "DM.AGE", "DM.SEX_DEC", "BCLC.BCLC_DEC",
          "CP.CPGRA", "ECOG.ECOG"
        ],
        "outcome": [
          "VT.PADHDAT", "VT.PADRDAT",
          "VT.DEDAT1", "VT.DEDAT2"
        ]
      },
      "feasibility_assessment": "Cohort and follow-up summaries are feasible; survival analyses require an availability check."
    },
    "low_knowledge": {
      "ontology_mapping": {
        "DM.SEX_DEC": "Biological sex",
        "DM.AGE": "Age",
        "SG.SGDAT": "Surgical procedure date",
        "VT.PADHDAT": "Recurrence date"
      },
      "validation_status": {
        "DM.HEIGHT": "ambiguous: numeric values mixed with unknown tokens",
        "VT.PADHDAT": "valid_but_sparse",
        "VT.DEDAT2": "invalid_for_primary_survival_endpoint"
      },
      "analysis_constraints": [
        "Use only fields listed in field_profiles.",
        "Treat unknown or partial date tokens as missing unless the method supports interval censoring.",
        "Do not infer recurrence or death outcomes from missing date fields."
      ]
    },
    "ambiguity_flags": [
      "DM.HEIGHT", "DM.WEIGHT",
      "VT.PADHDAT", "VT.DEDAT2"
    ]
  }
}
\end{lstlisting}

\subsection{RWS-IR schema and static checks}
\label{app:ir_static_checks}
The RWS-IR is the typed study object validated before code generation. It records the study design, analysis grain, join graph, entity keys, temporal anchors, variable roles, methods, required artifacts, and claim policy. Table~\ref{tab:rws_ir_schema} lists the machine-readable fields and the admission checks applied before execution.

\begin{table*}[t]
\centering
\scriptsize
\setlength{\tabcolsep}{4pt}
\renewcommand{\arraystretch}{1.08}
\caption{Machine-readable fields in the RWS-IR and static checks before execution.}
\label{tab:rws_ir_schema}
\begin{tabularx}{\textwidth}{@{}p{0.08\textwidth}p{0.18\textwidth}X X@{}}
\toprule
\textbf{Symbol} & \textbf{Serialized key} & \textbf{Contents} & \textbf{Static admissibility checks} \\
\midrule
$D$ & \texttt{study\_design} & Study type, estimand boundary, and whether the requested output is descriptive, associative, or inferential. & Must be compatible with available tables, observed design, and claim policy. \\
$G$ & \texttt{analysis\_grain} & Subject-, visit-, encounter-, field-, or record-level analysis unit. & Must match key multiplicities and downstream aggregation. \\
$J$ & \texttt{join\_graph} & Tables, join keys, cardinalities, and permitted join paths. & All tables and keys must exist in $\Phi_{L,S}$; declared cardinality must match profiled relationships. \\
$K$ & \texttt{entity\_keys} & Subject, visit, encounter, treatment, and artifact identifiers. & Required keys must be present and nondegenerate under the declared grain. \\
$T$ & \texttt{temporal\_anchors} & Index date, event dates, censoring dates, windows, and ordering rules. & Date fields must be parseable or declared as partial/proxy dates; impossible orderings are disallowed. \\
$V$ & \texttt{variable\_roles} & Exposure, outcome, censoring, covariate, subgroup, proxy, and reporting variables. & Clinical roles must be supported by $\Phi_{L,K}$, a validated proxy flag, or a recorded assumption. \\
$M$ & \texttt{method} & Statistical method, fallback method, denominators, missingness policy, and diagnostics. & Method must be compatible with design, grain, variable support, and minimum data requirements. \\
$A$ & \texttt{required\_artifacts} & Required result tables, figures, manifests, logs, and report sections. & Artifacts must be executable in the sandbox and verifiable by manifests. \\
$\Pi_q$ & \texttt{claim\_policy} & Allowed and forbidden claim types, language constraints, and support levels. & Claim policy must be no stronger than the design and evidence support. \\
\bottomrule
\end{tabularx}
\end{table*}

Given a candidate $Z_0$, \texttt{IRCheckstatic} returns an accepted RWS-IR $Z$ and a set of static violations $U_{\mathrm{static}}$. A hard violation is any reference to a missing table, missing field, invalid key, unsupported clinical role, incompatible analysis grain, impossible temporal anchor, method that requires unavailable variables, or claim policy exceeding the design. A soft violation is a downgradeable limitation, such as sparse endpoint support, partial date availability, missing units, or a proxy variable that can support descriptive but not causal claims. Only $U_{\mathrm{static}}=\emptyset$ enters the execution workspace. When hard violations remain, ProCARE either repairs the RWS-IR, emits an infeasibility statement, or downgrades the requested analysis according to prompt specificity.

\begin{table*}[t]
\centering
\scriptsize
\setlength{\tabcolsep}{4pt}
\renewcommand{\arraystretch}{1.08}
\caption{Static violation types emitted by \texttt{IRCheckstatic}.}
\label{tab:static_violation_types}
\begin{tabularx}{\textwidth}{@{}p{0.18\textwidth}X X@{}}
\toprule
\textbf{Violation type} & \textbf{Trigger} & \textbf{Allowed repair} \\
\midrule
Missing schema reference & A table, field, or key in $Z_0$ is absent from $\Phi_{L,S}$. & Replace with a profiled field, remove the variable, or declare infeasibility for \textsc{Goal+Method}. \\
Unsupported semantic role & Exposure, endpoint, censoring, or confounder role is not validated in $\Phi_{L,K}$. & Use a validated proxy with limitation, downgrade to descriptive support, or stop. \\
Invalid join or grain & Join cardinality or aggregation conflicts with the declared analysis grain. & Change grain, revise the join graph, or aggregate before merging. \\
Invalid temporal anchor & Index, event, censoring, or follow-up dates are missing, partial, nonparseable, or ordered incorrectly. & Use a supported anchor, restrict to analyzable records, or report feasibility only. \\
Method infeasibility & The requested method requires unsupported design assumptions, insufficient events, or unavailable covariates. & Switch to a supported fallback method or emit an infeasibility statement. \\
Claim-policy excess & $\Pi_q$ allows causal, safety-critical, economic-cost, or generalization claims unsupported by the data. & Restrict $\Pi_q$ to descriptive, feasibility, or proxy-association claims. \\
\bottomrule
\end{tabularx}
\end{table*}

\subsection{RWS-IR and evidence certificate examples}
\label{app:rws_ir_certificate_examples}
Listing~\ref{lst:rws-ir-example} gives a compact RWS-IR instance for a recurrence-free survival task. The example shows the fields used by the static checker: study design, analysis grain, join graph, temporal anchors, variable-role bindings, method assumptions, required artifacts, and claim policy.

\begin{lstlisting}[language=json,basicstyle=\ttfamily\scriptsize,caption={Compact RWS-IR example for a recurrence-free survival task.},label={lst:rws-ir-example}]
{
  "study_design": "retrospective_cohort",
  "analysis_grain": "subject",
  "population": {
    "source_tables": ["DM", "SG", "VT"],
    "subject_key": "Subject_Id",
    "inclusion": ["non_missing(DM.Subject_Id)", "surgery_record_exists(SG)"]
  },
  "join_graph": [
    {"left": "DM.Subject_Id", "right": "SG.Subject_Id", "type": "one_to_one"},
    {"left": "DM.Subject_Id", "right": "VT.Subject_Id", "type": "one_to_many"}
  ],
  "temporal_anchors": {
    "index_date": "SG.SGDAT",
    "event_date_candidates": ["VT.PADHDAT", "VT.PADRDAT"],
    "censoring_date_candidates": ["VT.VTDAT1", "VT.VTDAT2"]
  },
  "variable_roles": {
    "exposure": "surgery cohort",
    "outcome": "recurrence-free survival proxy",
    "covariates": ["DM.AGE", "DM.SEX_DEC", "BCLC.BCLC_DEC", "CP.CPGRA"]
  },
  "method": {
    "primary": "Kaplan-Meier summary if temporal checks pass",
    "fallback": "descriptive follow-up availability when event dates are sparse"
  },
  "required_artifacts": ["cohort_flow", "followup_summary", "km_curve_or_feasibility_table"],
  "claim_policy": {
    "allowed": ["descriptive", "association_with_limitation"],
    "blocked": ["causal_effect", "safety_detail", "submission_grade_claim"]
  }
}
\end{lstlisting}

Listing~\ref{lst:certificate-example} gives a representative evidence certificate. The certificate records the claim text, its support level, the validated artifacts and code region that produced it, the RWS-IR fields that authorize the claim, and the profile entries or assumptions used by the admission rule.

\begin{lstlisting}[language=json,basicstyle=\ttfamily\scriptsize,caption={Evidence certificate example for an admitted factual claim.},label={lst:certificate-example}]
{
  "claim_id": "q_rfs_03",
  "claim_text": "Recurrence-free survival can be summarized for subjects with valid surgery and follow-up dates.",
  "claim_type": "descriptive_feasibility",
  "support_level": "profile_and_artifact_supported",
  "trace": {
    "artifact": "results/followup_summary.json",
    "figure": "figures/km_or_feasibility.pdf",
    "code_region": "analysis.py::build_rfs_dataset",
    "rws_ir_fields": [
      "temporal_anchors.index_date",
      "temporal_anchors.event_date_candidates",
      "method.primary",
      "claim_policy.allowed"
    ],
    "profile_entries": ["SG.SGDAT", "VT.PADHDAT", "VT.PADRDAT", "VT.VTDAT1", "VT.VTDAT2"],
    "assumptions": ["partial or unknown date tokens are treated as missing"]
  },
  "admission": {
    "trace_check": true,
    "policy_check": true,
    "language_check": true
  }
}
\end{lstlisting}

\subsection{Execution validators and violation-guided repair}
\label{app:execution_validators}
Execution-time checks are applied after code generation because row-level temporal ordering, manifest completeness, figure creation, and artifact consistency cannot be fully resolved statically. Validators return a localized violation set $U_t=V_{\mathrm{exec}}(Z_t,C_t,O_t,E_t,\Phi,S_t)$. Each violation records the failed contract clause, offending code or artifact, nearest valid profile entries, severity, and permitted repair actions. Listing~\ref{lst:violation_schema} shows the JSON schema used for one violation record.

\begin{lstlisting}[language=json,basicstyle=\ttfamily\scriptsize,caption={Schema for one execution-time contract violation.},label={lst:violation_schema}]
{
  "violation_id": "u_017",
  "stage": "execution",
  "contract_clause": "temporal_anchor.index_date_before_event",
  "severity": "hard",
  "offending_artifact": "analysis.py",
  "code_region": "build_survival_dataset",
  "observed_failure": "event_date precedes SG.SGDAT for at least one record",
  "nearest_profile_entries": ["SG.SGDAT", "VT.PADHDAT", "VT.PADRDAT"],
  "permitted_actions": ["filter_invalid_temporal_pairs", "report_temporal_limitation"],
  "blocks_acceptance": true
}
\end{lstlisting}

\begin{table*}[t]
\centering
\scriptsize
\setlength{\tabcolsep}{4pt}
\renewcommand{\arraystretch}{1.08}
\caption{Execution validators used before accepting an artifact bundle.}
\label{tab:execution_validators}
\begin{tabularx}{\textwidth}{@{}p{0.16\textwidth}X X@{}}
\toprule
\textbf{Validator} & \textbf{Check} & \textbf{Acceptance criterion} \\
\midrule
Entry-point validator & Runs \texttt{analysis.py} or \texttt{main.py} in the sandbox. & Exit code is zero and no failure sentinel such as \texttt{TASK\_FAILED} appears. \\
Schema validator & Parses code and generated outputs for table, field, and key usage. & Every referenced table, field, join key, and output column is traceable to $\Phi_{L,S}$ or a derived artifact. \\
Grain and join validator & Checks duplicate inflation, one-to-many joins, and final denominator consistency. & Output denominators match the declared analysis grain or disclose aggregation rules. \\
Temporal validator & Recomputes index, event, censoring, and follow-up windows. & Invalid date strings and impossible orderings are filtered, logged, or reported as limitations. \\
Method validator & Checks cohort filters, variables, denominators, covariates, and statistical choices against $Z.M$. & Method matches the admitted RWS-IR or a declared fallback. \\
Artifact validator & Checks result and figure manifests, file existence, and generated figure formats. & Required tables, figures, manifests, and logs exist and are nonempty when applicable. \\
Claim-preparation validator & Builds the evidence map used by the report generator. & Each reportable statistic has a source artifact, producing code region, and profile/IR trace. \\
\bottomrule
\end{tabularx}
\end{table*}

\begin{algorithm}[t]
\small
\caption{Execution acceptance and localized repair}
\label{alg:execution_acceptance}
\begin{algorithmic}[1]
\Require Accepted RWS-IR $Z$, profile $\Phi$, workspace policy $\Pi$, maximum attempts $B$
\Ensure Accepted code $C^*$, artifacts $O$, evidence $E$, assumptions $A$, or failure trace
\For{$t=1,\ldots,B$}
  \State Generate or edit code $C_t$ conditioned on $Z$, $\Phi$, $\Pi$, and prior violations $U_{t-1}$.
  \State Execute the selected entry point in the sandbox and collect logs, result manifests, and figure manifests.
  \State Compute $U_t = V_{\mathrm{exec}}(Z,C_t,O_t,E_t,\Phi,S_t)$.
  \If{$U_t=\emptyset$}
    \State Accept $C^*=C_t$ and return $(C^*,O_t,E_t,A_t)$.
  \Else
    \State Write $U_t$ to \texttt{violation\_set.json} with localized repair actions.
  \EndIf
\EndFor
\State Return the final failure trace or a supported downgrade with explicit limitations.
\end{algorithmic}
\end{algorithm}

\subsection{Evidence certificates and claim admission}
\label{app:claim_admission}
A factual claim is any report sentence that asserts a dataset property, cohort count, variable availability, statistical result, visualization interpretation, clinical association, feasibility conclusion, or limitation derived from the data. Section headings, procedural statements, and generic background statements are not counted unless they contain dataset-specific factual content. For ProCARE, certificates are emitted during report generation. For baselines without explicit certificates, the same checker performs post-hoc reconstruction from the submitted code, logs, artifacts, and report.

Each claim $q$ receives a certificate $e_q=\langle o_q,c_q,z_q,\phi_q,a_q\rangle$, where $o_q$ is a supporting artifact, $c_q$ is the executed code region that created the artifact, $z_q$ is the RWS-IR field or method authorizing the claim, $\phi_q$ is the relevant profile entry, and $a_q$ is an optional recorded assumption or limitation. The trace check passes only when all required links exist and agree on variables, denominators, analysis grain, temporal windows, and method. The policy check passes only when the claim type and wording are allowed by $Z.\Pi_q$ at the available support level.

\begin{table*}[t]
\centering
\scriptsize
\setlength{\tabcolsep}{4pt}
\renewcommand{\arraystretch}{1.08}
\caption{Claim types used by the certificate gate.}
\label{tab:claim_types}
\begin{tabularx}{\textwidth}{@{}p{0.18\textwidth}X X@{}}
\toprule
\textbf{Claim type} & \textbf{Allowed support} & \textbf{Examples of blocked wording} \\
\midrule
Descriptive & Validated table, count, summary statistic, figure, or profile entry. & Claims that imply generalization beyond the analyzed cohort. \\
Feasibility or data-quality & Profile entries, missingness summaries, temporal-validity logs, or execution artifacts. & Treating missingness or feasibility as clinical outcome evidence. \\
Proxy association & Valid proxy flag, artifact support, and explicit limitation. & Causal treatment-effect language or safety conclusion language. \\
Causal or comparative effect & Requires design assumptions, adjustment plan, and method support declared in $Z$. & Causal language when only descriptive or proxy support is available. \\
Safety-critical or economic-cost & Requires detailed safety records or direct cost/billing fields. & Detailed AE/SAE term claims when only screening pages exist; cost-effectiveness claims without cost variables. \\
\bottomrule
\end{tabularx}
\end{table*}

Let $N_q$ be the number of extracted factual claims, $C_q$ the number with admitted evidence certificates, $U_q$ the number without admissible evidence, and $L_q$ the number whose language exceeds the claim policy. We compute
\[
\mathrm{CTC}=C_q/N_q,\qquad \mathrm{UCR}=U_q/N_q,\qquad \mathrm{CLV}=L_q/N_q.
\]
A claim can be both unsupported and a language violation, but each metric is normalized by the same factual-claim denominator. For the claim-diagnostic table in the main paper, metrics are computed per run and then averaged across the 12 tasks in each prompt-specificity group; the overall row averages across all 36 HCC tasks.

\subsection{Profile-grounded state artifacts}
\label{app:state_artifacts}
The workflow state is represented by typed JSON artifacts, workspace files, skills, and validation logs. No vector database is used. Each downstream module reads the relevant state artifact from named files or typed prompt payloads, so intermediate decisions remain auditable without re-querying raw tables.
Concretely, the persistent dataset state is the cached profile $\Phi$, while the per-job state $\mathcal{S}_t$ may contain \texttt{excel\_info.json}, \texttt{rws\_ir.json}, \texttt{violation\_set.json}, \texttt{execution\_spec.md}, generated code, execution logs, result manifests, figure manifests, evidence certificates, report sections, and disclosed assumptions. These files are bookkeeping carriers for the profile contract: downstream stages may reuse a fact only when it can be traced to profiled tables, validated fields, executable artifacts, or explicitly recorded assumptions.

\begin{table*}[t]
\centering
\scriptsize
\setlength{\tabcolsep}{4pt}
\renewcommand{\arraystretch}{1.08}
\caption{Operational mapping between profile-grounded state roles and implementation artifacts.}
\label{tab:state-artifacts}
\begin{tabularx}{\textwidth}{@{}l X X X@{}}
\toprule
\textbf{State role} & \textbf{Stored artifacts} & \textbf{Example contents} & \textbf{Accessed by} \\
\midrule
Execution trace & \texttt{exec\_logs.json}, \texttt{exec\_status.json}, \texttt{opencode\_live.log}, \texttt{violation\_set.json}, audit records. & execution traces, contract violations, validation outcomes, disclosed assumptions. & violation-guided execution, report generator, frontend status view. \\
Working context & \texttt{excel\_info.json}, \texttt{rws\_ir.json}, \texttt{execution\_spec.md}, selected sheet payloads. & selected sheets, required fields, join keys, variable roles, temporal anchors, method specification, claim policy. & IR compiler, execution workspace, report context builder. \\
Profile semantics & $\Phi^{H,K}$, $\Phi^{L,K}$, ontology mappings, reporting prompts. & exposure/outcome/covariate roles, plausibility checks, clinical terminology, section semantics. & IR compiler, code generator, report DAG nodes. \\
Certified outputs & result manifests, figure manifests, evidence certificates, report sections, disclosed assumptions. & validated tables and figures, claim traces, support levels, limitations, section drafts. & claim checker, report DAG, final report renderer. \\
Operational policy & IR compilation, profiling, coding, figure, validation, and report-generation policies. & coding rules, artifact schemas, figure manifest schema, section templates, validation gates, claim policies. & execution workspace, report builder. \\
\bottomrule
\end{tabularx}
\end{table*}

\begin{algorithm}[t]
\small
\caption{Profile-grounded state update and access in ProCARE}
\label{alg:state-update}
\begin{algorithmic}[1]
\Require User query $q$, dataset $D$, profile cache $\mathcal{C}$
\Ensure Analysis artifacts, report, and updated job state
\State Compute dataset hash $h=\mathrm{MD5}(D)$.
\State Load $\Phi$ from $\mathcal{C}[h]$ if available; otherwise profile $D$ and cache $\Phi$.
\State Create job workspace $W$ and write the profile into \texttt{excel\_info.json}.
\State Compile candidate RWS-IR $Z_0$ from $q$ and the high-level profile.
\State Run static IR checks against $\Phi$ and the job state.
\State Write accepted $Z$ to \texttt{rws\_ir.json} and write execution context to \texttt{execution\_spec.md}.
\State Run violation-guided execution using $Z$, low-level profile blocks, and operational policy artifacts.
\State Append execution logs, violation sets, validation results, and figure manifests to the job state.
\State Generate certified report sections from the reporting DAG using $\Phi$, $Z$, logs, figures, evidence certificates, and existing sections.
\end{algorithmic}
\end{algorithm}

\subsection{Ontology mapping coverage}
The knowledge profiler maps raw fields to biomedical concepts from UMLS, SNOMED CT, ICD-10, and LOINC through candidate generation followed by type-compatibility and value-range checks. Overall field-level mapping coverage is 0.83. Coverage for critical task fields, including exposures, outcomes, confounders, and temporal anchors, is 0.91.

Unmapped fields remain in $\Phi_{L,K}$ with failure tags. The main failure categories are abbreviations or local jargon without a stable ontology match (0.07), type incompatibility between candidate concepts and observed values (0.05), ambiguity among indistinguishable candidates (0.03), and mixed units or inconsistent coding systems (0.02).

\subsection{Artifact bundle contract}
\label{app:artifact_bundle}
The prompt schema contains the task definition, optional guidance, and required deliverables. Each run returns a validated RWS-IR with required tables, join keys, variables, temporal anchors, methods, artifacts, and claim policy; one executable entry point (\texttt{analysis.py} or \texttt{main.py}) with optional helper modules for data loading and artifact generation; and a structured scientific report with evidence certificates. Generated SQL or pandas operations may reference only tables and columns present in the profile.

The contract depends on prompt specificity. Under \texttt{GOAL+METHOD}, the agent follows the requested variables and stops when required data are unavailable. Under \texttt{GOAL} and \texttt{GOAL+APPROACH}, the agent may generate the closest feasible analysis, but the report must disclose data limitations and method deviations.

The artifact bundle is the unit of evaluation. A bundle is accepted for full scoring only if it contains the files in Table~\ref{tab:artifact_bundle_schema}. Missing execution-dependent artifacts produce N/A for metrics that cannot be computed, while schema, report, and claim diagnostics are still computed when the submitted artifacts allow post-hoc checking.

\begin{table*}[t]
\centering
\scriptsize
\setlength{\tabcolsep}{4pt}
\renewcommand{\arraystretch}{1.08}
\caption{Required artifact-bundle schema.}
\label{tab:artifact_bundle_schema}
\begin{tabularx}{\textwidth}{@{}p{0.20\textwidth}p{0.24\textwidth}X@{}}
\toprule
\textbf{Artifact} & \textbf{Typical file name} & \textbf{Evaluation use} \\
\midrule
RWS-IR & \texttt{rws\_ir.json} & PF, method validation, temporal and grain checks, claim policy. \\
Executable entry point & \texttt{analysis.py} or \texttt{main.py} & XE, CQ, artifact generation, code-region traces. \\
Execution logs & \texttt{exec\_logs.json}, \texttt{exec\_status.json} & Runtime pass/fail, violation localization, evidence reconstruction. \\
Results manifest & \texttt{results\_manifest.json} & Result-table existence, denominators, generated statistics. \\
Figure manifest & \texttt{figure\_manifest.json} & Figure existence, caption/axis metadata, chart-fidelity checks. \\
Violation set & \texttt{violation\_set.json} & VRR and repair diagnostics. \\
Evidence certificates & \path{evidence_certificates.json} & CTC, UCR, CLV, report traceability. \\
Assumptions and limitations & \texttt{assumptions.json} or report section & Downgrade handling and policy-compatible language checks. \\
Final report & \texttt{report.md}, \texttt{report.json}, or rendered report & RS, SL, factual-claim extraction, evidence-policy matching. \\
\bottomrule
\end{tabularx}
\end{table*}

\subsection{Profile-constrained coding harness}
\label{app:coding-harness}
The code-generation stage is implemented as a controlled execution workspace. The workspace writes the profile, RWS-IR, local policies, generated code, figures, logs, violation sets, and validation records into one job directory. The coding agent can edit and execute files inside this workspace, while the backend reruns the selected entry point and rejects incomplete artifact bundles. Table~\ref{tab:coding-harness-artifacts} maps the execution components to concrete implementation artifacts.

\begin{table*}[t]
\centering
\small
\setlength{\tabcolsep}{4pt}
\renewcommand{\arraystretch}{1.08}
\caption{Concrete artifacts used by the profile-constrained coding harness.}
\label{tab:coding-harness-artifacts}
\begin{tabularx}{\textwidth}{@{}l X X@{}}
\toprule
\textbf{Harness component} & \textbf{Concrete artifact} & \textbf{Purpose} \\
\midrule
Workspace $W$ & Isolated job directory with generated scripts, plots, logs, and report fragments. & Keeps one RWS task reproducible and separates generated files from source code. \\
Context $X$ & \texttt{rws\_ir.json}, \texttt{excel\_info.json}, and \texttt{execution\_spec.md}. & Supplies objective, selected sheets, required fields, join keys, temporal anchors, method choices, and claim policy. \\
Policies $\Pi$ & Local coding instructions, figure policy, \texttt{publication\_plotting.py}, and artifact schema. & Converts profile and reporting requirements into file-level rules that the coding agent must follow. \\
Tools $\mathcal{T}$ & File editing, Python execution, local skill use, and sandbox validation. & Allows edit-run-fix behavior inside the controlled workspace. \\
Validators $\mathcal{V}$ & Entry-point execution, \texttt{TASK\_FAILED} sentinel check, \texttt{figure\_manifest.json}, and PNG/PDF/SVG existence checks. & Turns soft generation instructions into completion gates for executable artifacts. \\
Traces $L$ & \texttt{opencode\_live.log}, \texttt{exec\_logs.json}, \texttt{exec\_status.json}, and \texttt{violation\_set.json}. & Records evidence for violation-guided repair, failure reporting, and later report grounding. \\
\bottomrule
\end{tabularx}
\end{table*}

The constraint intent remains the same, but enforcement happens through observable gates instead of a separate symbolic solver. Schema constraints are checked against \texttt{excel\_info.json}; methodological constraints are checked against the RWS-IR and profile semantics; artifact constraints are checked by execution and manifest validation. If the profile does not support a required variable, temporal anchor, or statistical procedure, the execution workspace requires a failed run or an explicit limitation rather than an unsupported substitute.

\subsection{DAG reporting dependency rules}
\label{app:scoring}
Report generation uses a dependency DAG $G=(T,E)$. Each node $t\in T$ is an atomic reporting task that either constructs a validated evidence artifact or renders a report section from validated artifacts. Each directed edge $(t_i,t_j)\in E$ indicates that $t_j$ requires artifacts produced by $t_i$.
Table~\ref{tab:dag-symbols} defines the task metadata used by the graph.

\begin{table}[t]
\centering
\small
\setlength{\tabcolsep}{4pt}
\renewcommand{\arraystretch}{1.08}
\caption{Task metadata used in the reporting DAG.}
\label{tab:dag-symbols}
\begin{tabularx}{\linewidth}{@{}l X@{}}
\toprule
\textbf{Symbol} & \textbf{Meaning} \\
\midrule
$\mathrm{prod}(t)$ & Artifacts produced by task $t$, such as tables, figures, statistics, or text. \\
$\mathrm{req}(t)$ & Prerequisites used by task $t$, such as variables, cohorts, temporal anchors, or method assumptions. \\
\bottomrule
\end{tabularx}
\end{table}

\paragraph{Required chapter schema for RS.}
The minimal required chapters are listed in Appendix~\ref{app:report_schema}. RS is computed as the fraction of required chapters covered with non-trivial content.

\paragraph{Support score and edge weights.}
For each dependency edge $(t_i,t_j)$, we assign a weight:
\begin{equation}
w_{ij} = s_{ij} \cdot \sigma\!\left(\AlphaSupport + \BetaSupport \, c_{ij}\right),
\label{eq:wij}
\end{equation}
where $\sigma(x)=\frac{1}{1+e^{-x}}$ maps to $[0,1]$ and $s_{ij}\in\{0,1\}$ is a feasibility gate indicating whether $t_j$ can run from the available artifacts and recorded assumptions.

We compute $c_{ij}$ from the four deterministic signals in Table~\ref{tab:support-signals}:
\begin{equation}
c_{ij} = \WVar S_{\mathrm{var}} + \WTemp S_{\mathrm{temp}} + \WSize S_{\mathrm{size}} + \WCons S_{\mathrm{cons}}.
\label{eq:cij}
\end{equation}

All experiments use $\alpha=\AlphaSupport$, $\beta=\BetaSupport$, $w_{\mathrm{var}}=\WVar$, $w_{\mathrm{temp}}=\WTemp$, $w_{\mathrm{size}}=\WSize$, $w_{\mathrm{cons}}=\WCons$, and $N_{\min}=\NMin$. These constants are fixed before evaluation. They encode the ordering preference that validated variable support, temporal validity, effective sample size, and method-design consistency should be present before a result section is promoted above a limitation or feasibility section. The traversal rule changes report ordering and section expansion only; it does not change executed statistics or certify unsupported claims.

\begin{table}[t]
\centering
\small
\setlength{\tabcolsep}{4pt}
\renewcommand{\arraystretch}{1.08}
\caption{Support signals for reporting-task prioritization.}
\label{tab:support-signals}
\begin{tabularx}{\linewidth}{@{}l X@{}}
\toprule
\textbf{Signal} & \textbf{Definition} \\
\midrule
$S_{\mathrm{var}}$ & Variable availability in the analysis-ready cohort, with missingness penalties. One instantiation is $\frac{1}{|\mathcal{V}|}\sum_{v\in\mathcal{V}}\max(0,1-\mathrm{miss}(v))$. \\
$S_{\mathrm{temp}}$ & Temporal validity, defined as $1-r_{\mathrm{invalid}}$, where $r_{\mathrm{invalid}}$ is the fraction of missing, invalid, or impossible temporal prerequisites. \\
$S_{\mathrm{size}}$ & Sample adequacy, defined as $\min(1,N_{\mathrm{eff}}/\NMin)$ after filtering and joins. \\
$S_{\mathrm{cons}}$ & Method-design consistency, with partial credit only when execution is feasible under recorded assumptions. \\
\bottomrule
\end{tabularx}
\end{table}

\paragraph{Ordering policy.}
We generate report outlines with a weighted topological traversal. At each step, among tasks whose prerequisites are satisfied, we select:
\begin{equation}
t^{\star}=\arg\max_{t\in \mathrm{frontier}} \sum_{(u,t)\in E} w_{ut}.
\label{eq:topo}
\end{equation}

\subsection{Instantiated reporting DAG}
\label{app:dag-example}
For the recurrence-free survival benchmark instance, the task uses \texttt{SG.SGDAT} as the index date, the earliest non-missing \texttt{VT.PADHDAT}/\texttt{VT.PADRDAT} as the recurrence event date, and the latest non-missing \texttt{VT.VTDAT1}/\texttt{VT.VTDAT2} as censoring time. In the workbook, the surgery cohort contains 1,295 subjects, 1,215 subjects have analyzable RFS follow-up, and 304 subjects have valid recurrence events. The same graph shows evidence gating: because the \texttt{AE} and \texttt{SAE} detail tables contain no valid records, detailed AE profiling is blocked and the evidence gap is propagated to the limitations section.

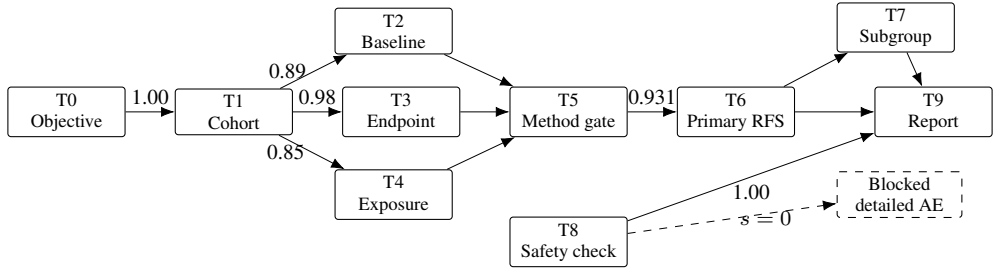
\begin{figure*}[t]
\centering
\footnotesize
\begin{tikzpicture}[
    node distance=0.75cm and 0.65cm,
    task/.style={draw, rounded corners=1.2pt, align=center, inner sep=2.2pt, minimum width=1.55cm, minimum height=0.58cm, font=\scriptsize},
    blocked/.style={draw, dashed, rounded corners=1.2pt, align=center, inner sep=2.2pt, minimum width=1.65cm, minimum height=0.58cm, font=\scriptsize},
    edge/.style={-{Latex[length=1.8mm]}, line width=0.35pt},
    blockededge/.style={-{Latex[length=1.8mm]}, dashed, line width=0.35pt}
]
\node[task] (t0) {T0\\Objective};
\node[task, right=of t0] (t1) {T1\\Cohort};
\node[task, above right=0.45cm and 0.55cm of t1] (t2) {T2\\Baseline};
\node[task, right=of t1] (t3) {T3\\Endpoint};
\node[task, below right=0.45cm and 0.55cm of t1] (t4) {T4\\Exposure};
\node[task, right=of t3] (t5) {T5\\Method gate};
\node[task, right=of t5] (t6) {T6\\Primary RFS};
\node[task, above right=0.45cm and 0.55cm of t6] (t7) {T7\\Subgroup};
\node[blocked, below right=0.45cm and 0.55cm of t6] (t8b) {Blocked\\detailed AE};
\node[task, right=1.05cm of t6] (t9) {T9\\Report};
\node[task, below=1.05cm of t5] (t8) {T8\\Safety check};

\draw[edge] (t0) -- node[above] {1.00} (t1);
\draw[edge] (t1) -- node[left] {0.89} (t2);
\draw[edge] (t1) -- node[above] {0.98} (t3);
\draw[edge] (t1) -- node[left] {0.85} (t4);
\draw[edge] (t2) -- (t5);
\draw[edge] (t3) -- (t5);
\draw[edge] (t4) -- (t5);
\draw[edge] (t5) -- node[above] {0.931} (t6);
\draw[edge] (t6) -- (t7);
\draw[edge] (t6) -- (t9);
\draw[edge] (t7) -- (t9);
\draw[edge] (t8) -- node[below] {1.00} (t9);
\draw[blockededge] (t8) -- node[right] {$s=0$} (t8b);
\end{tikzpicture}
\caption{Instantiated reporting DAG for the recurrence-free survival task. Solid edges denote feasible artifact dependencies and show selected support scores; the dashed edge records the blocked AE-detail analysis, which is routed to limitations rather than promoted as a safety result.}
\label{fig:rfs_dag}
\end{figure*}

\begin{table*}[t]
\centering
\small
\setlength{\tabcolsep}{3.5pt}
\renewcommand{\arraystretch}{1.08}
\caption{Instantiated report-task DAG for the recurrence-free survival benchmark instance. Each node declares required inputs and produced artifacts. Edges encode artifact prerequisites; infeasible edges are not selected and instead propagate limitations.}
\label{tab:dag-rfs-nodes}
\begin{tabularx}{\textwidth}{@{}l l X X l@{}}
\toprule
\textbf{Node} & \textbf{Task} & \textbf{Required inputs} & \textbf{Produced artifacts} & \textbf{Report role} \\
\midrule
T0 & Objective binding & User prompt, heterogeneous profile, report schema & Objective, endpoint definition, required section plan & Study overview \\
T1 & Cohort and attrition & \texttt{SG.Subject\_Id}, \texttt{SG.SGDAT} & Analysis cohort, attrition table, index-date assumptions & Overview / Attrition \\
T2 & Baseline covariates & T1, \texttt{DM}, \texttt{BCLC}, \texttt{CP}, \texttt{ECOG} & Table 1, missingness table, covariate map & Baseline characteristics \\
T3 & Endpoint and censoring & T1, \texttt{VT.PADHDAT}, \texttt{VT.PADRDAT}, \texttt{VT.VTDAT1}, \texttt{VT.VTDAT2} & RFS dataset, event/censor flags, temporal validity log & Primary analysis \\
T4 & Exposure matrix & T1, \texttt{TACE}, \texttt{TART}, \texttt{IMU}, \texttt{PATE} & Treatment exposure matrix, dated exposure availability & Primary / Subgroup \\
T5 & Method feasibility gate & T2, T3, T4, procedural policy & KM feasibility decision and caveats & Methods \\
T6 & Primary RFS analysis & T3, T4, T5 & KM curve, RFS summary table, primary result text & Primary analysis \\
T7 & Subgroup/sensitivity & T2, T4, T6 & Subgroup tables and caveats & Sensitivity analysis \\
T8 & Safety evidence check & \texttt{AE\_HQ}, \texttt{SAE\_HQ}, \texttt{AE}, \texttt{SAE} & Safety limitation statement; blocked detailed AE profiling & Safety / Limitations \\
T9 & Report synthesis & Prior artifacts and evidence map & Final structured report and claim-evidence map & Final report \\
\bottomrule
\end{tabularx}
\end{table*}

\begin{table*}[t]
\centering
\small
\setlength{\tabcolsep}{3.5pt}
\renewcommand{\arraystretch}{1.08}
\caption{Selected dependency edges and support scores for the instantiated DAG. The feasibility gate $s_{ij}$ blocks hard-infeasible dependencies. The support score $c_{ij}$ summarizes deterministic evidence from variable availability, temporal validity, sample adequacy, and method-design consistency.}
\label{tab:dag-rfs-edges}
\begin{tabularx}{\textwidth}{@{}l X c c@{}}
\toprule
\textbf{Edge} & \textbf{Dependency rationale} & $\mathbf{s_{ij}}$ & $\mathbf{c_{ij}}$ \\
\midrule
T0 $\rightarrow$ T1 & Cohort construction requires the objective, index date, and output contract. & 1 & 1.00 \\
T1 $\rightarrow$ T2 & Baseline summaries must be computed within the index cohort; BCLC and Child--Pugh coverage is high, while age missingness must be disclosed. & 1 & 0.89 \\
T1 $\rightarrow$ T3 & RFS endpoints and censoring must be aligned to surgery date; 1,215 subjects have usable follow-up and one temporal anomaly is flagged. & 1 & 0.98 \\
T1 $\rightarrow$ T4 & Treatment exposure grouping must be aligned to the index cohort; treatment proxies cover 1,104 subjects but dated exposures are incomplete. & 1 & 0.85 \\
T5 $\rightarrow$ T6 & The primary RFS analysis is selected only after the method gate passes; the combined gate has $S_{\mathrm{var}}=0.828$, $S_{\mathrm{temp}}=0.999$, $S_{\mathrm{size}}=1.000$, and $S_{\mathrm{cons}}=1.000$. & 1 & 0.931 \\
T8 $\rightarrow$ detailed AE profiling & Detailed AE/SAE profiling requires non-empty detail tables; both detail tables contain no valid records. & 0 & 0.00 \\
T8 $\rightarrow$ T9 & The empty AE/SAE detail tables support a limitation statement rather than unsupported safety claims. & 1 & 1.00 \\
\bottomrule
\end{tabularx}
\end{table*}

\subsection{Standard report schema}
\label{app:report_schema}
The reporting engine targets the sections in Table~\ref{tab:report-schema}.

\begin{table}[t]
\centering
\small
\setlength{\tabcolsep}{4pt}
\renewcommand{\arraystretch}{1.08}
\caption{Required RWS report sections used by the RS metric.}
\label{tab:report-schema}
\begin{tabularx}{\linewidth}{@{}l X@{}}
\toprule
\textbf{Section} & \textbf{Expected content} \\
\midrule
Study Overview & Cohort definition and objective statement. \\
Attrition Analysis & Patient disposition and eligibility filtering steps. \\
Baseline Characteristics & Demographics, clinical history, baseline vitals, and baseline labs. \\
Primary Analysis & Main efficacy or exposure-outcome results, such as KM curves or hazard ratios. \\
Safety Profiling & Adverse-event screening summaries and available safety proxies. \\
Sensitivity/Subgroup Analysis & Checks across predefined strata. \\
Limitations & Data constraints, method limitations, and potential biases. \\
\bottomrule
\end{tabularx}
\end{table}

\section{Experimental Protocol and Reproducibility}
\label{app:experimental_protocol}

\subsection{Execution environment and cost accounting}
\label{app:cost_accounting}
Generated analysis code is rerun from a clean job workspace in a Python 3.9 sandbox with network access disabled. The execution image includes \texttt{pandas}, \texttt{numpy}, \texttt{scipy}, \texttt{matplotlib}, \texttt{seaborn}, and \texttt{statsmodels}. The evaluator records stdout, stderr, return code, generated files, result manifests, figure manifests, violation sets, and failure sentinels. LLM calls are made outside the code-execution sandbox; generated code cannot fetch external data during scoring.

Token use is reported for ProCARE because baselines use different interfaces and are not directly comparable at the prompt level. Profiling is a reusable dataset-level cost. For the primary HCC schema, profiling uses approximately $T_{\mathrm{prof}}\approx65\mathrm{k}$ tokens, and each query uses approximately $T_{\mathrm{inf}}\approx38\mathrm{k}$ tokens for IR compilation, repair prompts, and certified reporting. For $N$ tasks on the same dataset,
\[
\bar{T}(N)=T_{\mathrm{inf}}+T_{\mathrm{prof}}/N.
\]
Thus $\bar{T}(1)\approx103\mathrm{k}$ tokens, $\bar{T}(20)\approx41.3\mathrm{k}$ tokens, and $\bar{T}(36)\approx39.8\mathrm{k}$ tokens per query.

\subsection{Systems and baseline adaptation}
\label{app:baselines}
All methods receive the same natural-language task prompt, raw data access boundary, deliverable request, and inference budget. Direct Gemini-3-Pro baselines and direct Gemini-3-Pro scaffolded controls use the same model as the default ProCARE backbone. Claude Code and Codex are evaluated as native code-agent baselines: they can edit files and run code through their coding-agent interfaces, but they do not receive ProCARE's profile object, typed RWS-IR, evidence-certificate admission policy, or certified reporting DAG unless they infer or generate analogous files from the prompt. ProCARE additionally constructs and uses the heterogeneous profile, typed RWS-IR, execution validators, evidence certificates, and certified reporting DAG. Baselines are evaluated under a post-hoc contract checker so that schema use, execution, artifact completeness, and claim traceability are measured against the same profile-derived constraints.

\begin{table*}[t]
\centering
\scriptsize
\setlength{\tabcolsep}{4pt}
\renewcommand{\arraystretch}{1.08}
\caption{Baseline adaptation and scoring rules.}
\label{tab:baseline-protocol}
\begin{tabularx}{\textwidth}{@{}p{0.18\textwidth}X X@{}}
\toprule
\textbf{Method family} & \textbf{Run protocol} & \textbf{Scoring rule} \\
\midrule
ReAct and ToolLLM & Agentic tool-use baselines are asked to inspect the tables, write executable analysis code, generate artifacts, and produce a report under the same prompt. & Deterministic validators execute the final entry point when provided; reports are checked post hoc for evidence support. \\
TableGPT-Agent and ChatExcel-Max & Table-analysis agents are asked to generate the requested analysis and report. & If no executable artifact bundle is produced, execution-dependent metrics are N/A; report and claim metrics are computed when possible. \\
Gemini-3-Pro direct baseline & Gemini-3-Pro receives the task and data-access instructions and is asked to return code, artifacts, and report content. & Generated code is executed in the same sandbox and evaluated by the same artifact and claim validators. \\
TableGPT2 & The table model is evaluated on table-analysis outputs and executable artifacts when available. & Metrics requiring a complete bundle are N/A when the method does not emit the required artifact. \\
Claude Code and Codex & Native coding-agent baselines receive the benchmark prompt, data-access path, artifact contract, and sandbox budget through their repository editing interfaces. They may create, edit, execute, and repair files within the job workspace. & The final workspace is re-executed from a clean scorer sandbox. Execution, artifact, clinical/statistical language, and claim-trace metrics are computed post hoc from the submitted files and report. \\
ProCARE & Full profile--IR--execution--certificate--DAG workflow. & All deterministic and judge metrics are computed on the accepted artifact bundle. \\
\bottomrule
\end{tabularx}
\end{table*}

A method is not penalized twice for a missing executable bundle. XE is zero when an executable entry point is submitted but fails, while metrics that require absent artifacts are marked N/A. Schema feasibility, report structure, scientific language, and post-hoc claim diagnostics are still computed whenever enough submitted content is available. Code-agent baselines typically produce executable bundles and stronger local repairs, but their reports are still scored without granting credit for ProCARE-only objects such as accepted RWS-IRs or admitted evidence certificates.

\subsection{Fairness-controlled scaffolded baselines}
\label{app:scaffolded_baselines}
To address the concern that the artifact contract may favor ProCARE, we conducted a scaffolded-baseline evaluation under the same final checker used for the main baseline comparison. These RQ1 fairness controls differ from the RQ2 mechanism ablations in Appendix~\ref{app:ablation_protocol}: ablations remove one component from ProCARE, whereas scaffolded baselines add evaluator-facing information and output schemas to external baselines or a direct Gemini-3-Pro run. Direct Gemini-3-Pro controls use the same provider endpoint, model identifier, decoding settings, and inference budget as the default ProCARE backbone. The scaffold gives non-ProCARE methods the same schema profile $\Phi$, artifact-bundle template, raw data access boundary, execution sandbox, and three-turn repair budget needed to satisfy the evaluator. It does not give baselines ProCARE's IR compiler, certificate admission gate, certified reporting DAG, or hidden planner state. Rows that mention a study-spec or claim-trace schema only require compatible JSON files for scoring.

The shared artifact template requires: \texttt{study\_spec.json}, \texttt{analysis.py} or \texttt{main.py}, \texttt{result\_manifest.json}, \texttt{figure\_manifest.json}, \texttt{claim\_trace.json} when the method attempts explicit claim tracing, \texttt{assumptions.json}, and \texttt{report.md}. For methods that do not emit a typed RWS-IR, the evaluator maps the submitted \texttt{study\_spec.json} into the canonical scoring schema before computing PF. For methods that do not emit explicit evidence certificates, CTC/UCR/CLV are computed by the same post-hoc evidence reconstruction used in Table~\ref{tab:baseline-protocol}.

\begin{table*}[t]
\centering
\scriptsize
\setlength{\tabcolsep}{4pt}
\renewcommand{\arraystretch}{1.08}
\caption{Inputs and controls used in the scaffolded-baseline fairness evaluation. These are baseline adaptations, not ProCARE component ablations.}
\label{tab:scaffolded_baseline_inputs}
\begin{tabularx}{\textwidth}{@{}p{0.24\textwidth}X X X@{}}
\toprule
\textbf{Condition} & \textbf{Information given to model} & \textbf{Output and repair control} & \textbf{Purpose} \\
\midrule
Profile/template baseline scaffold & Task prompt, raw data access, $\Phi$ summary, table/field dictionary, join keys, temporal-role candidates, and artifact template. & One initial generation; no validator feedback loop beyond final scoring. & Controls for missing schema and output-format information. \\
Profile/template/repair baseline scaffold & Same as above. & Three edit-run-repair turns with standardized validator feedback and identical execution budget. & Controls for unequal repair opportunity. \\
Direct Gemini-3-Pro + validator interface & Gemini-3-Pro receives $\Phi$, artifact template, sandbox logs, and validator feedback. & Three repair turns; no typed IR compiler or certificate admission gate. & Controls for deterministic validation feedback without ProCARE planning. \\
Direct Gemini-3-Pro + study-spec schema & Gemini-3-Pro receives $\Phi$ and must emit \texttt{study\_spec.json} before writing code and report. & No ProCARE IR compiler; validator feedback is limited to schema/artifact execution errors. & Controls for evaluator-compatible study-spec output. \\
Direct Gemini-3-Pro + claim-trace schema & Gemini-3-Pro receives $\Phi$, artifact template, and claim-trace file schema. & No certificate admission gate or certified DAG ordering. & Controls for evaluator-compatible claim-trace output. \\
ProCARE reference & Full profile, IR compiler, violation-guided execution, evidence certificates, and certified report DAG. & Same three repair turns and final checker. & Reports the integrated contract workflow for comparison. \\
\bottomrule
\end{tabularx}
\end{table*}

Table~\ref{tab:scaffolded_baseline_full} reports the final checker outputs for the scaffolded-baseline evaluation. Scaffolded baselines improve substantially over generic agents, reducing N/A rates and narrowing the execution gap, while the integrated ProCARE workflow remains strongest on violation resolution and claim-level traceability. The table should be read as a fairness-control comparison, not as evidence that individual ProCARE modules have been added to an otherwise unchanged LLM. All direct rows use Gemini-3-Pro.

\begin{table*}[t]
\centering
\footnotesize
\setlength{\tabcolsep}{2.5pt}
\renewcommand{\arraystretch}{1.12}
\caption{Full metric results for fairness-controlled scaffolded baselines. Higher is better except UCR and CLV. Rows are baseline adaptations, not ProCARE ablations; values are final checker outputs.}
\label{tab:scaffolded_baseline_full}
\begin{tabularx}{\textwidth}{@{}>{\RaggedRight\arraybackslash}X *{12}{c}@{}}
\toprule
\textbf{Baseline condition} & \multicolumn{8}{c}{\textbf{Performance}} & \multicolumn{4}{c}{\textbf{Contract diagnostics}} \\
\cmidrule(lr){2-9} \cmidrule(lr){10-13}
& {\textbf{GA}$\uparrow$} & {\textbf{PF}$\uparrow$} & {\textbf{XE}$\uparrow$} & {\textbf{CQ}$\uparrow$} & {\textbf{VC}$\uparrow$} & {\textbf{CF}$\uparrow$} & {\textbf{RS}$\uparrow$} & {\textbf{SL}$\uparrow$}
& {\textbf{VRR}$\uparrow$} & {\textbf{CTC}$\uparrow$} & {\textbf{UCR}$\downarrow$} & {\textbf{CLV}$\downarrow$} \\
\midrule
ReAct + fairness scaffold & 0.82 & 0.79 & 0.74 & 0.69 & 0.62 & 0.65 & 0.75 & 0.70 & 0.62 & 0.70 & 0.13 & 0.10 \\
ToolLLM + fairness scaffold & 0.79 & 0.80 & 0.68 & 0.65 & 0.58 & 0.62 & 0.72 & 0.67 & 0.58 & 0.66 & 0.15 & 0.12 \\
Direct Gemini-3-Pro + profile/template & 0.77 & 0.73 & 0.66 & 0.62 & 0.54 & 0.58 & 0.73 & 0.65 & 0.31 & 0.55 & 0.20 & 0.15 \\
Direct Gemini-3-Pro + validator repair & 0.80 & 0.78 & 0.79 & 0.72 & 0.60 & 0.64 & 0.75 & 0.69 & 0.68 & 0.65 & 0.15 & 0.11 \\
Direct Gemini-3-Pro + study-spec schema & 0.82 & 0.82 & 0.70 & 0.68 & 0.57 & 0.62 & 0.71 & 0.68 & 0.43 & 0.52 & 0.21 & 0.16 \\
Direct Gemini-3-Pro + claim-trace schema & 0.78 & 0.74 & 0.63 & 0.61 & 0.59 & 0.63 & 0.74 & 0.73 & 0.35 & 0.74 & 0.11 & 0.08 \\
\midrule
\textbf{ProCARE reference} & \textbf{0.86} & \textbf{0.84} & \textbf{0.92} & \textbf{0.88} & \textbf{0.72} & \textbf{0.74} & \textbf{0.80} & \textbf{0.79} & \textbf{0.81} & \textbf{0.90} & \textbf{0.06} & \textbf{0.03} \\
\bottomrule
\end{tabularx}
\end{table*}

\subsection{Ablation protocol}
\label{app:ablation_protocol}
Mechanism ablations remove one contract component at a time while keeping all remaining components fixed whenever the remaining workflow is defined. Table~\ref{tab:ablation_protocol} specifies the intervention behind the ablation table in the main paper.

\begin{table*}[t]
\centering
\scriptsize
\setlength{\tabcolsep}{4pt}
\renewcommand{\arraystretch}{1.08}
\caption{Ablation definitions for contract mechanisms.}
\label{tab:ablation_protocol}
\begin{tabularx}{\textwidth}{@{}p{0.20\textwidth}X X@{}}
\toprule
\textbf{Variant} & \textbf{Removed component} & \textbf{What remains fixed} \\
\midrule
w/o profile $\Phi$ & The model does not receive the heterogeneous profile as a contract object. It may inspect raw table metadata and generated summaries, but generation lacks profile-grounded repair hints. & Prompt, budget, execution sandbox, reporting request, and post-hoc evaluator. \\
w/o RWS-IR & The model writes code and a report without producing the typed RWS-IR. & Raw data access, profile summaries when available, execution sandbox, and post-hoc claim diagnostics. PF is N/A because no typed IR exists. \\
w/o violation-guided repair & Runtime validators detect violations, but localized violation sets are not fed back into the next edit. & Profile, RWS-IR, execution sandbox, artifact validators, certificates, and report policy. \\
w/o evidence certificates & The report is generated without explicit claim certificates and without the claim-admission gate. & Profile, RWS-IR, execution repair, artifacts, and DAG section ordering. CTC for emitted explicit certificates is zero; post-hoc support can still be inspected separately. \\
w/o certified DAG ordering & Evidence certificates remain available, but report sections are generated in a fixed or unconstrained order without dependency-aware traversal. & Profile, RWS-IR, execution repair, artifact validators, and claim admission. \\
Full ProCARE & No contract component is removed. & Complete profile--IR--repair--certificate--DAG workflow. \\
\bottomrule
\end{tabularx}
\end{table*}

\subsection{Model access, decoding, and version identifiers}
\label{app:model_access}
Backbone-sensitivity experiments keep profiling, RWS-IR compilation, violation-guided execution, certified reporting, prompts, and scoring scripts fixed. The table records the paper-level model identifiers and roles used in the experiments. Provider-returned request metadata is retained in run logs; when a provider alias differs from the paper-level display name, the logged identifier is the reproducibility target. Generation runs use the released implementation's production decoding settings, with temperature 0.2 for Gemini/openai-compatible providers and 0.1 for DeepSeek-compatible providers; judge models are run at temperature 0.

\begin{table*}[t]
\centering
\scriptsize
\setlength{\tabcolsep}{3pt}
\renewcommand{\arraystretch}{1.08}
\caption{Model access, version metadata, and evaluation roles.}
\label{tab:model_access_metadata}
\begin{tabularx}{\textwidth}{@{}p{0.14\textwidth}p{0.22\textwidth}p{0.18\textwidth}X@{}}
\toprule
\textbf{Role} & \textbf{Paper-level model identifier} & \textbf{Access mode} & \textbf{Use in evaluation} \\
\midrule
Backbone & GPT-5.2 & Proprietary API & Backbone-sensitivity run under the fixed ProCARE workflow. \\
Backbone & GPT-4 Turbo & Proprietary API & Backbone-sensitivity run under the fixed ProCARE workflow. \\
Backbone & Claude Opus 4.5 & Proprietary API & Backbone-sensitivity run under the fixed ProCARE workflow. \\
Backbone & Claude Sonnet 4.5 & Proprietary API & Backbone-sensitivity run under the fixed ProCARE workflow. \\
Backbone & Claude-3 Opus & Proprietary API & Backbone-sensitivity run under the fixed ProCARE workflow. \\
Backbone, direct baseline, and scaffolded control & Gemini-3-Pro & Proprietary API & Default ProCARE backbone; also used for the direct baseline and all Direct Gemini-3-Pro scaffolded fairness-control rows. \\
Backbone and judge & Gemini-2.5-Pro & Proprietary API & Backbone-sensitivity run and one blinded semantic judge. \\
Backbone & Qwen3-Max & API endpoint & Backbone-sensitivity run under the fixed ProCARE workflow. \\
Backbone & \makecell[l]{\texttt{Qwen3-235B-A22B-}\\\texttt{Instruct-2507}} & API endpoint or hosted open-source model & Backbone-sensitivity run. \\
Backbone & DeepSeek-V3.2 & API endpoint or hosted open-source model & Backbone-sensitivity run. \\
Backbone & DeepSeek-R1-0528 & API endpoint or hosted open-source model & Backbone-sensitivity run and non-valid-bundle baseline category. \\
Code-agent baseline & Claude Code (Claude Sonnet 4.5 backend) & Proprietary coding-agent interface & Baseline run through the native Claude Code repository-editing workflow; provider-returned CLI/model metadata is retained in the run log. \\
Code-agent baseline & Codex (\texttt{gpt-5.3-codex} backend) & Proprietary coding-agent interface & Baseline run through the native Codex repository-editing workflow; CLI session metadata and final workspace state are retained for reproduction. \\
Judge & Qwen2.5-1M & API endpoint & Second blinded semantic judge. \\
\bottomrule
\end{tabularx}
\end{table*}

\subsection{Reproducibility package}
\label{app:repo_commands}
The anonymous release contains the manuscript sources, system code, prompt definitions, the fully de-identified HCC benchmark tables, Public-RWS bundle metadata, profile artifacts, evaluation scripts, and figure-generation scripts. The HCC release includes the record-level tables needed to rerun the 36 primary tasks, with direct identifiers removed and random benchmark IDs retained for reproducible linkage.

\begin{table*}[t]
\centering
\scriptsize
\setlength{\tabcolsep}{4pt}
\renewcommand{\arraystretch}{1.08}
\caption{Repository components used for reproduction.}
\label{tab:repo_layout}
\begin{tabularx}{\textwidth}{@{}p{0.25\textwidth}X@{}}
\toprule
\textbf{Path} & \textbf{Contents} \\
\midrule
\texttt{paper/} & Manuscript, supplement, bibliography, NeurIPS style file, build script, and paper figures. \\
\texttt{code/backend/} & Profiling, planning, code-generation, execution, report-generation, and API services. \\
\texttt{code/backend/tests/} & Local tests for profile construction, planning, code execution, OpenCode integration, and report generation. \\
\texttt{code/frontend/} & Interface for project setup, profile inspection, RWS-IR review, execution monitoring, and report review. \\
\path{code/output/knowledge_base/profiles/} & Cached profile artifacts used by the implementation. \\
\path{rws_hcc_benchmark/} & Fully de-identified HCC benchmark tables, data dictionary, checksums, task metadata, and release terms for reproducing the 36 primary tasks. \\
\path{rws_public_datasets_bundle/} & Public-RWS source manifests, checksums, table shapes, and public dataset files. \\
\texttt{figure\_code/} & Scripts used to regenerate paper diagnostic figures. \\
\bottomrule
\end{tabularx}
\end{table*}

\begin{lstlisting}[basicstyle=\ttfamily\footnotesize,breaklines=true,breakatwhitespace=true,columns=fullflexible,caption={Local reproduction commands.},label={lst:repro_commands}]
cd code/backend
python -m app.cli --excel <workbook.xlsx> --topic "<benchmark prompt>" --refine
python tests/run_all.py

cd ../../paper
build.cmd nips appendix
\end{lstlisting}

\section{Additional Results and Diagnostics}
\label{app:additional_results}

\subsection{Claim-diagnostic interpretation}
\label{app:claim_metrics}
More specific prompts increase claim traceability because they constrain variables, temporal anchors, methods, and expected artifacts before reporting. The dominant downgrades are language-policy failures rather than arithmetic inconsistencies: causal phrasing from descriptive analyses, safety claims based only on screening pages, or subgroup conclusions without sufficient support. During certification, these claims are rewritten as limitations or removed from the admitted factual-claim set.

\subsection{Backbone-sensitivity interpretation}
\label{app:backbone_interpretation}
Backbone substitution changes only the base LLM used for planning, code editing, and report drafting. Profiling, RWS-IR schema, validators, evidence-certificate admission, report schema, and evaluation remain fixed. Stronger backbones improve GA, PF, VRR, and CTC because they use profile constraints and repair feedback more reliably. Execution-only differences are smaller than planning and reporting differences because deterministic validation and localized repair bound the accepted artifact space after code is generated.

\section{System Interface and Workflow}
\label{app:system_interface}
\begin{figure*}[t]
    \centering
    \begin{subfigure}[b]{0.48\textwidth}
        \centering
        \includegraphics[width=\linewidth]{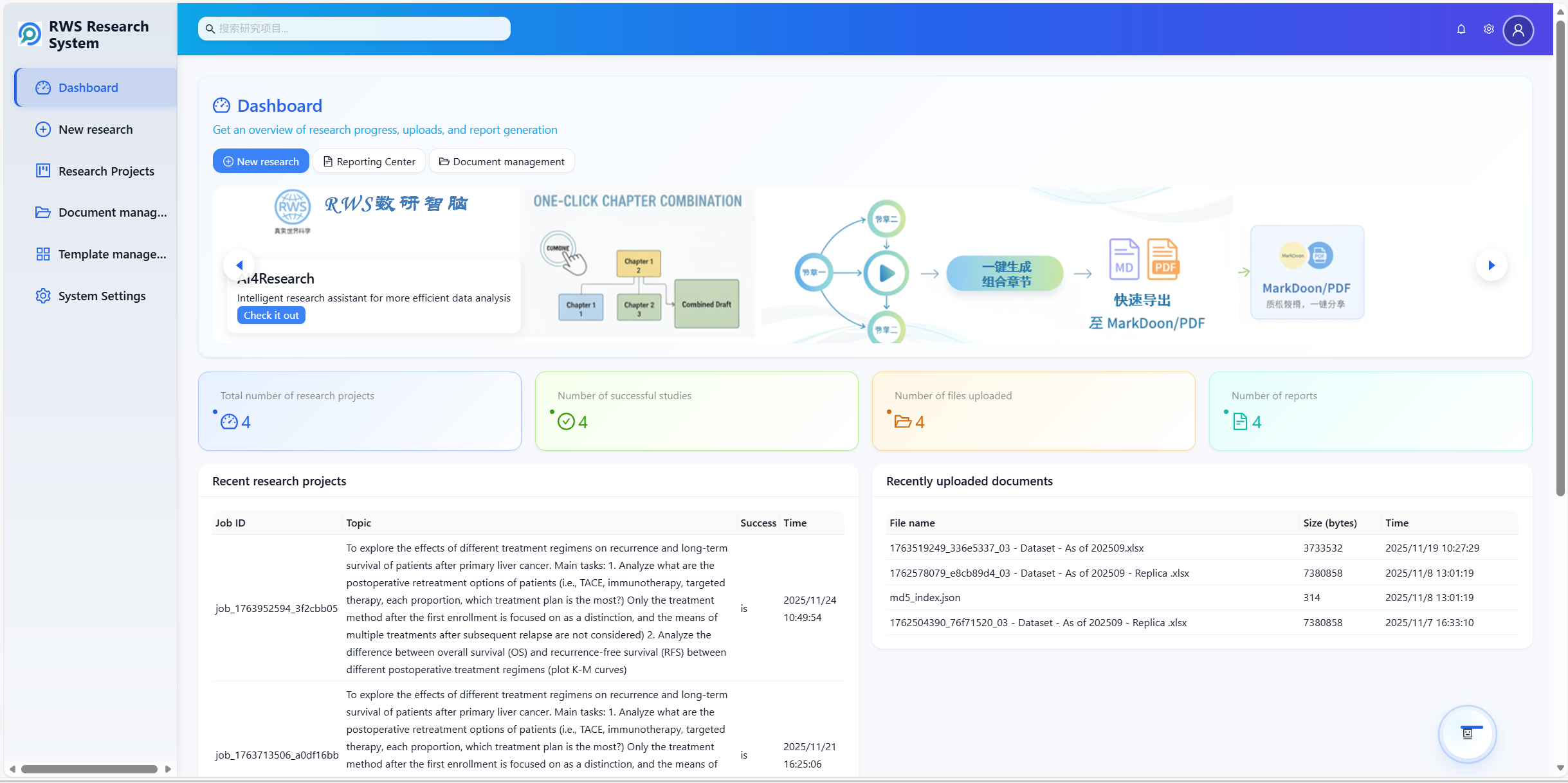}
        \caption{System dashboard for managing RWS projects and cohort status.}
        \label{fig:sys_dashboard}
    \end{subfigure}
    \hfill
    \begin{subfigure}[b]{0.48\textwidth}
        \centering
        \includegraphics[width=\linewidth]{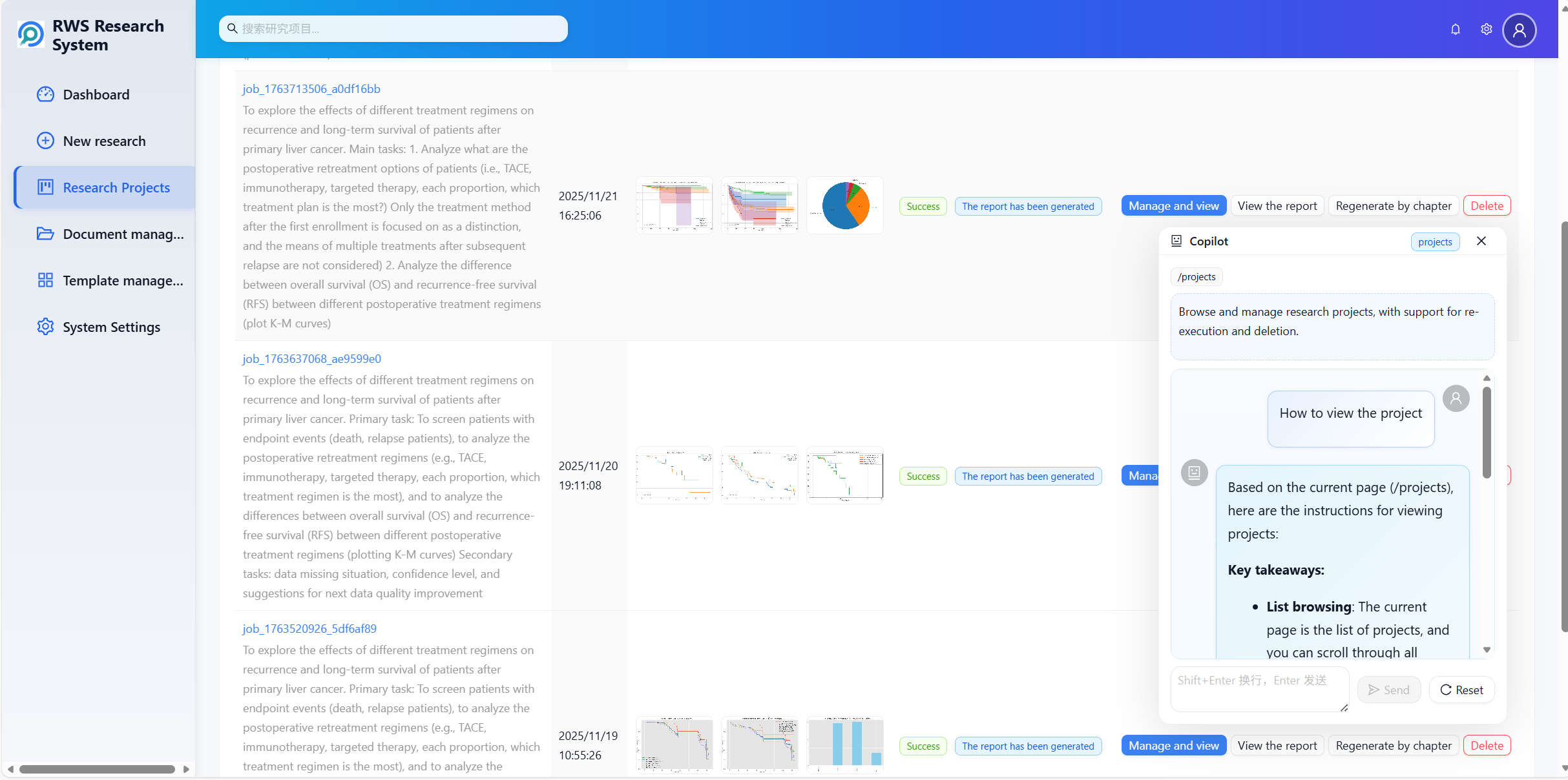}
        \caption{Profile panel for reviewing structural metadata ($\Phi^{S}$) and ontology mappings ($\Phi^{K}$).}
        \label{fig:sys_panel}
    \end{subfigure}
    
    \vspace{0.8em}
    
    \begin{subfigure}[b]{0.48\textwidth}
        \centering
        \includegraphics[width=\linewidth]{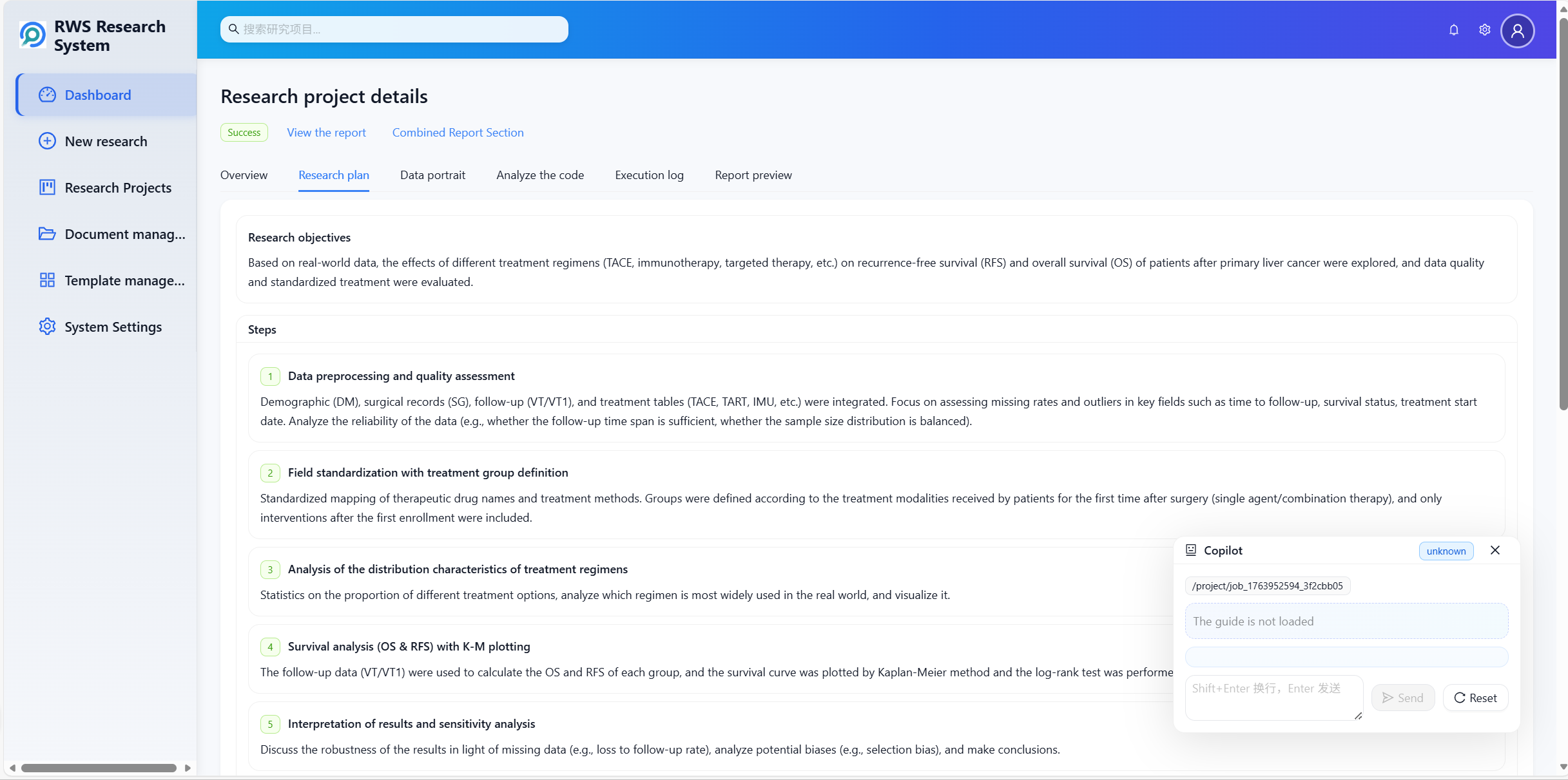}
        \caption{RWS-IR view showing goal compilation ($T_{\mathrm{goal}} \rightarrow Z$).}
        \label{fig:sys_plan}
    \end{subfigure}
    \hfill
    \begin{subfigure}[b]{0.48\textwidth}
        \centering
        \includegraphics[width=\linewidth]{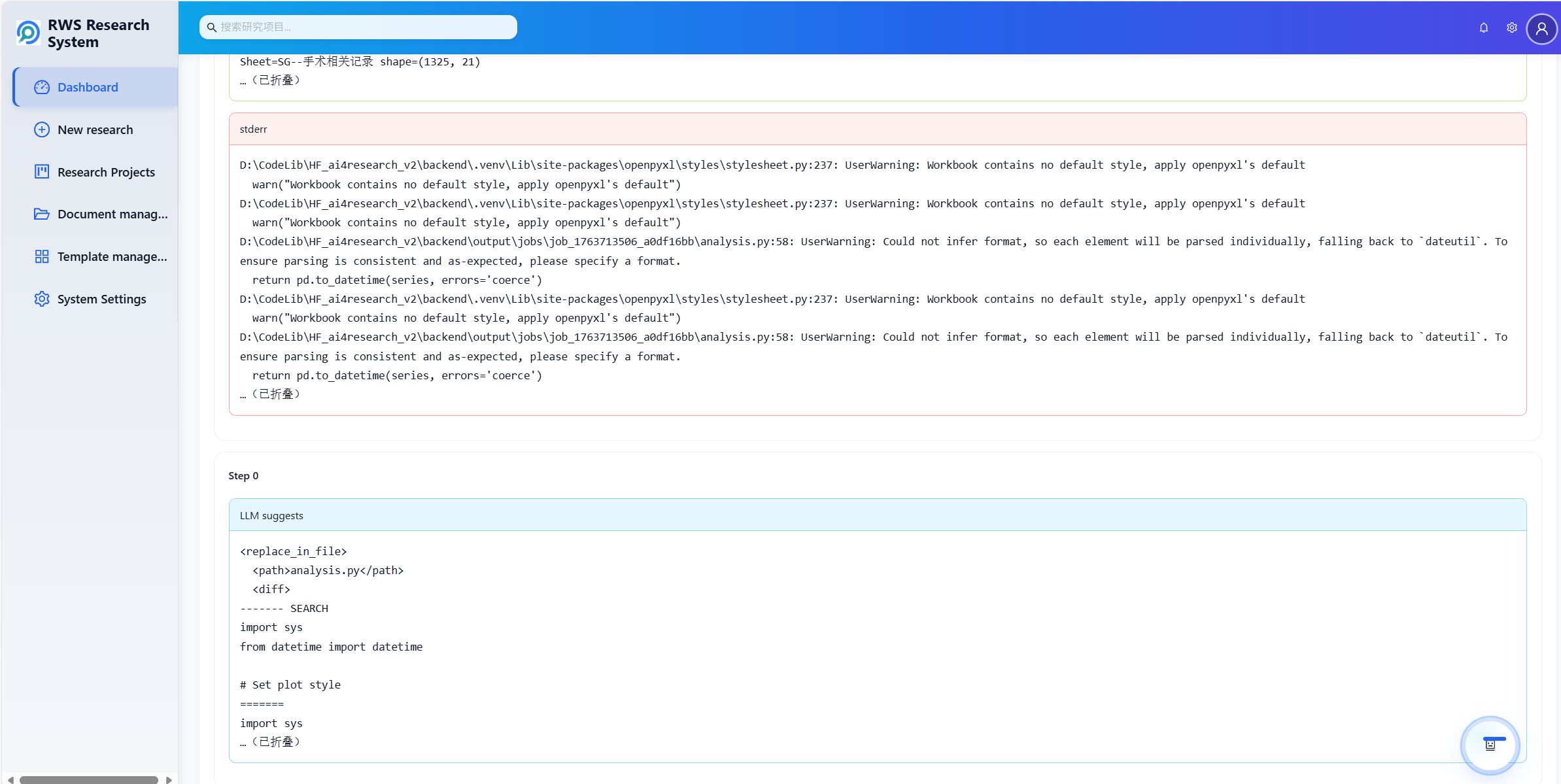}
        \caption{Coding harness view with reasoning traces, sandbox logs, and verification status.}
        \label{fig:sys_coding}
    \end{subfigure}
    
    \caption{ProCARE interface states for project setup, profile inspection, RWS-IR compilation, and coding execution.}
    \label{fig:system_overview}
\end{figure*}

\begin{figure*}[t]
    \centering
    \begin{subfigure}[b]{0.48\textwidth}
        \centering
        \includegraphics[width=\linewidth]{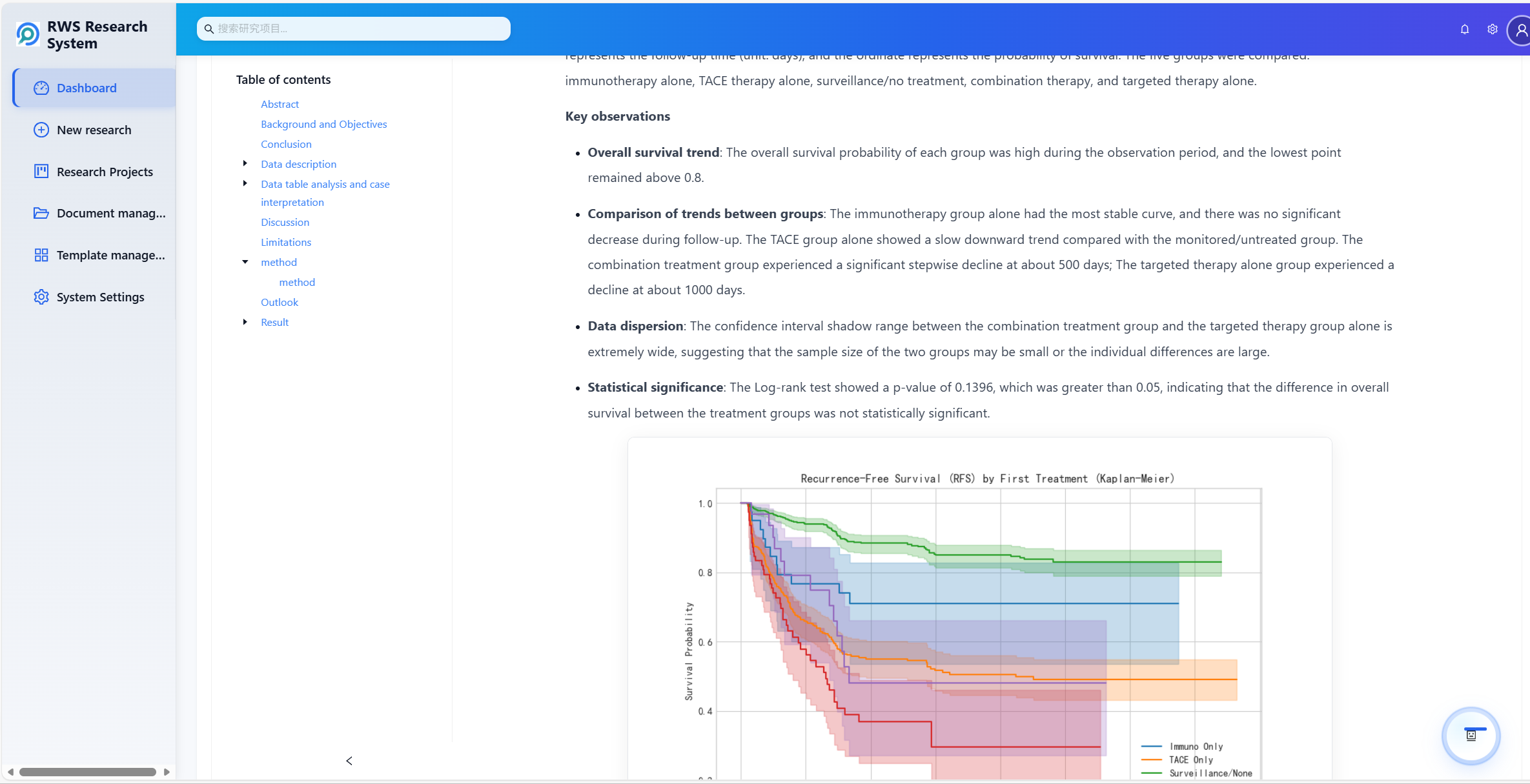}
        \caption{Report review view with generated tables, charts, and evidence links.}
        \label{fig:sys_report_review}
    \end{subfigure}
    \hfill
    \begin{subfigure}[b]{0.48\textwidth}
        \centering
        \includegraphics[width=\linewidth]{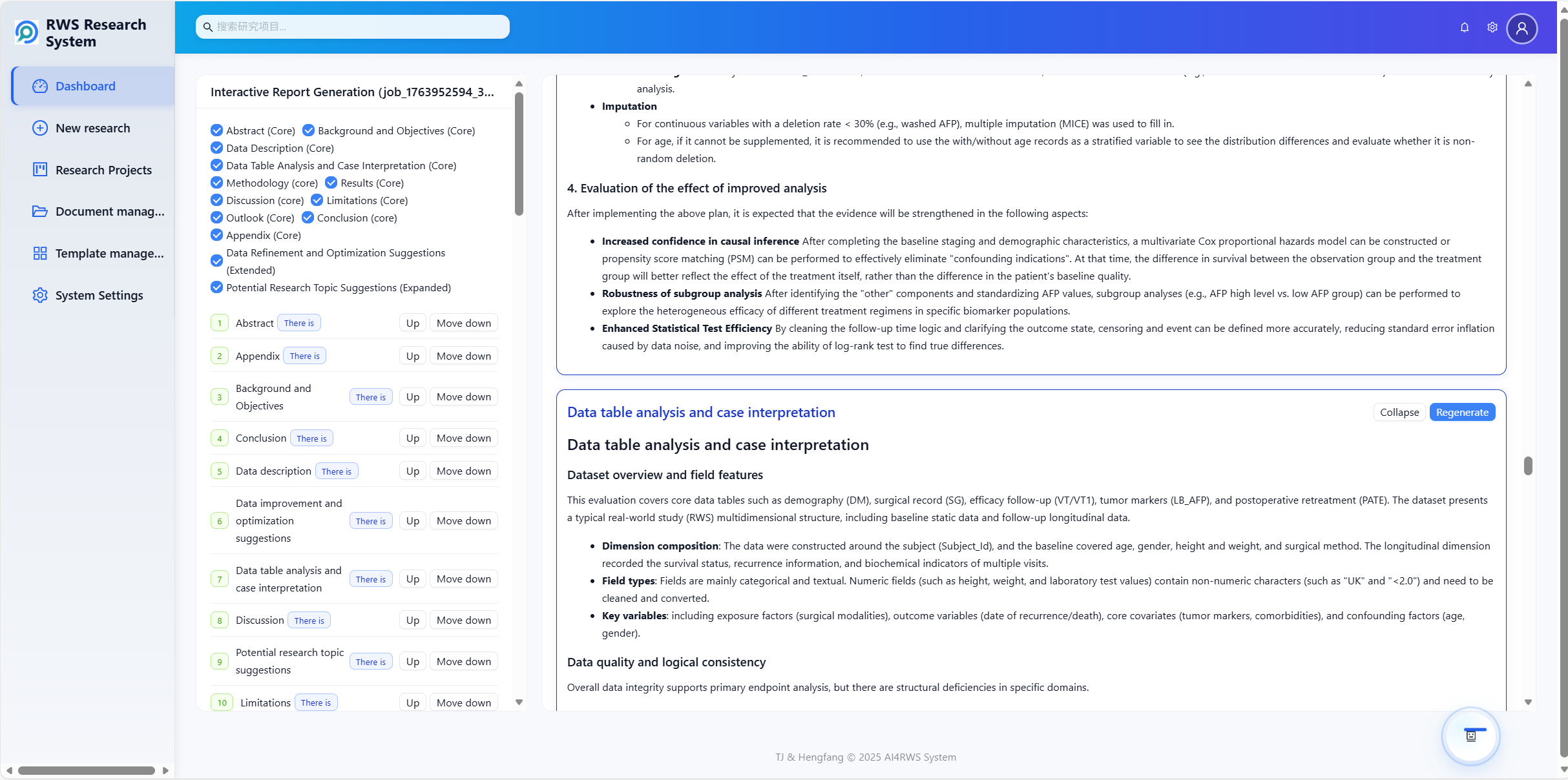}
        \caption{Section-wise report generation with data-bound report blocks.}
        \label{fig:sys_report_gen}
    \end{subfigure}

    \caption{Report-facing interface states for reviewing generated artifacts, evidence links, and section-wise report content.}
    \label{fig:system_reporting_views}
\end{figure*}

Figures~\ref{fig:system_overview} and~\ref{fig:system_reporting_views} show the implementation interface states. They are included as workflow evidence rather than method exposition: the panels identify where profile artifacts, IR traces, execution logs, generated figures, and section-level report blocks are inspected during a run.

\section{Evaluation Protocol and Rubrics}
\label{app:rubrics}

Evaluation uses deterministic validators for feasibility and execution checks, and LLM judges for semantic quality.

\subsection{Metrics Overview}
Table~\ref{tab:appendix-metrics} defines the metrics used for end-to-end RWS automation. All scores are normalized to $[0,1]$.

\begin{table*}[t]
\centering
\small
\setlength{\tabcolsep}{4pt}
\renewcommand{\arraystretch}{1.08}
\caption{Metrics used in the appendix evaluation protocol.}
\label{tab:appendix-metrics}
\begin{tabularx}{\textwidth}{@{}l l X@{}}
\toprule
\textbf{Metric} & \textbf{Type} & \textbf{Scoring rule} \\
\midrule
RWS-IR Feasibility (PF) & Deterministic + judge & Validates the RWS-IR against $\Phi^S$ and penalizes non-existent tables, columns, invalid joins, unsupported temporal anchors, or incompatible analysis grains. Residual ambiguity is checked by the judge. \\
Code Executability (XE) & Deterministic & Sandbox execution success rate, $n_{\mathrm{pass}}/N$; runtime errors and invalid output files count as failures. \\
Violation Resolution Rate (VRR) & Deterministic & Fraction of detected execution-time contract violations resolved before the accepted run. \\
Report Structure (RS) & Deterministic & Fraction of required report sections in Table~\ref{tab:report-schema} covered with non-trivial content. \\
Claim Trace Coverage (CTC) & Deterministic & Fraction of final factual claims linked to valid evidence certificates. \\
Unsupported Claim Rate (UCR) & Deterministic & Fraction of final factual claims without admissible evidence support. \\
Claim Language Violation (CLV) & Deterministic + judge & Fraction of factual claims whose wording exceeds the support level allowed by the evidence certificate and RWS-IR claim policy. \\
Goal Alignment (GA) & Judge & Alignment with the clinical objective, target population, and requested endpoint. \\
Code Quality (CQ) & Judge & Soundness of data handling, missingness handling, joins, and statistical implementation. \\
Visual Clarity (VC) & Judge & Readability and labeling quality of generated figures. \\
Chart Fidelity (CF) & Judge & Consistency between plotted values, execution logs, and report statements. \\
Scientific Language (SL) & Judge & Clinical tone, hedging, and limitation disclosure. \\
\bottomrule
\end{tabularx}
\end{table*}

\subsection{Deterministic validators}
\label{app:deterministic_validators}
Deterministic validators are used whenever the property can be checked from the artifact bundle. PF checks whether the accepted RWS-IR references only profiled tables, fields, keys, clinical roles, temporal anchors, and methods. XE checks whether the submitted entry point exits successfully, avoids failure sentinels, and writes the required result and figure manifests. Artifact completeness requires declared output files to exist, be nonempty, and be referenced by the corresponding manifest. Temporal consistency checks row-level ordering between index, exposure, event, and censoring dates when those roles are declared in $Z$. Chart-result consistency checks whether plotted variables, group labels, sample sizes, and reported summary values match execution logs or result tables.

All metrics are normalized to $[0,1]$. XE is $n_{\mathrm{pass}}/N$ over tasks. PF is one for a statically admissible RWS-IR, zero for a hard-invalid RWS-IR, and partially credited only for residual semantic ambiguity adjudicated by the fixed rubric. RS is the fraction of required report sections with task-relevant content. VRR, CTC, UCR, and CLV are computed from logged violations and extracted factual claims as described in Appendix~\ref{app:execution_validators} and Appendix~\ref{app:claim_admission}.

For baselines that do not implement ProCARE's contract internally, the evaluator reconstructs the same diagnostics from final artifacts. A baseline can receive CTC credit only when a factual claim can be linked to an executed result, code region, profile-supported field, or explicit assumption. Missing artifacts or unsupported variables block traceability.

\subsection{LLM judge protocol}
We use \emph{Gemini-2.5-Pro} and \emph{Qwen2.5-1M} as independent judges with temperature 0. Judges are blinded to method identity and receive the same artifact bundle: input prompt, generated RWS-IR, code, execution logs, result and figure manifests, evidence certificates when available, assumptions, limitations, and final report. Scores are normalized to $[0,1]$ and averaged across judges. Schema hallucination and artifact inconsistency are handled as hard rubric failures for the affected dimensions.

\subsection{Expert-review protocol}
\label{app:expert_review}
Expert review is performed on the 36 primary HCC benchmark instances. The panel contains five blinded reviewers: three board-certified clinical experts with hepatobiliary oncology or hepatology experience (C1--C3) and two statistics/data experts with clinical data-analysis experience (S1--S2). Reviewers receive anonymized artifact bundles containing the task prompt, study specification, code or execution logs when relevant, figures, assumptions, limitations, evidence certificates when available, and final report. Method identity is masked in the review package.

Reviewers score each stage on a five-point Likert scale: 1 indicates an invalid or irrelevant artifact, 3 indicates a partially correct artifact with important limitations, and 5 indicates a clinically and methodologically appropriate artifact with limitations disclosed. The clinical panel focuses on clinical objective alignment, endpoint and covariate validity, clinical interpretability of figures, and claim scope. The statistics/data panel focuses on data linkage, analysis grain, statistical method choice, code reproducibility, figure-result consistency, and whether reported results are supported by executed artifacts. Code scores in the main table use the statistics/data panel only; the other stage scores aggregate all five reviewers because both clinical and statistical/data judgments are relevant.

The raw evaluation data are represented in the long format in Table~\ref{tab:expert_raw_form}. The required stage-level row set contains $36\times(5+2+5+5)=612$ reviewer--instance--stage records before optional criterion-level comments: five reviewers score RWS-IR, visualization, and reporting, while S1--S2 score code implementation. Additional criterion-specific rows can be added with the same keys when a reviewer records a subscore or a major error flag.

\begin{table*}[t]
\centering
\scriptsize
\setlength{\tabcolsep}{4pt}
\renewcommand{\arraystretch}{1.08}
\caption{Raw expert-rating data format used before aggregation. One row stores one reviewer--instance--stage score.}
\label{tab:expert_raw_form}
\begin{tabularx}{\textwidth}{@{}p{0.18\textwidth}p{0.24\textwidth}X@{}}
\toprule
\textbf{Column} & \textbf{Allowed values or format} & \textbf{Definition} \\
\midrule
\texttt{review\_id} & stable string & Unique row identifier. \\
\texttt{benchmark\_id} & HCC-001--HCC-036 & Primary benchmark instance identifier. \\
\texttt{objective} & cohort, safety-proxy, efficacy, resource-use & Benchmark objective family. \\
\texttt{prompt\_level} & goal, goal+approach, goal+method & Prompt-specificity level. \\
\texttt{method\_id} & submitted method identifier & Method label masked during review and restored only for analysis. \\
\texttt{stage} & rws\_ir, code, visualization, reporting & Workflow stage being scored. \\
\texttt{artifact\_id} & path or hash & Submitted artifact bundle, code file, figure set, or report identifier. \\
\texttt{reviewer\_id} & C1, C2, C3, S1, S2 & De-identified reviewer identifier. \\
\texttt{reviewer\_role} & clinical or statistics\_data & Reviewer expertise group. \\
\texttt{criterion\_id} & global\_stage\_score or rubric subcriterion & Overall stage score or optional subcriterion. \\
\texttt{score\_1\_5} & integer 1--5 & Raw Likert score before normalization. \\
\texttt{major\_error\_flag} & 0 or 1 & Reviewer-marked hard error, such as invalid endpoint, impossible join, or irreproducible code. \\
\texttt{claim\_flag} & 0, 1, or NA & Whether the reviewed report or figure interpretation contains a claim exceeding evidence support. \\
\texttt{comment\_code} & none, schema, method, temporal, claim, figure, reproducibility, other & Structured reason for downgrade, if any. \\
\texttt{free\_text\_comment} & de-identified text & Short reviewer note used for adjudication and error analysis. \\
\texttt{timestamp\_utc} & ISO-8601 timestamp & Completion time of the review row. \\
\bottomrule
\end{tabularx}
\end{table*}

For each benchmark instance $i$, reviewer $r$, and stage $s$, the raw Likert score is $x_{irs}\in\{1,\ldots,5\}$. Stage-level expert scores are first averaged across the eligible reviewer set $R_s$:
\[
E_{is}=\frac{1}{|R_s|}\sum_{r\in R_s}x_{irs}.
\]
Scores are normalized to $[0,1]$ for calibration plots by $\tilde{E}_{is}=(E_{is}-1)/4$, but the main paper reports mean$\pm$SD on the original 1--5 scale. Spearman $\rho$ is computed between per-instance expert scores and the corresponding automated stage score: $(\mathrm{GA}+\mathrm{PF})/2$ for RWS-IR, CQ for code, $(\mathrm{VC}+\mathrm{CF})/2$ for visualization, and SL for reporting. Confidence intervals for mean expert scores use a $t$ interval over 36 instance-level means; confidence intervals for Spearman correlations use Fisher transformation.

\begin{table*}[t]
\centering
\scriptsize
\setlength{\tabcolsep}{4pt}
\renewcommand{\arraystretch}{1.08}
\caption{Expert panel responsibilities and raw scoring volume.}
\label{tab:expert_panel_roles}
\begin{tabularx}{\textwidth}{@{}p{0.16\textwidth}p{0.18\textwidth}X c@{}}
\toprule
\textbf{Stage} & \textbf{Eligible reviewers} & \textbf{Primary review focus} & \textbf{Raw records} \\
\midrule
RWS-IR & C1--C3, S1--S2 & Clinical goal, endpoint validity, data linkage, analysis grain, and statistical design feasibility. & 180 \\
Code & S1--S2 & Reproducibility, executable implementation, statistical method use, artifact writing, and result consistency. & 72 \\
Visualization & C1--C3, S1--S2 & Clinical interpretability, sample-size labeling, figure-result agreement, and avoidance of unsupported visual claims. & 180 \\
Reporting & C1--C3, S1--S2 & Clinical language, limitation disclosure, statistical wording, and claim-to-evidence support. & 180 \\
\bottomrule
\end{tabularx}
\end{table*}

\begin{table*}[t]
\centering
\scriptsize
\setlength{\tabcolsep}{4pt}
\renewcommand{\arraystretch}{1.08}
\caption{Stage-level expert score summary on the 36 primary HCC benchmark instances. CI is the 95\% confidence interval for the instance-level mean.}
\label{tab:expert_score_summary}
\begin{tabularx}{\textwidth}{@{}p{0.15\textwidth}c c c c c@{}}
\toprule
\textbf{Stage} & \textbf{Instance means} & \textbf{Raw records} & \textbf{Mean$\pm$SD} & \textbf{95\% CI} & \textbf{Median [IQR]} \\
\midrule
RWS-IR & 36 & 180 & 4.38$\pm$0.48 & 4.22--4.54 & 4.33 [4.00, 4.67] \\
Code & 36 & 72 & 4.42$\pm$0.21 & 4.35--4.49 & 4.42 [4.25, 4.58] \\
Visualization & 36 & 180 & 4.50$\pm$0.53 & 4.32--4.68 & 4.67 [4.00, 5.00] \\
Reporting & 36 & 180 & 4.13$\pm$0.64 & 3.91--4.35 & 4.17 [3.67, 4.67] \\
\bottomrule
\end{tabularx}
\end{table*}

\begin{table*}[t]
\centering
\scriptsize
\setlength{\tabcolsep}{4pt}
\renewcommand{\arraystretch}{1.08}
\caption{Inter-rater reliability for the human expert panel. Stages scored by five reviewers use two-way random-effects ICC for averaged ratings; the two-reviewer code stage uses weighted Cohen's $\kappa_w$.}
\label{tab:expert_reliability}
\begin{tabularx}{\textwidth}{@{}p{0.15\textwidth}p{0.18\textwidth}p{0.18\textwidth}c c X@{}}
\toprule
\textbf{Stage} & \textbf{Panel} & \textbf{Statistic} & \textbf{Estimate} & \textbf{95\% CI} & \textbf{Interpretation} \\
\midrule
RWS-IR & C1--C3, S1--S2 & ICC(2,k) & 0.84 & 0.72--0.91 & Strong agreement on specification validity. \\
Code & S1--S2 & $\kappa_w$ & 0.76 & 0.61--0.88 & Substantial agreement on reproducibility and statistical implementation. \\
Visualization & C1--C3, S1--S2 & ICC(2,k) & 0.79 & 0.66--0.88 & Strong agreement on figure clarity and consistency. \\
Reporting & C1--C3, S1--S2 & ICC(2,k) & 0.77 & 0.63--0.87 & Substantial-to-strong agreement on language and claim scope. \\
\bottomrule
\end{tabularx}
\end{table*}

\begin{table*}[t]
\centering
\scriptsize
\setlength{\tabcolsep}{4pt}
\renewcommand{\arraystretch}{1.08}
\caption{Correlation between automated metrics and aggregated expert scores. $q$ is the Benjamini--Hochberg adjusted $p$ value across the four stage-wise correlations.}
\label{tab:expert_auto_stats}
\begin{tabularx}{\textwidth}{@{}p{0.15\textwidth}p{0.20\textwidth}p{0.18\textwidth}c c c c@{}}
\toprule
\textbf{Stage} & \textbf{Expert aggregation} & \textbf{Automated score} & \textbf{$\rho$} & \textbf{95\% CI} & \textbf{$p$} & \textbf{$q$} \\
\midrule
RWS-IR & Mean of C1--C3, S1--S2 & $(GA+PF)/2$ & 0.82 & 0.67--0.90 & $<0.001$ & $<0.001$ \\
Code & Mean of S1--S2 & $CQ$ & 0.74 & 0.54--0.86 & $<0.001$ & $<0.001$ \\
Visualization & Mean of C1--C3, S1--S2 & $(VC+CF)/2$ & 0.75 & 0.55--0.86 & $<0.001$ & $<0.001$ \\
Reporting & Mean of C1--C3, S1--S2 & $SL$ & 0.71 & 0.49--0.84 & $<0.001$ & $<0.001$ \\
\bottomrule
\end{tabularx}
\end{table*}

\subsection{Semantic Rubric Anchors}
Judges assign 1--5 Likert scores for Goal Alignment (GA), Code Quality (CQ), Visual Clarity (VC), and Scientific Language (SL). Scores are normalized to $[0,1]$ with a linear transformation. Table~\ref{tab:rubric_anchors} lists the anchor definitions.

\begin{table*}[t]
\centering
\caption{Semantic rubric anchors for LLM judges.}
\label{tab:rubric_anchors}
\small
\begin{tabularx}{\textwidth}{@{}l X X X@{}}
\toprule
\textbf{Metric} & \textbf{Score 5} & \textbf{Score 3} & \textbf{Score 1} \\
\midrule
\textbf{Goal Alignment} & Addresses the clinical objective, target population, and endpoints with explicit limitation disclosure. & Partially addresses the task with minor errors in inclusion criteria or endpoints. & Gives an irrelevant analysis or misunderstands the clinical goal. \\
\midrule
\textbf{Code Quality} & Uses sound pandas or SQL logic, handles missing data and types correctly, and keeps the implementation modular. & Runs but remains fragile, often because of hardcoded values or inefficient joins. & Uses an incorrect method, applies the wrong statistical test, or contradicts the plan. \\
\midrule
\textbf{Visual Clarity} & Produces figures with clear axes, legends, labels, and titles that support the reported results. & Produces legible charts with missing units, overlapping labels, or default styling. & Produces misleading, illegible, or unrelated visualizations. \\
\midrule
\textbf{Scientific Language} & Uses clinical language with appropriate hedging and clear limitation disclosure. & Uses a general scientific tone but has verbose phrasing or imprecise medical terminology. & Uses casual language or makes unsupported claims. \\
\bottomrule
\end{tabularx}
\end{table*}

\section{Prompts and Templates}
\label{app:prompts}

\subsection{Benchmark prompt template}
Each benchmark instance is rendered into a common prompt with three fields: \textbf{Task}, optional \textbf{Guidance}, and \textbf{Deliverables}. The task states the study objective. Guidance supplies approach-level or method-level constraints for \textsc{Goal+Approach} and \textsc{Goal+Method}. Deliverables require a validated RWS-IR, one executable entry point with generated artifacts, and a structured report with assumptions, limitations, and evidence certificates. Hallucinated tables or columns are disallowed.

\subsection{IR compiler prompt template}
The IR compiler receives the user prompt, high-level structural profile, high-level knowledge profile, output contract, and claim-policy rules. It returns JSON containing study design, analysis grain, join graph, entity keys, temporal anchors, variable-role bindings, method specification, required artifacts, assumptions, and allowed or blocked claim types. If the profile does not support a requested variable or method, the compiler must either downgrade the task or return an infeasibility trace.

\subsection{Coding harness prompt template}
The coding harness receives the accepted RWS-IR, low-level field profile, selected sheets, local figure policy, artifact schema, and any previous violation set. It writes one executable entry point, generates declared artifacts, and produces figure and result manifests. Repairs must address the localized violation set without introducing fields or methods absent from the profile.

\subsection{Report and certificate prompt template}
The report generator receives the accepted RWS-IR, result artifacts, figure manifest, execution logs, assumptions, limitations, and claim-policy rules. Each factual sentence must be paired with an evidence certificate or rewritten as a limitation. The generator must not convert descriptive proxy artifacts into causal, safety-critical, regulatory, or treatment-effect claims.

\subsection{Judge prompts}
The judge receives the ground-truth data profile $\Phi$ so that schema hallucinations can be checked against the same table and field constraints used by the validators.

\paragraph{System prompt.} Defines the reviewer role and scoring constraints.

\begin{promptbox}{Prompt}
\label{prompt:judge_system}
\textbf{Prompt 1. Expert Judge System Prompt}

You are a Senior Clinical Data Scientist reviewing Real-World Study (RWS) automation artifacts.

Your task is to evaluate an AI agent's artifacts against a specific \textbf{Ground Truth Schema}.

\textbf{Scoring rules:}
\begin{enumerate}[left=0pt, nosep]
    \item \textbf{Schema Hallucination = 0:} Any usage of tables/columns not present in the provided \texttt{[DATA\_PROFILE]} results in a score of 0 for relevant metrics, regardless of code quality.
    \item \textbf{Consistency:} The numbers in the Final Report must be consistent with the Execution Logs.
    \item \textbf{Structured rationale:} Provide a concise evidence-based rationale before assigning JSON scores.
\end{enumerate}
\end{promptbox}

\paragraph{User prompt.} Injects the profile and artifact bundle.

\begin{promptbox}{Prompt}
\label{prompt:judge_user}
\textbf{Prompt 2. Context-Injection User Prompt}

\textbf{[PART 1: GROUND TRUTH CONTEXT]}\\[0.3em]
\textsc{User Query:} \{user\_query\}\\[0.3em]
\textsc{Data Profile Constraints ($\Phi$):}\\
\textit{(Valid tables, columns, and join keys derived from the dataset)}\\
\{profile\_snapshot\_str\}

\vspace{0.6em}
\textbf{[PART 2: AGENT ARTIFACTS]}
\begin{enumerate}[left=0pt, nosep, label=\arabic*.]
    \item \textsc{RWS-IR:} \{agent\_rws\_ir\_json\}
    \item \textsc{Code:} \{generated\_code\}
    \item \textsc{Logs:} \{execution\_logs\}
    \item \textsc{Visuals:} \{list\_of\_generated\_files\}
    \item \textsc{Evidence Certificates:} \{evidence\_certificate\_json\}
    \item \textsc{Report:} \{final\_report\_content\}
\end{enumerate}

\vspace{0.6em}
\textbf{[PART 3: TASK]}\\
Analyze the artifacts and score them based on the rubric.

\textbf{Structured assessment}
\begin{itemize}[left=0pt, nosep]
    \item \textit{Schema Check:} Does the code use ONLY columns listed in [Data Profile Constraints]?
    \item \textit{Logic Check:} Is the statistical method appropriate for \{user\_query\}?
    \item \textit{Consistency:} Do chart findings match the log data?
\end{itemize}

\textbf{JSON output}\\
Return a JSON object with keys: GA, CQ, VC, CF, SL.\\
Format: \texttt{\{"Analysis": "...", "Scores": \{...\}, "Failure\_Mode": "None/Hallucination/Logic"\}}
\end{promptbox}

\section{Data Access, Privacy, and Responsible Use}
\label{app:responsible_release}
The released materials include the reproducible benchmark logic and the fully de-identified record-level HCC benchmark tables. The public package includes prompts, profiles, schema artifacts, aggregate summaries, de-identified HCC tables, HCC data dictionary, checksums, public-data manifests, and evaluation scripts. Direct participant identifiers and identity-linking fields are removed from the HCC release, while random benchmark subject and visit identifiers are retained to preserve reproducible table joins and longitudinal task execution. Public-RWS datasets are accessed from their original public sources, with source metadata and checksums recorded in the release documentation.

The HCC benchmark is released for research reproduction and auditing of RWS automation methods. The release terms prohibit attempted re-identification, participant linkage to external data sources, and clinical use of generated outputs. Any downstream deployment on institutional data requires local governance review, schema revalidation, and confirmation that the site-specific claim policy matches the available data.

ProCARE is evaluated as a research automation and auditability system, not as an autonomous clinical decision system. Evidence certificates establish traceability to implemented validators and artifacts. They do not establish clinical truth, causal validity, regulatory sufficiency, or prospective safety. The primary HCC benchmark is single-center and disease-specific. Public-RWS-18 tests schema transfer, but it does not replace multi-institutional longitudinal RWS validation. Porting ProCARE to OMOP, FHIR, or another institutional schema requires rebuilding the profile, revalidating RWS-IR checks, and confirming that claim policies match the new data source and study design.

\end{document}